\documentclass[preprint,12pt,authoryear]{elsarticle}

\usepackage[T1]{fontenc}
\usepackage{amssymb}
\usepackage{amsmath}
\usepackage{booktabs}
\usepackage{multirow}
\usepackage{tabularx}
\usepackage{array}
\usepackage{xcolor}
\usepackage{url}
\usepackage{hyperref}
\usepackage{rotating}
\usepackage[section]{placeins}
\usepackage{tikz}
\usetikzlibrary{arrows.meta,positioning}

\journal{AI Open}

\graphicspath{{figures/}}
\newcolumntype{Y}{>{\raggedright\arraybackslash}X}
\newcolumntype{P}[1]{>{\raggedright\arraybackslash}p{#1}}
\newcommand{\IEEEPARstart}[2]{#1#2}

\begin{document}

\begin{frontmatter}

\title{Intelligent Automation for Embodied Benchmark Construction: Pipelines, Embodiments, Simulators, and Trends}

\author[inst1]{Jinshan Lai}
\author[inst2]{Jianwei Hu}
\author[inst2]{Baoyang Jiang}
\author[inst1]{Fengchun Zhang}
\author[inst3]{Leyuan Wang}
\author[inst3]{Haotian Li}
\author[inst3]{Yida Wang}
\author[inst4]{Tingxuan Huang}
\author[inst5]{Xi Ren}
\author[inst2]{Qiang Ma\corref{cor1}}
\ead{maqiang@qiyuanlab.com}
\cortext[cor1]{Corresponding author.}

\affiliation[inst1]{
  organization={University of Electronic Science and Technology of China},
  city={Chengdu},
  country={China}
}

\affiliation[inst2]{
  organization={Qiyuan Lab},
  city={Beijing},
  country={China}
}

\affiliation[inst3]{
  organization={Beijing University of Posts and Telecommunications},
  city={Beijing},
  country={China}
}

\affiliation[inst4]{
  organization={Tsinghua University},
  city={Beijing},
  country={China}
}

\affiliation[inst5]{
  organization={Beihang University},
  city={Beijing},
  country={China}
}

\begin{abstract}
Embodied intelligence now spans navigation, household assistance, manipulation, autonomous driving, aerial agents, and multimodal large-model control. This expansion has made benchmark construction a central bottleneck for reliable evaluation. Unlike static datasets, embodied benchmarks combine task specifications, environments, robot data, demonstrations, annotations, metrics, evaluation scripts, and release policies into a single evaluation system. This survey reviews the literature through a five-stage construction pipeline: requirement and task construction, data acquisition, data cleaning and annotation, benchmark suite generation and metric definition, and evaluation execution with diagnostic feedback. For each stage, the survey analyzes the transition from manual curation to traditional automation, foundation-model assistance, and agentic closed-loop workflows. It also compares qualitative construction costs across human labor, data and asset acquisition, compute and simulation, validation and debugging, governance and maintenance, and rework risk. The main conclusion is that automation does not simply reduce benchmark cost. Instead, it often shifts cost toward validation, auditability, version control, and long-term governance. Progress in embodied evaluation will therefore depend not only on larger benchmark suites, but also on construction pipelines that are diagnosable, auditable, and responsibly refreshable.
\end{abstract}



\end{frontmatter}

\section{Introduction}
\label{sec:introduction}

\IEEEPARstart{E}{mbodied} intelligence has grown from a relatively narrow focus on navigation and perception into a broader research program that includes language grounding, household activity execution, manipulation, social interaction, and generalist robot learning. Progress in simulation, robot data collection, and foundation models for vision, language, and action has accelerated this shift \cite{zhou2024foundation,luo2024cyberphysical,brohan2022rt1,brohan2023rt2,driess2023palme,feng2025embodiedagi}. The same progress has also exposed a less visible limitation: in many settings, the quality and maintainability of the benchmark now constrain what can be inferred from model results. Benchmark construction can no longer be treated as a background engineering step. As models, simulators, and robot datasets scale, the ability to build, validate, update, and govern benchmarks has become a research problem in its own right.

The difficulty is not only that embodied tasks are harder than static image or text recognition. An embodied benchmark is a compound artifact that combines environment assets, object semantics, physical rules, task templates, language interfaces, trajectories or demonstrations, success conditions, evaluation scripts, baselines, and release governance. Representative suites in navigation and household assistance, manipulation, real-robot data, and embodied perception show the same pattern: the benchmark is produced through a chain of design, acquisition, annotation, suite assembly, execution, and maintenance rather than released as a single static dataset \cite{savva2019habitat,ramakrishnan2021hm3d,deitke2020robothor,shridhar2020alfred,padmakumar2022teach,li2022igibson2,savva2021behavior,james2020rlbench,gu2023maniskill2,liu2023libero,khazatsky2024droid,openx2023,wang2024embodiedscan,yenamandra2023homerobot}.

This construction view differs from the usual treatment of training corpora. Training data are often optimized for coverage, scale, and annotation throughput. Evaluation benchmarks must also preserve hidden-test integrity, metric stability, documented updates, contamination resistance, and interpretable failure reports. Work on Dynabench, DataPerf, BenchmarkCards, BetterBench, Audit Cards, and GPAI evaluation standards has made a similar argument in other areas of AI: evaluation infrastructure requires governance rules rather than only larger datasets \cite{kiela2021dynabench,mazumder2023dataperf,sokol2024benchmarkcards,reuel2024betterbench,staufer2025auditcards,paskov2024gpai}. Embodied AI inherits these concerns and adds simulator drift, asset licensing, embodiment interfaces, hardware safety envelopes, and release policies for interactive testbeds.

The bottleneck has shifted. Earlier embodied work was often limited by model capacity and environment availability. The central question is no longer only whether agents can be trained in richer environments, but whether the benchmark can expose the specific reasons that an embodied system fails. Many current research questions instead depend on whether a benchmark can reveal long-horizon reasoning failures, contact errors, unsafe plans, domain gaps, or weak cross-embodiment generalization. Navigation and instruction-following benchmarks established standardized measurements for route completion, spatial grounding, and dialogue-conditioned execution \cite{anderson2018vision,das2018embodiedqa,thomason2020cvdn,qi2020reverie,ku2020rxr,savva2019habitat}. Household and manipulation benchmarks added state changes, object affordances, articulated assets, and long-horizon task decomposition \cite{shridhar2020alfred,shridhar2021alfworld,savva2021behavior,srivastava2022behavior1k,gu2023maniskill2,mees2022calvin}. Recent suites for VLA and MLLM agents further introduce open-ended instructions, safety constraints, structured error taxonomies, and governance-oriented metrics \cite{zhang2024vlabench,li2024eai,cheng2025embodiedeval,yang2025embodiedbench,yin2024safeagentbench,qin2026embodiedgovbench}.

The construction process has changed as well. Manual curation remains central for value judgment, task relevance, and safety review, but several stages are now partly automated. Traditional automation includes programmatic task templates, scripted solvers, procedural generation, GPU-based simulation, and regression-based evaluation servers \cite{yu2020metaworld,zhu2020robosuite,makoviychuk2021isaacgym,deitke2022procthor}. Foundation-model-assisted workflows use LLMs or VLMs to generate tasks, rewrite language, suggest rewards, propose layouts, or pre-annotate semantic content \cite{ahn2022saycan,liang2023codeaspolicies,wang2023gensim,ma2024holodeck,ma2024eureka,huang2023voxposer}. Recent work moves further toward agentic closed-loop construction, where systems propose tasks, generate environments, run solvers, filter failures, and revise benchmarks under limited human supervision \cite{hua2024gensim2,yang2023robogen,vuong2024dreureka,fan2024grutopia,xu2024infiniteworld,li2026affordsim}. These automation levels describe how benchmark artifacts are constructed and checked. They should not be read as a quality ranking, because a carefully maintained traditional benchmark may be more reliable than a poorly audited foundation-model-assisted or agentic benchmark.

Despite these advances, the literature remains fragmented. Existing surveys usually focus on embodied AI capabilities, robot foundation models, or LLM-based agents rather than on how benchmarks themselves are produced and maintained \cite{zhou2024foundation,luo2024cyberphysical,zhang2025llmagent}. As a result, key design patterns such as simulator-independent task descriptions, demonstration scaling, embodied perception-grounded annotation, automatic failure diagnosis, and benchmark version governance are rarely discussed under one unified framework.

This paper surveys intelligent automation for embodied benchmark construction. Instead of organizing the literature only by application domain, it treats the benchmark itself as a production system and asks how its artifacts are specified, generated, validated, released, and maintained. The review covers benchmark and tooling papers available through May 12, 2026, including archival publications, official project pages, and recent arXiv preprints when they define emerging benchmark families. Four questions guide the survey: how embodied benchmarks have evolved across task and embodiment regimes, which stages make up benchmark construction, how far each stage has been automated and by which techniques, and how automation changes scale, reproducibility, diagnostic power, and governance. The unit of analysis is therefore the benchmark-construction process rather than a model family, a robot carrier, or a single leaderboard score.

\subsection{Survey Methodology and Scope}

Because the survey spans benchmark papers, simulator papers, dataset and tooling releases, and governance-oriented evaluation work, it follows an explicit screening protocol rather than an ad hoc reading list. The search covered IEEE Xplore, the ACM Digital Library, Google Scholar, arXiv, and official benchmark or project pages when those pages served as primary release records. Search strings combined terms such as embodied benchmark, robot benchmark, embodied evaluation, manipulation benchmark, autonomous driving simulator, aerial embodied benchmark, VLA benchmark, benchmark governance, and embodied world model. A record was retained when it introduced or substantially redefined a benchmark, simulator substrate, evaluation suite, benchmark-construction toolchain, or release and governance framework directly relevant to embodied evaluation.

Papers were excluded when their contribution was purely algorithmic and did not define a reusable benchmark, simulator, dataset, or evaluation interface. Duplicates across archival venues, arXiv versions, and project pages were removed, with the most informative primary source retained when possible. The bibliography used in the current draft contains 110 cited sources. It includes benchmark-defining papers, simulator papers, dataset and tooling releases, governance-aware evaluation systems, and a smaller set of surveys and documentation frameworks used for positioning. For the stage-wise benchmark matrix used during drafting, each candidate benchmark was coded only when a primary paper, official release page, or documented repository provided evidence about its construction process. When the evidence did not justify a higher automation level, the coding remained conservative rather than inferred from later reuse or community reputation. The protocol does not claim exhaustive coverage of every embodied paper; its purpose is to make the benchmark-construction corpus auditable, reproducible, and aligned with the scope of the review.

\subsection{Positioning Against Existing Surveys}

Prior reviews remain important, but they use different units of analysis. Most embodied AI surveys organize the literature by capability, model family, or application domain. Surveys on foundation models emphasize pretraining, transfer, and policy generalization. Agent-evaluation surveys focus on LLM-based evaluation methodology more broadly. Governance-oriented benchmark work typically discusses documentation and audit structure, but not embodied construction pipelines. Table~\ref{tab:survey-positioning} summarizes this distinction.

\begin{table*}[t]
\caption{Positioning against existing survey families.}
\label{tab:survey-positioning}
\centering
\scriptsize
\setlength{\tabcolsep}{1.5pt}
\renewcommand{\arraystretch}{1.1}
\begin{tabularx}{\textwidth}{@{}p{2.2cm} Y Y Y Y Y Y@{}}
\toprule
\textbf{Survey Family} & \textbf{Embodied Bmks.} & \textbf{Pipeline} & \textbf{Automation} & \textbf{Carrier-Aware} & \textbf{Sim. / World Models} & \textbf{Governance} \\
\midrule
General embodied AI surveys & Yes & No & Partial & Partial & Partial & No \\
Robot foundation model surveys & Partial & No & Partial & Partial & Partial & No \\
LLM / agent evaluation surveys & Partial & No & Yes & No & No & Partial \\
Benchmark governance work & Partial & No & No & No & No & Yes \\
\textbf{This survey} & \textbf{Yes} & \textbf{Yes} & \textbf{Yes} & \textbf{Yes} & \textbf{Yes} & \textbf{Yes} \\
\bottomrule
\end{tabularx}
\end{table*}

Representative prior surveys in these families include the general embodied AI reviews of \cite{luo2024cyberphysical,feng2025embodiedagi}, the robot foundation model survey of \cite{zhou2024foundation}, the agent-evaluation survey of \cite{zhang2025llmagent}, and benchmark documentation / governance work such as \cite{sokol2024benchmarkcards,reuel2024betterbench,staufer2025auditcards,paskov2024gpai}. Table~\ref{tab:survey-positioning} clarifies that the present manuscript studies a different object of analysis, namely the benchmark-construction pipeline itself.

The survey makes three contributions. First, it formulates embodied benchmark construction as a five-stage pipeline that connects requirements, data, labels, suites, metrics, execution, and diagnostic feedback. Second, it compares each stage using the same four-level automation ladder, formalized below as the Construction Automation and Auditability Rubric (CAAR): manual construction, traditional automation, foundation-model assistance, and agentic closed-loop construction. Third, it introduces a qualitative cost lens showing that automation does not simply reduce construction cost; it often transfers effort from manual authoring and collection to validation, debugging, audit, governance, and later rework. These contributions give the paper a single analytical frame for reading diverse benchmarks across navigation, household activity, manipulation, driving, aerial systems, legged robots, real-robot datasets, and VLA/MLLM evaluation.

Several anchor benchmarks recur across the paper to keep this frame concrete. ALFRED illustrates how task schemas, language, demonstrations, annotations, and metrics interact in household instruction following. BEHAVIOR and BEHAVIOR-1K show how object states and activity definitions reshape acquisition, annotation, and scoring. Habitat and Habitat 2.0 show the role of executable simulation and challenge infrastructure. DROID and Open X-Embodiment show the acquisition and governance burden of real-robot data. Recent systems such as GenSim, RoboGen, GRUtopia, SafeAgentBench, EmbodiedEval, and EmbodiedGovBench illustrate the move toward foundation-model-assisted and agentic construction.

\begin{figure}[!htbp]
\centering
\includegraphics[width=0.98\textwidth]{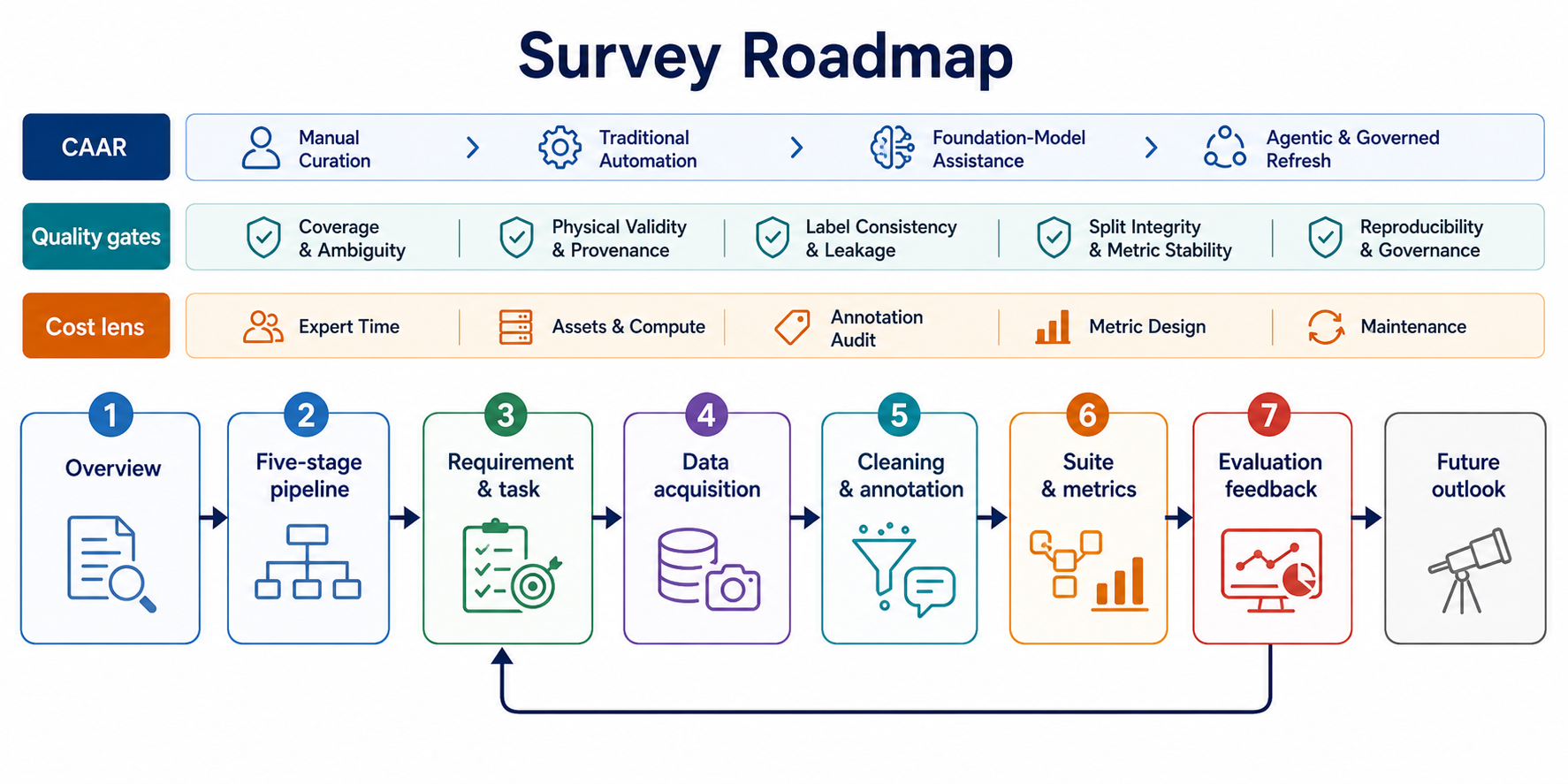}
\caption{Roadmap view of the revised survey structure. The paper first defines the five-stage pipeline and then studies each stage through automation level, quality gates, and qualitative construction cost.}
\label{fig:paper-architecture}
\end{figure}

Figure~\ref{fig:paper-architecture} summarizes the manuscript structure. The paper is not a catalogue of benchmark papers. It first defines the five-stage construction process and then assigns one chapter to each stage, so that automation level, quality control, and construction cost can be compared with a consistent vocabulary.

Section~\ref{sec:pipeline} defines the five-stage pipeline, CAAR automation levels, cost dimensions, and the stage-wise cost matrix. Sections~\ref{sec:automation}--\ref{sec:evaluation-execution-diagnostic-feedback} study the five stages in order: requirement and task construction, data acquisition, data cleaning and annotation, benchmark suite generation and metric definition, and evaluation execution with diagnostic feedback. Section~\ref{sec:future} discusses open challenges, cost transfer, and benchmark compilers. Section~\ref{sec:conclusion} concludes the paper. The future of embodied evaluation is therefore not only larger benchmarks, but maintainable benchmark-construction pipelines.

\section{Five-Stage Pipeline for Embodied Benchmark Construction}
\label{sec:pipeline}

An embodied benchmark is more informative when analyzed as a production pipeline rather than as a finished dataset. Capability requirements, task schemas, scenes, trajectories, annotations, metrics, evaluation traces, and release records are produced at different points, and each artifact has distinct failure modes. Automation also changes the cost structure of these artifacts. A method that reduces manual authoring may increase validation work, simulation cost, audit burden, or later rework. The pipeline is also not a one-way assembly line. Evaluation failures should feed back into upstream requirements, data, labels, and metrics, because many benchmark defects become visible only after agents are actually run.

\begin{figure*}[t]
\centering
\includegraphics[width=0.98\textwidth]{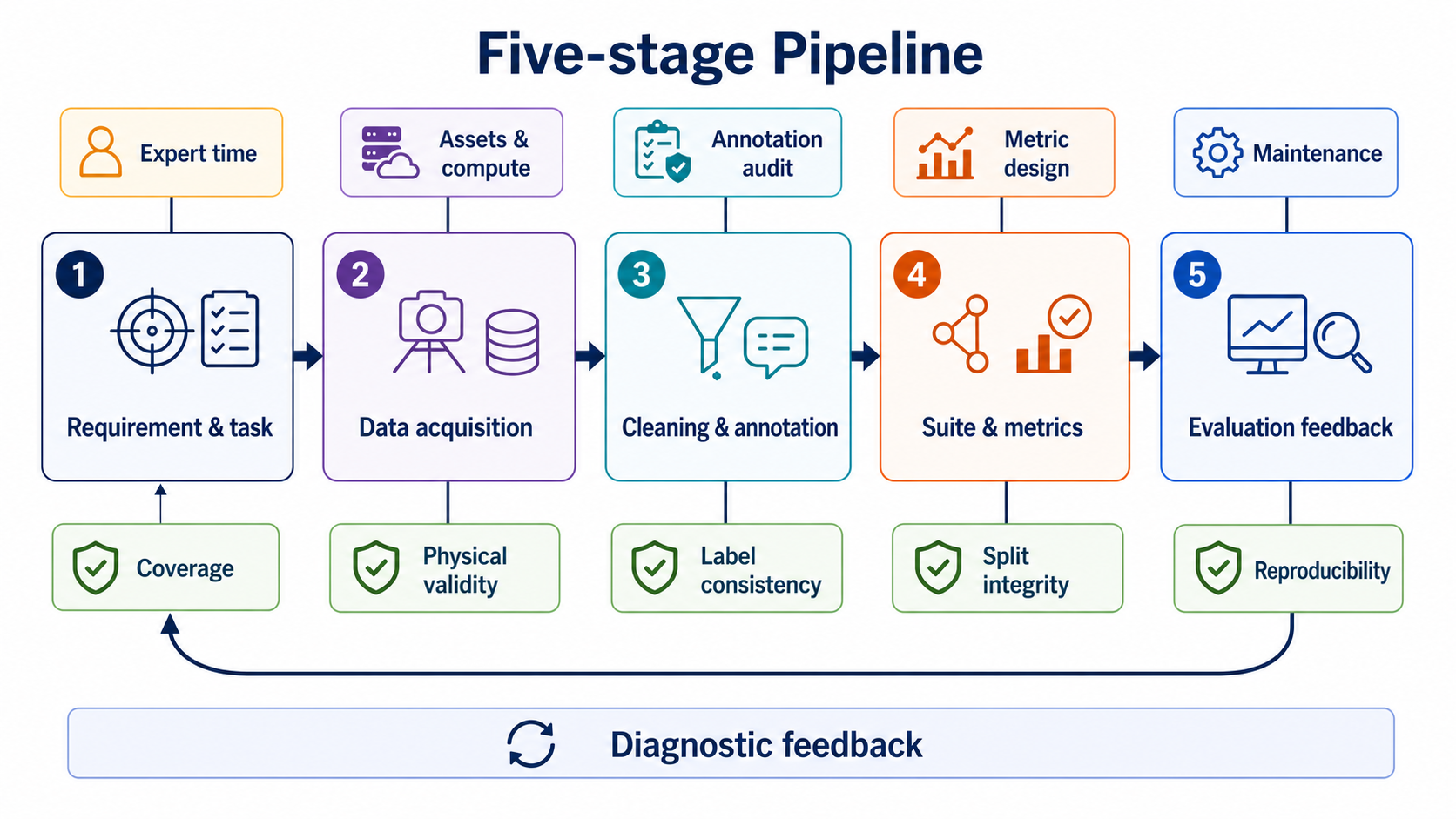}
\caption{Five-stage pipeline used in this survey. Each stage produces a different benchmark artifact, carries a different dominant cost, and requires its own quality gate. The feedback arrow indicates that diagnostic failures and release experience should revise the upstream requirement and task design rather than merely update leaderboard scores.}
\label{fig:pipeline}
\end{figure*}

The paper uses five stages: requirement and task construction, data acquisition, data cleaning and annotation, benchmark suite generation and metric definition, and evaluation execution and diagnostic feedback. Figure~\ref{fig:pipeline} presents this process view. The middle row gives the stages, the lower row identifies the quality gates that limit downstream error propagation, and the upper row summarizes the main cost profile of each stage. The first stage connects capability requirements to executable task schemas. The second stage combines environment, asset, simulator, and trajectory acquisition. The third stage links cleaning with semantic and diagnostic annotation. The fourth stage covers suite assembly and metrics, while the fifth stage covers execution, diagnosis, release feedback, and maintenance.

\begin{table*}[t]
\caption{Running examples showing how one benchmark can involve all five construction stages.}
\label{tab:running-examples}
\centering
\scriptsize
\setlength{\tabcolsep}{2pt}
\renewcommand{\arraystretch}{1.04}
\begin{tabularx}{\textwidth}{@{}p{2.2cm} Y Y Y Y Y@{}}
\toprule
\textbf{Anchor} & \textbf{Task Construction} & \textbf{Data Acquisition} & \textbf{Cleaning / Annotation} & \textbf{Suite and Metrics} & \textbf{Evaluation Feedback} \\
\midrule
ALFRED & Household goals, language instructions, subgoal structure & AI2-THOR scenes and demonstrations & Object states, action traces, and subgoal labels & Seen/unseen splits and task success metrics & Failure analysis for grounding, planning, and state change \\
BEHAVIOR / BEHAVIOR-1K & Activity definitions and object-state requirements & Interactive household scenes, objects, and task assets & Predicate, state, and affordance checks & Long-horizon task suites and goal-state metrics & Diagnosis of infeasible states and activity coverage gaps \\
Habitat family & Navigation, rearrangement, and interaction tasks & Scanned or simulated indoor environments & Scene metadata, object states, and simulator validation & Challenge splits, SPL-style metrics, and baselines & Leaderboard traces and simulator-version maintenance \\
DROID / Open X-Embodiment & Real-robot manipulation task framing & Large-scale robot trajectories and demonstrations & Sensor alignment, task metadata, and trajectory filtering & Dataset splits and policy-evaluation protocols & Release governance, data rights, and cross-embodiment comparability \\
RoboGen / SafeAgentBench & Model-assisted or safety-oriented task generation & Generated simulation assets or hazardous scenarios & Validity filtering, safety labels, and audit checks & Diagnostic metrics for failures, unsafe actions, or invalid plans & Closed-loop revision under explicit audit constraints \\
\bottomrule
\end{tabularx}
\end{table*}

Table~\ref{tab:pipeline-summary} summarizes the five-stage view. The representative systems are not exhaustive; they indicate where the literature provides concrete examples of each stage. The table also introduces recurring quality gates. A coverage check asks whether the benchmark spans the intended capability space rather than only the easiest cases to generate. An ambiguity check asks whether tasks, instructions, states, and success conditions can be interpreted consistently by users and maintainers. A leakage check asks whether evaluation samples, hidden tests, labels, or metric definitions could be exposed through training data, public examples, or model-assisted construction tools.

\begin{table*}[t]
\caption{Five-stage benchmark construction pipeline, main artifacts, and recurring quality gates.}
\label{tab:pipeline-summary}
\centering
\tiny
\setlength{\tabcolsep}{2pt}
\renewcommand{\arraystretch}{1.02}
\begin{tabularx}{\textwidth}{@{}p{2.6cm} Y Y Y Y@{}}
\toprule
\textbf{Stage} & \textbf{Core Artifacts} & \textbf{Typical Automation Levers} & \textbf{Quality Gates} & \textbf{Representative Systems} \\
\midrule
Requirement and task construction & Capability targets, task schemas, preconditions, success and failure conditions & Templates, symbolic task grammars, LLM-generated task variants, agentic task refinement & Coverage check, ambiguity check, safety and value-boundary review & ALFRED, BEHAVIOR-1K, RLBench, GenSim, VLABench \cite{shridhar2020alfred,srivastava2022behavior1k,james2020rlbench,wang2023gensim,zhang2024vlabench} \\
Data acquisition & Environments, assets, demonstrations, trajectories, sensor streams, synthetic scenes & Simulator resets, procedural generation, scripted collection, synthetic demonstrations, world-model generation & Physical validity, provenance, embodiment-interface consistency, sim-to-real plausibility & Matterport3D, Habitat, ProcTHOR, DROID, RoboGen \cite{chang2017matterport3d,savva2019habitat,deitke2022procthor,khazatsky2024droid,yang2023robogen} \\
Data cleaning and annotation & Semantic labels, state labels, action traces, failure tags, affordance labels, metadata & Rule filters, replay validation, VLM pre-annotation, weak supervision, self-filtering audits & Label consistency, temporal alignment, hidden-state validity, contamination and leakage checks & ScanRefer, Ego4D, EmbodiedScan, ALFRED, MimicGen \cite{chen2020scanrefer,grauman2021ego4d,wang2024embodiedscan,shridhar2020alfred,mandlekar2023mimicgen} \\
Benchmark suite generation and metric definition & Splits, hidden tests, scoring scripts, baselines, diagnostic protocols, benchmark cards & Automatic split generation, template metrics, LLM-assisted metric design, solver-based sanity checks & Split integrity, metric stability, baseline sanity, gaming resistance & Habitat Challenge, ALFRED, BEHAVIOR-1K, EvalAI, RoboEval \cite{savva2019habitat,shridhar2020alfred,srivastava2022behavior1k,wu2019evalai,wang2025roboeval} \\
Evaluation execution and diagnostic feedback & Evaluation server, logs, traces, leaderboards, error taxonomies, update records & Continuous evaluation, automatic trace analysis, model-assisted diagnosis, dynamic benchmark refresh & Reproducibility, versioning, auditability, refresh governance & Dynabench, EmbodiedEval, EmbodiedBench, SafeAgentBench, EmbodiedGovBench \cite{kiela2021dynabench,cheng2025embodiedeval,yang2025embodiedbench,yin2024safeagentbench,qin2026embodiedgovbench} \\
\bottomrule
\end{tabularx}
\end{table*}

\subsection{Automation Levels}

To compare systems without reducing automation to a single number, the survey uses a Construction Automation and Auditability Rubric (CAAR). CAAR is a coding scheme for reading the literature rather than a universal score to optimize. It is descriptive rather than normative: A2 or A3 does not mean that a benchmark is better, only that more of its construction process is assisted by foundation models or closed-loop agents. It separates four construction modes that recur across benchmark papers. Manual construction relies on experts to define, author, annotate, and inspect benchmark artifacts. Traditional automation uses scripts, templates, procedural rules, solvers, and evaluation servers. Foundation-model-assisted construction uses LLMs, VLMs, MLLMs, VLAs, or generative models to propose tasks, scenes, annotations, metrics, or diagnoses, while humans remain responsible for acceptance. Agentic closed-loop construction allows a system to propose artifacts, run checks, diagnose failures, and revise the benchmark under bounded human oversight. When a paper does not provide direct evidence that a stage was automated, the safer code is manual or unspecified rather than an inferred higher level. This conservative rule is important because many benchmark papers use automated tools during implementation without documenting them as part of the released benchmark-construction process.

\begin{table}[t]
\caption{Construction Automation and Auditability Rubric (CAAR) used to describe stage-wise maturity.}
\label{tab:caar-rubric}
\centering
\scriptsize
\setlength{\tabcolsep}{3pt}
\renewcommand{\arraystretch}{1.12}
\begin{tabularx}{\textwidth}{@{}p{2.3cm} Y Y Y@{}}
\toprule
\textbf{Level} & \textbf{Automation Coverage} & \textbf{Artifact Verifiability} & \textbf{Governance Maturity} \\
\midrule
Manual & Most artifacts are authored, collected, or inspected by humans & High for small suites, but limited by reviewer fatigue and local expertise & Informal release notes, static versions, limited audit trail \\
Traditional automation & Scripts, templates, procedural generators, replay validators, and evaluation servers automate bounded operations & Strong when artifacts are executable or rule-checkable & Versioning and challenge infrastructure may exist, but policies remain benchmark-specific \\
Foundation-model-assisted & Foundation models propose tasks, language, labels, scenes, rewards, or diagnoses & Variable; generated content needs external validation and human review & Documentation must record prompts, model versions, filtering rules, and known failure modes \\
Agentic closed-loop & Agents iteratively generate, test, filter, diagnose, and refresh benchmark artifacts & Potentially high if every loop has independent validators; weak if self-evaluation is trusted alone & Requires audit logs, rollback policies, hidden-test governance, and human accountability \\
\bottomrule
\end{tabularx}
\end{table}

\subsection{Construction Cost Dimensions}

Construction cost is treated qualitatively. Most benchmark papers do not report comparable dollar budgets, person-hours, robot uptime, or GPU-hours, and mixing anecdotal numbers would imply a level of precision that the literature does not support. Costs are therefore compared by their dominant source. The central hypothesis is cost transfer rather than cost elimination. Automation can reduce visible authoring, collection, labeling, or execution effort, but it often increases the less visible cost of validation, provenance tracking, audit, governance, and repair after release. Human labor cost covers expert authoring, teleoperation, annotation, review, and release management. Asset and data acquisition cost covers scanned scenes, object libraries, robot hardware, sensor calibration, demonstrations, and data rights. Compute and simulation cost covers rendering, physics, procedural generation, model inference, and large-scale evaluation runs. Validation and debugging cost covers solvability checks, replay failures, metric bugs, simulator drift, annotation audits, and baseline sanity tests. Governance and maintenance cost covers documentation, licensing, benchmark cards, versioning, hidden-test rotation, safety review, and leaderboard operation. Rework risk captures the chance that downstream failures force benchmark builders to redesign earlier artifacts.

\begin{table}[t]
\caption{Qualitative cost dimensions used in Sections~\ref{sec:automation}--\ref{sec:evaluation-execution-diagnostic-feedback}.}
\label{tab:cost-dimensions}
\centering
\scriptsize
\setlength{\tabcolsep}{3pt}
\renewcommand{\arraystretch}{1.12}
\begin{tabularx}{\textwidth}{@{}p{3.0cm} Y Y@{}}
\toprule
\textbf{Cost Dimension} & \textbf{Meaning in Benchmark Construction} & \textbf{Typical Failure if Underestimated} \\
\midrule
Human labor cost & Expert design, manual collection, annotation, review, and maintenance & Narrow task coverage, slow updates, inconsistent review \\
Asset/data acquisition cost & Scenes, objects, demonstrations, trajectories, sensors, robot operation, and data rights & Poor realism, limited embodiment coverage, licensing gaps \\
Compute/simulation cost & Rendering, physics, simulator rollout, model inference, and synthetic generation & Small-scale validation, unstable generation, unreproducible runs \\
Validation/debugging cost & Solvability, replay, annotation, metric, baseline, and simulator checks & Invalid tasks, broken labels, hidden leakage, inflated scores \\
Governance and maintenance cost & Documentation, versioning, hidden tests, release policy, audit logs, safety review & Benchmark drift, weak accountability, poor comparability over time \\
Rework risk & Probability that downstream failures require upstream redesign & Expensive late-stage repairs and inconsistent benchmark versions \\
\bottomrule
\end{tabularx}
\end{table}

\subsection{Stage-Wise Cost Matrix}

Automation rarely removes cost; more often, it changes where the cost is paid. Table~\ref{tab:automation-evolution} makes this transfer explicit. Manual construction concentrates cost in expert labor and slow review. Traditional automation reduces repeated work but introduces engineering and simulator-maintenance cost. Foundation-model-assisted workflows expand scale and linguistic or semantic diversity, but they raise validation cost because generated artifacts may appear plausible without being physically valid or evaluation-safe. Agentic closed-loop workflows can reduce the cost of continuous refresh, but they make auditability, comparability, rollback, and governance more demanding.

\begin{table*}[t]
\caption{Stage-wise construction-cost shift across automation levels. Entries are qualitative cost bottlenecks rather than reported monetary budgets.}
\label{tab:automation-evolution}
\centering
\scriptsize
\setlength{\tabcolsep}{3pt}
\renewcommand{\arraystretch}{1.15}
\begin{tabularx}{\textwidth}{@{}p{2.5cm} Y Y Y Y@{}}
\toprule
\textbf{Pipeline Stage} & \textbf{Manual} & \textbf{Traditional Automation} & \textbf{Foundation-Model-Assisted} & \textbf{Agentic Closed-Loop} \\
\midrule
Requirement and task construction & High expert-design cost; low scale & Lower template-writing cost, but grammar design becomes a bottleneck & Lower ideation cost; higher ambiguity and safety-review cost & Lower refresh cost; higher accountability and scope-control cost \\
Data acquisition & High collection and asset cost; slow coverage growth & Lower repeated collection cost; higher simulator and scripting cost & Lower scene and demonstration expansion cost; higher generation-validation cost & Lower continuous acquisition cost; higher compute, audit, and rollback cost \\
Data cleaning and annotation & High human labeling cost; variable consistency & Lower rule-check cost; higher rule-maintenance cost & Lower pre-labeling cost; higher hallucination and hidden-state verification cost & Lower repeated filtering cost; higher independent audit and leakage-control cost \\
Benchmark suite and metrics & High expert assembly cost; stable but narrow metrics & Lower split and scoring cost; higher infrastructure and baseline cost & Lower metric drafting cost; higher gaming and construct-validity review cost & Lower dynamic update cost; higher comparability and governance cost \\
Evaluation and feedback & High manual execution and diagnosis cost & Lower execution cost; higher server and regression-maintenance cost & Lower trace-summarization cost; higher explanation-validation cost & Lower refresh labor; higher versioning, audit, safety, and community-governance cost \\
\bottomrule
\end{tabularx}
\end{table*}

This matrix structures Sections~\ref{sec:automation}--\ref{sec:evaluation-execution-diagnostic-feedback}. Each section studies one pipeline stage, describes how automation has changed that stage, and returns to the same practical question: which cost is reduced, which cost is introduced, and which bottleneck remains. The entries should be read horizontally as cost movement within a stage, not vertically as a claim that later automation levels are always preferable. A benchmark may deliberately remain at a lower automation level when stability, interpretability, safety review, or community trust is more important than rapid expansion.

\section{Requirement and Task Construction}
\label{sec:automation}

\subsection{Role of Requirement and Task Construction}

Requirement and task construction defines what an embodied benchmark is intended to measure before data are collected, annotated, scored, or released. In embodied intelligence, broad capability claims such as navigation, household assistance, manipulation, autonomous driving, aerial reasoning, and safety-aware planning must be translated into task schemas that can be executed, evaluated, and audited \cite{anderson2018vision,shridhar2020alfred,savva2021behavior,james2020rlbench,dosovitskiy2017carla,shah2017airsim,yin2024safeagentbench}. These schemas specify capability targets, task boundaries, instruction formats, embodiment assumptions, observation and action interfaces, object and state vocabularies, preconditions, success conditions, failure conditions, and exclusion rules. The guiding question for this stage is therefore: what capability should the benchmark measure, and which task assumptions must be fixed before environments, demonstrations, annotations, metrics, or evaluation servers are built?

This stage is necessary because downstream construction can only operate on what the task schema makes explicit. In vision-and-language navigation, route instructions, goal locations, referring expressions, dialogue context, and multilingual grounding determine whether evaluation emphasizes route completion, localization, SPL, or language-grounded search \cite{anderson2018vision,thomason2020cvdn,qi2020reverie,ku2020rxr}. In household activity and manipulation benchmarks, high-level goals must be linked to subgoals, object states, affordances, action spaces, and goal predicates before demonstrations or success checks can be interpreted \cite{shridhar2020alfred,shridhar2021alfworld,savva2021behavior,srivastava2022behavior1k,gu2023maniskill2}. In embodied question answering and spatial reasoning, task definitions specify whether an agent answers from a static scene, a situated viewpoint, or an evidence-gathering interaction process \cite{das2018embodiedqa,azuma2022scanqa,ma2022sqa3d}. Thus, requirement construction defines the ontology of possible behavior for the rest of the benchmark pipeline.

\begin{figure*}[t]
\centering
\includegraphics[width=0.98\textwidth]{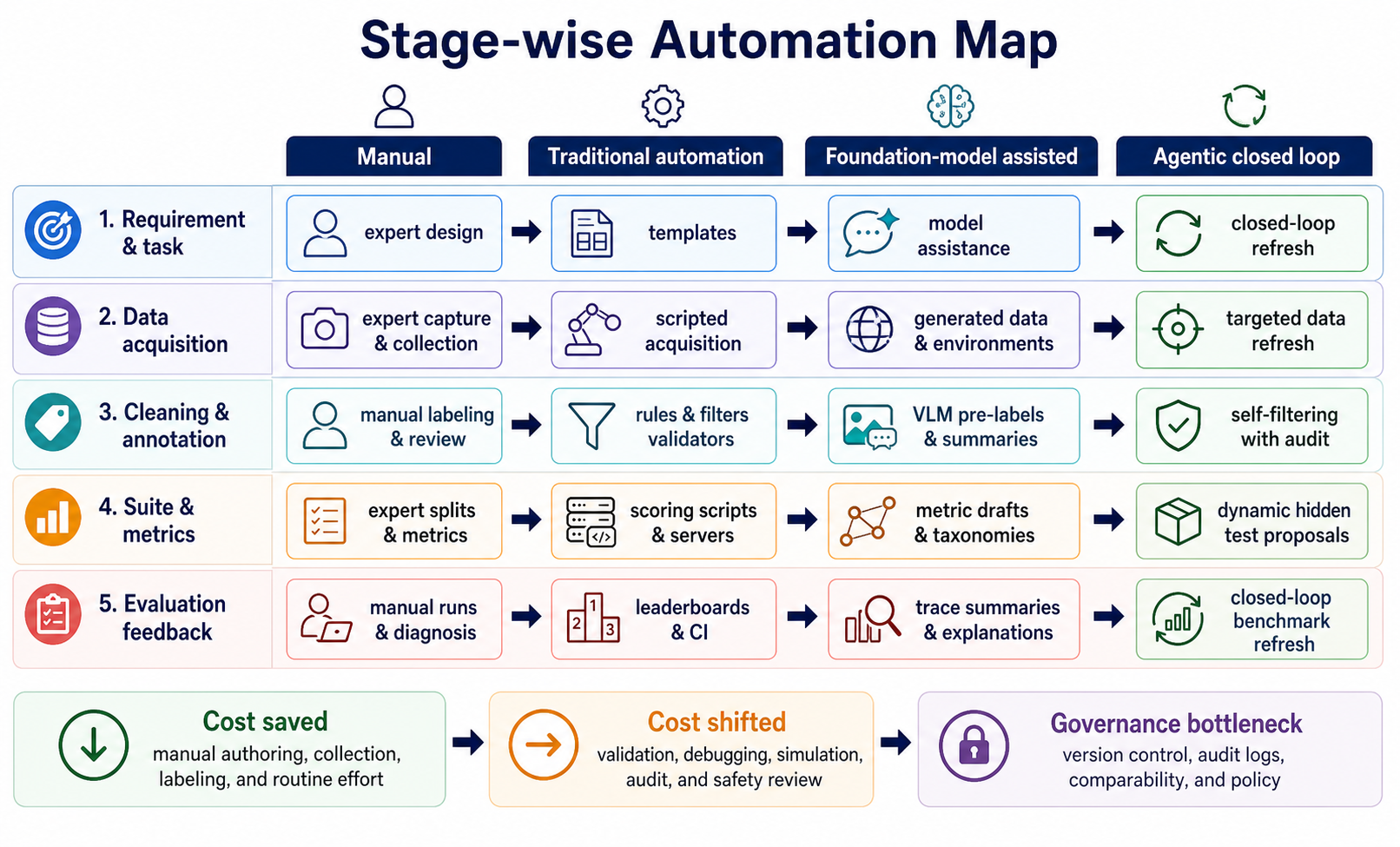}
\caption{Stage-wise automation map for embodied benchmark construction. Each row corresponds to one pipeline stage, and each column corresponds to one CAAR automation level. The bottom row states the main cost-transfer pattern: automation reduces some manual construction costs, but shifts effort toward validation, simulation, audit, version control, and governance.}
\label{fig:timeline}
\end{figure*}

Figure~\ref{fig:timeline} summarizes how requirement and task construction evolves across the four CAAR automation levels used throughout this survey. The figure also situates this stage relative to data acquisition, cleaning and annotation, suite generation and metric definition, and evaluation feedback. The key point is not that later automation levels are necessarily better, but that each level changes which parts of task construction are authored, generated, checked, and governed.

\subsection{Manual Expert Design}

Manual expert design refers to benchmark construction in which capability targets, task categories, success criteria, and exclusion rules are mainly specified by researchers, domain experts, or crowd-authored protocols. Room-to-Room and related vision-and-language navigation benchmarks define tasks through human instructions, routes, viewpoints, target goals, and path-following assumptions \cite{anderson2018vision,thomason2020cvdn,qi2020reverie,ku2020rxr}. ALFRED defines household instruction following through manually designed task types, natural-language instructions, subgoal sequences, object interactions, and goal conditions in AI2-THOR scenes \cite{kolve2017ai2thor,shridhar2020alfred}. BEHAVIOR and BEHAVIOR-1K define everyday activities through object states, predicates, and long-horizon household goals \cite{savva2021behavior,srivastava2022behavior1k}. ScanQA and SQA3D show a related pattern in 3D question answering, where the relation among scene content, spatial viewpoint, question, and answer space must be fixed before evaluation can be performed \cite{azuma2022scanqa,ma2022sqa3d}.

The main strength of manual task construction is conceptual control. Expert-designed schemas can specify which capabilities matter, which failures should count, which hazards are out of scope, and which simplifications are acceptable for scientific comparison. This is especially important when task construction involves ambiguous instructions, safety boundaries, norm-sensitive requests, or governance-relevant failures \cite{yin2024safeagentbench,qin2026embodiedgovbench}. The limitation is that manual construction expands slowly and may reflect the assumptions of a limited design team or annotation protocol. The dominant cost is therefore expert labor and consensus building, while the main quality gates are coverage, ambiguity, feasibility, and safety.

\subsection{Traditional Automation}

Traditional automation turns human-defined requirements into executable templates, task grammars, predicates, scripted samplers, and simulator APIs. In manipulation, RLBench, Meta-World, robosuite, and ManiSkill2 organize task families through parameterized objects, goal states, robot actions, scripted demonstrations, reward functions, or solver interfaces \cite{james2020rlbench,yu2020metaworld,zhu2020robosuite,gu2023maniskill2}. In navigation and household interaction, AI2-THOR, Habitat, Habitat 2.0, RoboTHOR, ProcTHOR, and ALFWorld make tasks executable through scene loading, episode definitions, simulator states, action interfaces, procedural house generation, or text-to-simulation abstractions \cite{kolve2017ai2thor,savva2019habitat,szot2021habitat2,deitke2020robothor,deitke2022procthor,shridhar2021alfworld}. In driving and mobility, CARLA and MetaDrive express requirements through configurable scenarios, maps, traffic actors, routes, and rule-compliance conditions \cite{dosovitskiy2017carla,li2021metadrive}.

This level reduces repeated task-authoring cost because task instances can be regenerated from code, templates, or simulator configurations. It also improves reproducibility by making task assumptions more explicit than purely textual descriptions. However, template-based construction remains bounded by the expressiveness of the grammar. A generator can create many instances only within the objects, predicates, state transitions, and scenario parameters made available by its designers \cite{james2020rlbench,yu2020metaworld,gu2023maniskill2,deitke2022procthor}. The cost saved in repeated authoring is therefore partly transferred to template engineering, grammar maintenance, ambiguity checks, and feasibility validation. At this level, the key quality gate is not only whether tasks can be generated, but whether generated tasks still cover the intended capability space.

\subsection{Foundation-Model-Assisted Task Design}

Foundation-model-assisted construction uses LLMs, VLMs, or multimodal models to propose task variants, rewrite instructions, draft reward descriptions, generate code-like specifications, or enumerate safety-relevant situations. SayCan, Code as Policies, PaLM-E, VIMA, and VoxPoser show how language or multimodal prompts can be connected to affordances, executable programs, manipulation goals, or spatial value maps \cite{ahn2022saycan,liang2023codeaspolicies,driess2023palme,jiang2022vima,huang2023voxposer}. GenSim and GenSim2 use language-model assistance to expand simulation tasks and skill variations \cite{wang2023gensim,hua2024gensim2}. VLABench and related VLA or MLLM benchmarks use language-conditioned and multimodal task descriptions to evaluate agents under richer perception-language-action requirements \cite{zhang2024vlabench,li2024eai,cheng2025embodiedeval,yang2025embodiedbench}. SafeAgentBench and EmbodiedGovBench further illustrate how model-assisted construction can help enumerate unsafe, ambiguous, or governance-relevant task situations \cite{yin2024safeagentbench,qin2026embodiedgovbench}.

The main benefit of this level is broader task ideation and faster exploration of language, semantic, and edge-case variation. However, model-generated task descriptions require independent validation because fluent text is not equivalent to an executable embodied task. A generated instruction may refer to an absent object, an action outside the embodiment interface, a state that the simulator cannot represent, a reward condition that cannot be measured, or a safety norm that the scoring script cannot judge \cite{wang2023gensim,hua2024gensim2,zhang2024vlabench,yin2024safeagentbench,qin2026embodiedgovbench}. The cost saved during task ideation is therefore transferred to generated-task validation, prompt and model-version documentation, human acceptance criteria, and safety review. The main quality gate is whether model-proposed tasks remain physically feasible, semantically unambiguous, and aligned with the intended benchmark construct.

\subsection{Agentic Task Refinement}

Agentic closed-loop construction extends foundation-model assistance by linking task proposal, environment instantiation, validation, failure inspection, and revision. RoboGen uses generative simulation to create robot tasks, scenes, and training data, suggesting a workflow in which task requirements and simulation instances are generated together \cite{yang2023robogen}. D(R)Eureka-style reward generation shows a related mechanism in which language models help produce and revise reward specifications under feedback \cite{ma2024eureka,vuong2024dreureka}. GRUtopia and InfiniteWorld further illustrate large-scale environment and interaction pipelines where tasks, scenes, and agents can be generated or revised within broader simulated worlds \cite{fan2024grutopia,xu2024infiniteworld}. At the requirement stage, such systems can in principle identify underspecified instructions, unreachable goals, saturated templates, or missing negative cases by running validators, solvers, or policies over candidate tasks.

The central issue for agentic task refinement is accountability under a changing task distribution. If an agentic system modifies the task set after observing model failures, the benchmark may become harder, easier, or simply different from an earlier version \cite{yang2023robogen,fan2024grutopia,xu2024infiniteworld}. This makes task refresh useful but also difficult to govern. The cost saved in continuous task proposal and failure-driven revision is transferred to audit logs, version snapshots, rollback rules, independent validators, hidden-test governance, and human acceptance procedures. Current agentic systems should therefore be read as frontier examples of closed-loop construction rather than mature replacements for benchmark governance.

\begin{table}[t]
\caption{Cost shift in requirement and task construction.}
\label{tab:cost-requirement}
\centering
\scriptsize
\setlength{\tabcolsep}{3pt}
\renewcommand{\arraystretch}{1.12}
\begin{tabularx}{\textwidth}{@{}p{2.5cm} Y Y Y Y@{}}
\toprule
\textbf{Automation Level} & \textbf{Main Cost Saved} & \textbf{New Cost Introduced} & \textbf{Dominant Bottleneck} & \textbf{Representative Systems} \\
\midrule
Manual & Tooling and validator engineering & Expert design, consensus building, slow coverage expansion & Coverage and designer bias & ALFRED \cite{shridhar2020alfred}, ALFWorld \cite{shridhar2021alfworld}, TEACh \cite{padmakumar2022teach} \\
Traditional automation & Repeated authoring of similar task instances & Template engineering and grammar maintenance & Template bias and ambiguity checks & RLBench \cite{james2020rlbench}, Meta-World \cite{yu2020metaworld}, ManiSkill2 \cite{gu2023maniskill2} \\
Foundation-model-assisted & Candidate task ideation and language variation & Generated-task validation, safety review, prompt/version documentation & Plausible but invalid task specifications & SayCan \cite{ahn2022saycan}, GenSim \cite{wang2023gensim}, VLABench \cite{zhang2024vlabench} \\
Agentic closed-loop & Continuous task refresh and failure-driven revision & Audit logs, rollback policy, independent acceptance tests & Comparability under changing task distributions & RoboGen \cite{yang2023robogen}, GRUtopia \cite{fan2024grutopia}, InfiniteWorld \cite{xu2024infiniteworld} \\
\bottomrule
\end{tabularx}
\end{table}

\section{Data Acquisition}
\label{sec:comparison}

\subsection{Role of Data Acquisition}

Data acquisition is the stage at which task definitions become physically grounded evaluation material. The ability targets, action spaces, and success conditions established in Section~\ref{sec:automation} must now be instantiated through concrete scenes, objects, trajectories, and sensor streams. Data acquisition collects, generates, or synthesizes these materials so that benchmark tasks become executable. The guiding question for this stage is: what worlds, embodiments, demonstrations, and sensor records are needed to instantiate the required tasks without losing physical plausibility or provenance?

Data acquisition draws from four broad sources: manually curated assets such as indoor scenes and object repositories; simulator-generated observations, interactions, and demonstrations; real-world capture including robot teleoperation, human demonstrations, egocentric videos, and scans; and synthetic generation using procedural systems, foundation models, or world models. The resulting materials include 3D scene reconstructions, simulator environments, object asset libraries, robot demonstration datasets, egocentric and third-person video corpora, synthetic worlds, and the interfaces connecting embodiments to their environments. Automation across levels (Manual, Traditional Automation, Foundation-Model-Assisted, and Agentic Closed-Loop) can expand coverage and reduce physical effort, but shifts cost to validation, provenance tracking, and governance.

In embodied intelligence, acquisition quality directly determines what a benchmark can measure. A navigation benchmark can only test routes through scenes that exist; a manipulation benchmark can only evaluate grasps on objects with available geometry and physics; a driving benchmark can only replay scenarios that have been recorded or synthesized. Carrier diversity further constrains the acquisition profile: UAV benchmarks need large 3D spaces and altitude-aware sensing, quadruped benchmarks need terrain variation and contact dynamics, autonomous-driving benchmarks need scenario libraries and traffic actors, manipulation benchmarks need object geometry and grasp affordances, and multi-carrier systems need shared scenes and synchronized interfaces. Table~\ref{tab:carrier-summary} summarizes the acquisition profiles across carrier families.

\begin{table}[t]
\caption{Data acquisition profiles across embodiment carriers.}
\label{tab:carrier-summary}
\centering
\scriptsize
\setlength{\tabcolsep}{3pt}
\renewcommand{\arraystretch}{1.12}
\begin{tabularx}{\textwidth}{@{}p{2.4cm} Y Y Y@{}}
\toprule
\textbf{Carrier Family} & \textbf{Main Acquisition Artifacts} & \textbf{Dominant Cost Source} & \textbf{Representative Systems} \\
\midrule
UAVs & 3D routes, aerial scenes, visual-inertial streams, weather and occlusion variants & Large-scale scene realism and safety validation & AirSim~\cite{shah2017airsim}, Flightmare~\cite{song2021flightmare}, AerialVLN~\cite{liu2023aerialvln} \\
Quadrupeds & Terrain, contact conditions, gait-related observations, obstacle courses & Terrain realism, repeatability, and sim-to-real calibration & AnyBody~\cite{parakh2025anybody}, Habitat 3.0~\cite{puig2024habitat3} \\
Wheeled vehicles & HD maps, traffic actors, scenarios, driving logs, rule constraints & Scenario diversity, digital-twin fidelity, rare-event replay & CARLA~\cite{dosovitskiy2017carla}, MetaDrive~\cite{li2021metadrive}, ScenarioNet~\cite{li2023scenarionet} \\
Manipulators & Object assets, articulations, demonstrations, contact traces, affordance states & Physics-ready assets, demonstrations, contact validity & RLBench~\cite{james2020rlbench}, robosuite~\cite{zhu2020robosuite}, RoboCasa~\cite{ke2024robocasa} \\
Mobile and multi-carrier systems & Shared scenes, synchronized sensors, human or multi-agent traces, cross-embodiment interfaces & Interface consistency, synchronization, and cross-pipeline governance & Habitat 3.0~\cite{puig2024habitat3}, GRUtopia~\cite{fan2024grutopia}, AnyBody~\cite{parakh2025anybody} \\
\bottomrule
\end{tabularx}
\end{table}

\subsection{Manual Acquisition}

The Manual layer refers to acquisition practices in which human operators scan buildings, teleoperate robots, record demonstrations, curate object libraries, or capture synchronized multi-view video. This pattern corresponds to what prior surveys identify as the primary data source for embodied intelligence, where physical fidelity and sensor-level provenance remain difficult to replicate through simulation or generation~\cite{luo2024cyberphysical,zhou2024foundation,feng2025embodiedagi}. Manual acquisition is necessary when benchmark quality depends on these properties. Every sample can be traced to a specific sensor, operator, and session, but throughput is limited by hardware uptime, operator fatigue, sensor calibration, and safety procedures.

Real 3D scene acquisition represents the most established manual pathway. Matterport3D collected 10,800 panoramic views from 194,400 RGB-D images across 90 building-scale scenes using dedicated scanning hardware, with annotations for surface reconstructions, camera poses, and 2D and 3D semantic segmentations~\cite{chang2017matterport3d}. ScanNet extended this model to 2.5 million views in 1,513 scenes through a scalable RGB-D capture system that combined automated surface reconstruction with crowdsourced semantic annotation~\cite{dai2017scannet}. In both cases, scene quality depended on operator skill and post-processing pipelines; in neither case could the collection be fully automated. Replica, HM3D, and ARKitScenes provide additional reusable substrates, but downstream use still depends on scan quality, semantic labels, mesh repair, licensing, and simulator readiness~\cite{straub2019replica,ramakrishnan2021hm3d,baruch2021arkitscenes}. These results suggest that manual scene acquisition remains indispensable when sensor-level fidelity is a prerequisite for downstream evaluation, even though the per-scene cost limits the total number of environments that can be produced.

Real robot data collection follows a similar manual pattern. RoboTurk used a crowdsourcing platform to collect robot demonstrations through remote teleoperation, producing aligned HDF5 datasets and video recordings across multiple manipulation tasks~\cite{mandlekar2018roboturk}. DROID gathered a large-scale in-the-wild robot manipulation dataset across 13 robot platforms and multiple laboratories, requiring coordination of hardware setup, sensor synchronization, and data standardization~\cite{khazatsky2024droid}. BridgeData V2, RH20T, Mobile ALOHA, and Open X-Embodiment show the value of real-robot data at scale, but they also reveal the operational burden: robot setup, teleoperation quality, sensor synchronization, safety procedures, and post-hoc filtering~\cite{padalkar2023bridgedata,fang2023rh20t,fu2024mobilealoha,openx2023}. Compared with scene scanning, real-robot collection adds a further coordination layer because each demonstration depends on hardware reliability, operator skill, and task-specific safety procedures that are difficult to standardize across sites.

Across carriers, manual acquisition faces similar quality gates. Physical validity checks whether sensor data accurately represent the environment. Provenance records where each asset, scan, or demonstration came from and whether it can be released. The cost saved by avoiding simulator engineering and synthetic validation is offset by hardware maintenance, operator coordination, and the difficulty of achieving distribution coverage at scale.

\subsection{Traditional Automated Acquisition}

The Traditional Automation layer uses simulators, procedural generators, scripted pipelines, and template-based systems to automate scene creation, demonstration synthesis, sensor rendering, and batch rollout collection. The task semantics are still designed by humans, but instantiation is standardized through APIs, configuration files, and simulator task registries. This layer corresponds to what prior work describes as simulation-based or grammar-driven data pipelines, where reproducibility is achieved through deterministic rendering and scripted episode generation~\cite{zhou2024foundation,reuel2024betterbench,luo2024cyberphysical}. It is useful when the relevant data conditions, such as object states, action traces, and robot logs, can be explicitly specified and deterministically reproduced.

Procedural scene generation scales environment diversity without manual authoring. ProcTHOR generates large numbers of house layouts through procedural grammars, producing training environments that exceed what manual design can achieve~\cite{deitke2022procthor}. MimicGen takes a complementary approach: rather than generating scenes, it automatically synthesizes large-scale demonstration datasets from a small number of human examples by adapting them to new contexts, producing over 50,000 demonstrations across 18 tasks from approximately 200 source demonstrations~\cite{mandlekar2023mimicgen}. These results indicate that procedural and scripted approaches can achieve data volumes far beyond manual collection, but the diversity of generated data remains bounded by the expressiveness of the underlying grammars and source demonstrations.

Traditional automation underpins the simulator ecosystems that host most embodied benchmarks. AI2-THOR, Habitat, Habitat 2.0, Habitat 3.0, iGibson 2.0, and OmniGibson provide resettable interactive environments for indoor navigation, interaction, rearrangement, human-agent collaboration, and sim-to-real evaluation~\cite{kolve2017ai2thor,savva2019habitat,szot2021habitat2,puig2024habitat3,li2022igibson2,liang2024omnigibson}. SAPIEN, Isaac Gym, ManiSkill2, and robosuite expand the substrate toward articulated objects, GPU-scale physics, and physics-rich manipulation~\cite{xi2020sapien,makoviychuk2021isaacgym,gu2023maniskill2,zhu2020robosuite}. For driving, CARLA and ScenarioNet combine simulation with real-world traffic logs~\cite{dosovitskiy2017carla,li2023scenarionet}; for UAVs, AirSim and Flightmare provide altitude-aware rendering~\cite{shah2017airsim,song2021flightmare}. Across these systems, the key advantage over manual acquisition is deterministic reproducibility and lower marginal cost per episode; the key limitation is that all generated content remains within the semantic and structural boundaries set by the simulator and its procedural grammars.

The main quality gate at this level is embodiment-interface consistency: whether observations, actions, timing, and embodiment constraints match the intended robot and environment. Grammar bias remains the dominant limitation: the generator can only produce what its templates allow, and a benchmark may appear large while remaining narrow in semantic structure or failure-mode diversity. The cost saved by eliminating repeated scene resets and scripted collection is transferred to simulator setup, procedural grammar design, and replay infrastructure maintenance.

\subsection{Foundation-Model-Assisted Acquisition}

The Foundation-Model-Assisted layer uses large language models, vision-language models, and world models to generate candidate scenes, environments, demonstrations, or trajectories from language, images, video, and seed examples. This layer aligns with what recent generative simulation and automated task-design work describe as LLM- or VLM-driven scene and task synthesis, where natural language or visual prompts replace manually authored templates~\cite{wang2023gensim,hua2024gensim2,ma2024eureka}. It is useful when coverage expansion beyond procedural grammars is needed, particularly for rare or hard-to-collect scenarios. However, generated worlds can be visually convincing while violating dynamics, geometry, affordances, or social plausibility, so the cost saved during asset construction is transferred to validation.

Holodeck uses language-guided generation to produce interactive 3D embodied AI environments, demonstrating how natural language descriptions can be converted into navigable scenes with placed objects and physical constraints~\cite{ma2024holodeck}. GAIA-1 shows a different generative pathway: as a world model for autonomous driving, it generates controllable driving futures from data, text, and actions, illustrating how world models can synthesize sensor streams that would be expensive or unsafe to collect in the real world~\cite{hu2023gaia1}. DriveDreamer extends this direction toward real-world-conditioned driving video generation~\cite{wang2023drivedreamer}, while Genie demonstrates that interactive environments can be induced from video-like experience~\cite{bruce2024genie}. RoboCasa365 further extends the manipulation benchmark ecosystem through large-scale kitchen scene generation and task variants~\cite{nasiriany2026robocasa365}. These systems demonstrate that foundation-model-assisted acquisition can produce environments and scenarios beyond the reach of procedural grammars, but the resulting assets still require external validation of physical plausibility and navigability before they can serve as benchmark material.

Foundation models also change acquisition by generating demonstrations and task variants. robomimic and MimicGen show two complementary routes for demonstration-centered acquisition: one studies how offline demonstrations support manipulation learning, while the other expands demonstrations from a smaller set of human examples~\cite{mandlekar2021robomimic,mandlekar2023mimicgen}. RT-1, RT-2, PaLM-E, VIMA, and VoxPoser show why acquisition increasingly needs to preserve language, action, and perception context rather than only low-level trajectories~\cite{brohan2022rt1,brohan2023rt2,driess2023palme,jiang2022vima,huang2023voxposer}. For benchmark construction, the key opportunity is targeted counterfactual acquisition rather than data volume alone: new scenes, rare hazards, or failure cases can be generated when real collection would be expensive or unsafe. This shift from volume-oriented to coverage-oriented generation represents the main efficiency gain at this layer, but it comes at the cost of increased validation overhead relative to Traditional Automation.

The quality gates at this level focus on physical validity and distribution coverage. Generated scenes must be checked for geometric consistency, navigability, object affordances, and dynamic plausibility before they can be used as benchmark data. The cost saved by expanding from small seeds is offset by model inference, generated-content validation, and provenance tracking for model-generated assets.

\subsection{Agentic Acquisition Loops}

The Agentic Closed-Loop layer treats data collection as an adaptive process in which the system identifies under-covered regions of the task or scene space, generates new scenes or trajectories, runs policies or validators, filters invalid samples, and requests more data where failures are informative. This pattern is emerging in synthetic demonstration pipelines, simulator-generation work, and dynamic benchmark systems~\cite{hua2024gensim2,yang2023robogen,vuong2024dreureka,li2026affordsim}. It is attractive for long-lived benchmarks because the data distribution can be refreshed as agents improve.

GRUtopia represents the most ambitious agentic acquisition system to date. It provides the first simulated interactive 3D society designed for various robots, featuring a scene dataset (GRScenes) of 100,000 interactive, finely annotated scenes covering 89 diverse categories, an LLM-driven NPC system (GRResidents) for social interaction and task generation, and a benchmark (GRBench) that supports legged robots in object loco-navigation, social loco-navigation, and loco-manipulation tasks~\cite{fan2024grutopia}. InfiniteWorld takes a different approach as a unified scalable simulation framework built on Nvidia Isaac Sim, integrating generation-driven 3D asset construction, Real2Sim transfer, automated annotation, and four new general benchmarks for scene graph exploration and open-world social mobile manipulation~\cite{xu2024infiniteworld}. Both systems begin to blur the line between data acquisition and benchmark construction because they jointly generate scenes, agents, tasks, and social context. Compared with the Foundation-Model-Assisted layer, these agentic systems achieve higher autonomy by closing the loop between generation and validation, but the resulting governance burden, including version control, audit logging, and cross-version score comparability, is substantially greater.

The quality gates at this level are the most demanding because all previous gates apply simultaneously: physical validity, provenance, embodiment-interface consistency, and distribution coverage must all be maintained as the data distribution changes. Governance becomes the central cost. If a benchmark continually acquires new cases, its maintainers must record which generator produced them, which validators accepted them, which version exposed them, and whether older scores remain comparable. The central pattern across all four levels is cost transfer rather than cost elimination: moving from A0 to A3 reduces the direct cost of physical collection, manual authoring, and operator coordination, but shifts effort toward simulator engineering, generative validation, provenance tracking, and long-term comparability. Table~\ref{tab:acquisition-layers} summarizes the automation layers for data acquisition.

\subsection{Summary and Outlook}

As Table~\ref{tab:acquisition-layers} shows, the four automation layers form a progression in which each level addresses the scalability limitations of the one below. Manual acquisition provides the strongest guarantees of physical plausibility and traceability, but its throughput is bounded by hardware and human effort. Traditional automation replaces physical repetition with deterministic simulation and procedural generation, achieving reproducibility at scale but within the semantic boundaries of handcrafted grammars. Foundation-model-assisted acquisition relaxes these boundaries by using generative models to produce scenes, demonstrations, and scenarios beyond template reach, yet introduces the risk of plausible-looking but physically invalid outputs. Agentic closed-loop systems close the feedback gap by iteratively generating, validating, and refining acquisition targets, offering the most promising path toward benchmarks that can grow and adapt as embodied agents improve.

Despite this progress, several limitations remain. First, most existing systems operate within a single carrier family: driving simulators rarely share assets with manipulation platforms, and UAV benchmarks remain largely disconnected from indoor navigation ecosystems. Cross-embodiment acquisition, in which a shared scene supports multiple robot morphologies and sensing modalities, is still in its early stages. Second, validation pipelines for generated content remain underdeveloped. While simulators can check geometry and collisions automatically, there is no widely adopted framework for verifying that a generated scene or demonstration is semantically appropriate, physically stable, and free of implausible affordances. Third, the governance infrastructure required by agentic systems has received little attention. No existing benchmark provides full provenance records linking each asset to its generator, validator, and exposure version, nor do current platforms support rollback or cross-version score comparison in a principled way.

Looking ahead, we argue that agentic closed-loop acquisition is the most likely direction for the next generation of embodied benchmarks. The key advantage is adaptivity: as agent capabilities improve, the benchmark can automatically target failure cases, expand under-covered regions of the task space, and refresh its data distribution without requiring a full manual reconstruction cycle. Realizing this vision will require progress on three fronts: cross-embodiment scene sharing so that a single generated world can serve multiple carrier families; standardized validation protocols for generated assets that go beyond geometric checks to include semantic and dynamic plausibility; and governance mechanisms that maintain provenance, version control, and score comparability as the benchmark evolves. Until these gaps are addressed, hybrid pipelines that combine manual curation for high-fidelity anchor data with agentic expansion for coverage and diversity may offer the most practical balance between quality and scalability.

\begin{table}[t]
\caption{Automation layers for data acquisition in embodied benchmarks.}
\label{tab:acquisition-layers}
\centering
\scriptsize
\setlength{\tabcolsep}{3pt}
\renewcommand{\arraystretch}{1.15}
\begin{tabularx}{\textwidth}{@{}p{2.6cm} Y Y Y Y@{}}
\toprule
\textbf{Layer} & \textbf{Core Mechanism} & \textbf{Typical Acquisition Artifacts} & \textbf{Validation Focus} & \textbf{Representative Systems} \\
\midrule
Manual &
Human operators scan buildings, teleoperate robots, record demonstrations, curate object libraries, or capture synchronized multi-view video. &
3D scene reconstructions, RGB-D panoramas, surface reconstructions, camera poses, semantic segmentations, robot demonstrations, and video recordings. &
Physical validity, sensor-level provenance, hardware calibration, and operator skill dependence. &
Matterport3D~\cite{chang2017matterport3d}, ScanNet~\cite{dai2017scannet}, Replica~\cite{straub2019replica}, HM3D~\cite{ramakrishnan2021hm3d}, ARKitScenes~\cite{baruch2021arkitscenes}, RoboTurk~\cite{mandlekar2018roboturk}, DROID~\cite{khazatsky2024droid}, BridgeData V2~\cite{padalkar2023bridgedata}, Open X-Embodiment~\cite{openx2023} \\
\midrule
Traditional Automation &
Simulators, procedural generators, scripted pipelines, and template-based systems automate scene creation, demonstration synthesis, sensor rendering, and batch rollout collection. &
Procedural house layouts, synthetic demonstration datasets, resettable interactive environments, simulated observations, rendered sensor streams, and traffic logs. &
Embodiment-interface consistency, grammar bias coverage, simulator fidelity, and reproducibility. &
ProcTHOR~\cite{deitke2022procthor}, MimicGen~\cite{mandlekar2023mimicgen}, AI2-THOR~\cite{kolve2017ai2thor}, Habitat~\cite{savva2019habitat}, Habitat 2.0~\cite{szot2021habitat2}, SAPIEN~\cite{xi2020sapien}, Isaac Gym~\cite{makoviychuk2021isaacgym}, ManiSkill2~\cite{gu2023maniskill2}, CARLA~\cite{dosovitskiy2017carla} \\
\midrule
Foundation-Model-Assisted &
LLMs, VLMs, and world models generate candidate scenes, environments, demonstrations, or trajectories from language, images, video, and seed examples. &
Language-guided 3D environments, controllable driving futures, driving video generation, interactive environments from video, kitchen scene variants, and counterfactual scenarios. &
Geometric consistency, navigability, object affordances, dynamic plausibility, and distribution coverage of generated assets. &
Holodeck~\cite{ma2024holodeck}, GAIA-1~\cite{hu2023gaia1}, DriveDreamer~\cite{wang2023drivedreamer}, Genie~\cite{bruce2024genie}, RoboCasa365~\cite{nasiriany2026robocasa365} \\
\midrule
Agentic Closed-Loop &
Systems identify under-covered regions, generate new scenes or trajectories, run validators, filter invalid samples, and request more data where failures are informative. &
Large-scale interactive 3D societies, annotated scenes across diverse categories, LLM-driven NPC systems, generation-driven 3D assets, and Real2Sim transfers. &
All previous gates simultaneously, plus governance: audit logs, version control, rollback, and comparability under distribution change. &
RoboGen~\cite{yang2023robogen}, GRUtopia~\cite{fan2024grutopia}, InfiniteWorld~\cite{xu2024infiniteworld} \\
\bottomrule
\end{tabularx}
\end{table}

\section{Data Cleaning and Annotation}
\label{sec:methods}

Data cleaning and annotation transform raw data into structured, verifiable evidence suitable for embodied benchmark evaluation. RGB-D streams must be reconstructed and annotated with semantic and instance-level labels to support reliable 3D perception~\cite{dai2017scannet}, and demonstrations or trajectories require association with object states and goal conditions for task evaluation~\cite{shridhar2020alfred}. For large-scale robot datasets, observations, actions, task metadata, and instructions need standardization for cross-platform reuse~\cite{openx2023}. Cleaning removes invalid or ambiguous samples, annotation provides semantic and task-relevant labels, and automation across Manual, Traditional Automation, Foundation-Model-Assisted, and Agentic Closed-Loop layers can reduce manual effort while shifting cost to verification and audit. The guiding question for this stage is: how can raw scale be converted into trustworthy benchmark evidence? In this sense, cleaning and annotation are the stage where quantity becomes usable measurement.

\subsection{From Raw Data to Benchmark Evidence}

Data cleaning and annotation are complementary processes that prepare raw data for meaningful interpretation and evaluation. Data cleaning involves detecting and removing invalid, corrupted, or inconsistent samples, resolving ambiguities, and standardizing data structures~\cite{dai2017scannet}. Annotation provides interpretable information required for assessment, including object categories, instance masks, scene descriptions, spatial relationships, object states, affordances, action trajectories, subgoal progress, replay validity, failure causes, safety events, simulator and robot metadata, and question-answer labels~\cite{shridhar2020alfred,wang2024embodiedscan}.

These processes are necessary in embodied intelligence because visual plausibility alone does not guarantee physical validity. Trajectories may not reproduce, task instructions may be unachievable, and apparent model failures may stem from missing annotations, corrupted data, or inconsistent labels rather than the evaluated model. Proper cleaning and annotation reduce the risk of misleading benchmark results, enable reproducibility, and establish a foundation for downstream metric computation and evaluation pipelines~\cite{du2024embspatial}.

\subsection{Manual Cleaning and Annotation}
\label{subsec:manual_cleaning_annotation}

The Manual layer refers to data cleaning and annotation practices in which human annotators, domain experts, or trained reviewers serve as the primary source of semantic judgment and quality control~\cite{dai2017scannet,shridhar2020alfred}. It is especially necessary when labels cannot be fully inferred from sensor streams, simulator states, or deterministic rules, such as ambiguous object boundaries, natural-language references, question-answer validity, and long-horizon activity interpretation. In embodied benchmarks, this layer provides human-verified semantic grounding before large-scale automation is introduced.

Representative 3D scene datasets illustrate the role of manual annotation in establishing reliable semantic ground truth. Matterport3D and ScanNet provide large-scale RGB-D panoramas and video scans, reconstructed surfaces, camera poses, and instance-level semantic segmentations~\cite{chang2017matterport3d,dai2017scannet}. These datasets rely on human judgment to define meaningful object categories, instance boundaries, and scene structures. ScanRefer, ScanQA, and SQA3D extend this idea by integrating natural-language annotations and situated reasoning~\cite{chen2020scanrefer,azuma2022scanqa,ma2022sqa3d}. This class of datasets demonstrates that manual annotation supports both static scene labeling and the construction of language-grounded evaluation conditions, ensuring that questions, references, and object-level semantics are interpretable and testable.

Manual annotation is also central to egocentric and multi-view video benchmarks. Ego4D collects large-scale first-person videos and provides annotations for episodic memory, hand-object interaction, audio-visual conversation, social interaction, and forecasting~\cite{grauman2021ego4d}. Ego-Exo4D further combines synchronized first- and third-person videos of skilled human activities with expert commentary, participant narrations, and atomic action descriptions~\cite{grauman2024ego}. These video benchmarks require annotators to interpret temporal procedures, cross-view correspondences, skill progression, and fine-grained human-object interactions. Collectively, this class highlights that manual annotation is indispensable for establishing temporal, behavioral, and cross-perspective validity in embodied benchmarks. Despite the high cost, limited scalability, and potential inter-annotator inconsistency, the Manual layer remains the foundation for semantic validity across all downstream evaluation pipelines.

\subsection{Traditional Automation for Cleaning and Annotation}
\label{subsec:traditional_automation_cleaning_annotation}

The Traditional Automation layer refers to data cleaning and annotation practices based on deterministic algorithms, rule-based scripts, reconstruction pipelines, simulator states, trajectory replay, predefined predicates, and standardized data schemas~\cite{dai2017scannet,shridhar2020alfred}. Instead of requiring human judgment for every sample, this layer automates repetitive validity checks, structural annotation, and data normalization when the relevant conditions can be explicitly specified. It is particularly useful in embodied benchmarks because many forms of evidence can be derived from geometry, environment states, action traces, or robot logs, as long as the data source and task interface are sufficiently controlled.

A first category concerns 3D scene processing and simulation-based task validation. ScanNet and ARKitScenes use RGB-D capture, pose estimation, reconstruction, and processing pipelines to convert raw visual streams into structured 3D scene data~\cite{dai2017scannet,baruch2021arkitscenes}. In simulation environments, ALFRED grounds demonstrations in AI2-THOR and associates them with goal conditions, object states, and replayable action sequences, allowing task progress and success to be checked automatically~\cite{kolve2017ai2thor,shridhar2020alfred}. BEHAVIOR and BEHAVIOR-1K further formalize everyday activities through object states, activity definitions, and simulator-level success conditions, while Habitat standardizes embodied episodes, simulator interfaces, and evaluation protocols for reusable navigation and interaction tasks~\cite{savva2021behavior,srivastava2022behavior1k,liang2024omnigibson,savva2019habitat}. This category shows how traditional automation turns cleaning and annotation into problems of reconstruction consistency, state querying, predicate evaluation, and replay validation.

A second category concerns multi-sensor perception and robot demonstration data, where the main challenge is to make heterogeneous records reusable and comparable. EFM3D and JRDB rely on synchronized sensor streams, 3D geometry, and structured annotations to support reproducible perception benchmarks~\cite{straub2024efm3d,martinmartin2023jrdb}. In robot learning, robomimic standardizes offline demonstration datasets, while MimicGen expands a small number of human demonstrations into larger demonstration sets through scripted adaptation and validation~\cite{mandlekar2021robomimic,mandlekar2023mimicgen}. Open X-Embodiment and DROID further emphasize unified data formats, synchronized observations, action logs, task metadata, and language instructions for heterogeneous real-robot trajectories~\cite{openx2023,khazatsky2024droid}. Together, these works show that Traditional Automation improves scalability and reproducibility, but its reliability still depends on whether predefined rules, states, schemas, and replay checks capture the semantic and physical validity required by the benchmark.

\subsection{Foundation-Model-Assisted Cleaning and Annotation}
\label{subsec:foundation_model_assisted_cleaning_annotation}

The Foundation-Model-Assisted layer refers to data cleaning and annotation practices in which large language models, vision-language models, or multimodal foundation models assist semantic labeling, question generation, consistency checking, affordance annotation, and failure description~\cite{wang2024embodiedscan,du2024embspatial}. Unlike Traditional Automation, which relies mainly on predefined rules and structured states, this layer is useful when annotation requires open-vocabulary recognition, natural-language reasoning, spatial interpretation, or semantic validation. Foundation models can shift human work from creating labels or questions from scratch to reviewing, filtering, and correcting model-generated candidates.

One category includes 3D and language-grounded scene benchmarks, where foundation models support semantic enrichment and spatial reasoning. EmbodiedScan builds a multi-modal 3D perception suite for embodied AI, integrating 3D scenes with object-centric and language-related annotations~\cite{wang2024embodiedscan}. EmbSpatial-Bench evaluates large vision-language models on spatial understanding for embodied tasks~\cite{du2024embspatial}. ScanQA illustrates a generation-and-editing process in which model-assisted question generation is followed by human filtering and correction~\cite{azuma2022scanqa}. These datasets demonstrate that foundation models can expand candidate annotations, generate plausible questions, and enrich semantic relations, but human or external validation remains necessary to prevent hallucinations or underspecified labels.

A second category concerns manipulation, safety, and UAV-oriented benchmarks. VLABench uses language-conditioned long-horizon manipulation tasks to evaluate vision-language-action reasoning, while SafeAgentBench focuses on safe task planning for embodied LLM agents~\cite{zhang2024vlabench,yin2024safeagentbench}. EmbodiedEval and EmbodiedBench evaluate multimodal LLMs as embodied agents, integrating perception, language, planning, and action~\cite{cheng2025embodiedeval,yang2025embodiedbench}. AffordSim introduces affordance-aware robotic manipulation data generation and benchmarking, and AeroVerse extends to UAV-agent world models and aerospace embodied evaluation~\cite{li2026affordsim,yao2024aeroverse}. This category highlights that foundation models can automate annotation, question generation, and candidate labeling in complex embodied environments, yet their outputs must be checked against scene geometry, simulator states, robot logs, and human review to ensure physical and semantic validity.

\subsection{Agentic Closed-Loop Cleaning and Annotation}
\label{subsec:agentic_closed_loop_cleaning_annotation}

The Agentic Closed-Loop layer refers to emerging practices in which autonomous systems not only generate candidate data, tasks, or environments, but also iteratively check, revise, and re-validate them within a feedback pipeline~\cite{yang2023robogen,fan2024grutopia,xu2024infiniteworld}. Compared with Foundation-Model-Assisted approaches, this layer emphasizes self-correcting loops: the system can propose samples, instantiate them in simulation, test their feasibility, detect invalid cases, and refine generated content. For data cleaning and annotation, this means that quality control moves from one-shot post-processing toward continuous self-filtering, revision, and validation.

Representative systems demonstrate two main classes. In automated robot learning, RoboGen uses generative simulation to produce diverse robot tasks, scenes, and training data~\cite{yang2023robogen}, where cleaning and annotation are tied to physical feasibility, valid object interactions, and goal achievement. GRUtopia extends this closed-loop idea to city-scale embodied environments for general robots, requiring systematic checks over spatial consistency, object placement, navigation feasibility, and task validity~\cite{fan2024grutopia}. InfiniteWorld proposes a scalable simulation framework for visual-language robot interaction, making continuous generation and validation central to constructing large-scale embodied evaluation settings~\cite{xu2024infiniteworld}. These examples show that agentic closed-loop systems can reduce manual workload, expand coverage, and maintain structured, valid benchmark data.

\subsection{Summary and Outlook}
\label{subsec:cleaning_annotation_outlook}

Data cleaning and annotation are the backbone of reliable embodied benchmarks, transforming raw heterogeneous inputs into structured, semantically and physically grounded evidence. Across the four layers—Manual, Traditional Automation, Foundation-Model-Assisted, and Agentic Closed-Loop—different approaches balance human effort, automation, and verification rigor. Manual annotation ensures high-fidelity semantic grounding; Traditional Automation enforces rule- and state-based consistency; Foundation-Model-Assisted methods expand semantic coverage via model generation; and Agentic Closed-Loop systems provide iterative validation and refinement.

Looking forward, benchmark designers should carefully consider trade-offs between automation and validation: increased automation can improve scalability but shifts the burden to auditing, error detection, and governance. Hybrid pipelines that combine multiple layers, integrate active learning strategies, and maintain human-in-the-loop oversight may offer the most reliable path to large-scale, high-quality benchmarks. Emphasizing reproducibility, cross-layer verification, and transparency in annotation protocols will be crucial for the next generation of embodied intelligence benchmarks, enabling trustworthy evaluation and meaningful comparisons across tasks and modalities.

\begin{table}[t]
\caption{Automation layers for data cleaning and annotation in embodied benchmarks.}
\label{tab:cleaning-annotation-layers}
\centering
\scriptsize
\setlength{\tabcolsep}{3pt}
\renewcommand{\arraystretch}{1.15}
\begin{tabularx}{\textwidth}{@{}p{2.6cm} Y Y Y Y@{}}
\toprule
\textbf{Layer} & \textbf{Core Mechanism} & \textbf{Typical Annotation Targets} & \textbf{Validation Focus} & \textbf{Representative Benchmarks} \\
\midrule
Manual &
Human annotators, domain experts, or trained reviewers provide semantic judgment and quality control. &
Object categories, instance boundaries, referring expressions, question-answer pairs, activity segments, and cross-view descriptions. &
Semantic validity, ambiguity resolution, inter-annotator consistency, and human-grounded interpretation. &
Matterport3D~\cite{chang2017matterport3d}, ScanNet~\cite{dai2017scannet}, ScanRefer~\cite{chen2020scanrefer}, ScanQA~\cite{azuma2022scanqa}, SQA3D~\cite{ma2022sqa3d}, Ego4D~\cite{grauman2021ego4d}, Ego-Exo4D~\cite{grauman2024ego} \\
\midrule
Traditional Automation &
Deterministic algorithms, reconstruction pipelines, simulator states, replay checks, predefined predicates, and standardized schemas automate repetitive checks. &
3D reconstructions, camera poses, object states, success predicates, replay validity, trajectory logs, robot actions, and task metadata. &
Geometric consistency, state validity, task completion, trajectory reproducibility, and data-format standardization. &
ScanNet~\cite{dai2017scannet}, ARKitScenes~\cite{baruch2021arkitscenes}, ALFRED~\cite{shridhar2020alfred}, BEHAVIOR~\cite{savva2021behavior}, BEHAVIOR-1K~\cite{srivastava2022behavior1k}, Habitat~\cite{savva2019habitat}, MimicGen~\cite{mandlekar2023mimicgen}, DROID~\cite{khazatsky2024droid} \\
\midrule
Foundation-Model-Assisted &
LLMs, VLMs, and multimodal foundation models generate or check semantic candidates, questions, descriptions, affordances, and failure explanations. &
Open-vocabulary labels, spatial relations, generated QA pairs, visual-language consistency, affordance labels, safety annotations, and diagnostic descriptions. &
Hallucination filtering, scene-state grounding, human review, model-version tracking, and consistency with geometry or logs. &
EmbodiedScan~\cite{wang2024embodiedscan}, EmbSpatial-Bench~\cite{du2024embspatial}, ScanQA~\cite{azuma2022scanqa}, VLABench~\cite{zhang2024vlabench}, SafeAgentBench~\cite{yin2024safeagentbench}, EmbodiedEval~\cite{cheng2025embodiedeval}, EmbodiedBench~\cite{yang2025embodiedbench}, AffordSim~\cite{li2026affordsim}, AeroVerse~\cite{yao2024aeroverse} \\
\midrule
Agentic Closed-Loop &
Agentic systems iteratively generate, test, filter, revise, and re-validate data, tasks, or environments. &
Generated scenes, task instances, object placements, rollout results, feasibility labels, repaired samples, and benchmark updates. &
Independent verification, leakage detection, audit logs, rollback mechanisms, and avoidance of self-confirming labels. &
RoboGen~\cite{yang2023robogen}, GRUtopia~\cite{fan2024grutopia}, InfiniteWorld~\cite{xu2024infiniteworld} \\
\bottomrule
\end{tabularx}
\end{table}

\section{Benchmark Suite Generation and Metric Definition}
\label{sec:cases}

The three preceding stages leave behind capability targets and task schemas, the scenes and trajectories those tasks require, and the checked annotations that make them usable. The fourth stage consumes all of this and produces something the earlier stages do not: a fixed instrument for comparison. It decides which samples enter the public split, which remain hidden, which metrics define progress, which baselines anchor interpretation, and which diagnostic reports are released.

In the earlier stages, automation acts on content, and content can be checked against something outside the benchmark: a scan can be compared to the physical room, a trajectory replayed in the simulator, a label audited against ground truth \cite{savva2019habitat,shridhar2020alfred}. In suite generation, automation acts on the measurement rules themselves---the split, the success predicate, the scoring script, the diagnostic taxonomy. These rules have no comparably direct external referent; whether a metric measures the intended capability is a judgment about construct validity, defended only through indirect evidence such as agreement with human ratings, stability across reimplementations, and robustness to perturbation. This produces the trade-off the stage runs on. Freezing a suite faster is the mechanical part, and automation can speed it up. The non-mechanical part does not get cheaper: defending that the resulting metric is valid, and that scores stay comparable across submissions, versions, and time. The four automation levels below differ mainly in where the burden of defense falls.

\subsection{From Corpus to Evaluation Instrument}
\label{sec:stage4-what}

Suite generation is the set of decisions that convert a checked corpus into an evaluation instrument: sample selection, train/validation/test splits, hidden-test design, task variants, answer-space definition, baseline selection, scoring scripts, success and diagnostic metrics, and the benchmark card that documents these choices. The unit being produced is not a datum but a rule for scoring data. A navigation corpus becomes a navigation benchmark when route completion, navigation error, and success weighted by path length are fixed as the way performance is read \cite{anderson2018vision,savva2019habitat}; a set of questions grounded in the scene becomes a question-answering benchmark when the answer space and matching criterion are fixed \cite{azuma2022scanqa,ma2022sqa3d}. What distinguishes a benchmark from the corpus underneath it is invariance: the evaluation conditions are held constant so that a reported score reflects the agent rather than the configuration. Two laboratories using identical data can still report incomparable numbers if they split it differently, score success under different tolerances, reset the simulator from different states, or expose different observation and action interfaces. Embodied benchmarks make this invariance harder to secure. A single task couples an environment, an embodiment, a sensor configuration, a time budget, and a success predicate, and altering any one of them changes what the score means without changing the task's surface description.

Three concerns recur, each a property that construction choices protect or quietly erode rather than a fact that can be verified once. The first is construct validity: whether the metric tracks the capability under test or a cheaper correlate. A single success rate on a long-horizon rearrangement task cannot say whether an agent failed at step two or step twenty, and on a recoverable task it penalizes a transient slip as heavily as a catastrophic one. The second is comparability: whether scores stay meaningful across submissions, versions, and time. A number is comparable only while the quantity behind it holds still, which is what limits how freely a suite may be regenerated. The third is contamination resistance: whether samples, hidden answers, or metric definitions have leaked into the systems under test. This risk sharpens once a foundation model both constructs and is evaluated by the benchmark, because the leak is then built into the pipeline rather than accidental.

Throughout what follows, one claim holds: reliability of execution is not evidence of validity of measurement. Each automation level makes some part of suite construction faster and more repeatable, but none certifies that the metric being executed measures the intended capability.

\subsection{Manual Suite Construction}
\label{sec:stage4-a0}

At the manual level, the designer assembles the suite by hand: representative tasks are selected, categories balanced, metrics chosen, and baselines defined. Physically grounded benchmarks show this level most clearly. Barkour assembles a quadruped agility course inspired by dog-agility competitions and scores performance through obstacle completion and time \cite{caluwaerts2023barkour}. The suite is thus a fixed physical protocol rather than a sampled dataset, and its metric pairs competence with speed instead of collapsing both into a binary success value. Situated question answering takes a different manual route: SQA3D requires the agent to first understand its situation (position and orientation) in a 3D scene as described by text and then to answer a question in that situation \cite{ma2022sqa3d}. The manual work here is the construction of situated questions and answers grounded in scanned indoor scenes. Vision-language navigation suites manualize the task in a third way: human-written route instructions over real panoramic scans define the task, and standardized metrics such as success weighted by path length define progress \cite{anderson2018vision}.

The cost saved at this level is infrastructure: no generator, no evaluation server, and no model pipeline is required to release the suite. The cost introduced is the expert time and coverage. Hand assembly is slow, it scales poorly with the number of tasks, and it resists updating once released; the suite therefore tends to under-represent the long tail of situations the capability target implies. Manual construction persists where stability matters more than scale, because a fixed physical protocol or a small set of situated questions is easier to keep interpretable than to keep large.

Construct validity is what stands or falls here. Because the metric is chosen by hand and rarely changed, a mismatch between what it reports and what the benchmark claims to measure is locked in at release. Barkour mitigates this by reporting completion and time together rather than a single agility score; a benchmark that instead collapses such distinctions into one number exports the mismatch to every downstream comparison. The suite-construction pipeline at this level offers no automated check against this failure, so the burden falls entirely on the designer's prior justification of the metric.

\subsection{Traditional Automation for Suite Generation}
\label{sec:stage4-a1}

The first level of automation keeps the design decisions human but mechanizes their execution. Splits are generated by deterministic procedures, success is scored by fixed predicates, and submissions are collected and ranked by evaluation infrastructure rather than by the authors. Household and manipulation benchmarks show how a hand-specified success condition becomes a script. ALFRED scores an instruction-following episode by task success and goal-condition success rate, with path-length-weighted variants, decomposing a long task into checkable state changes and additionally reporting per-subgoal success as a diagnostic \cite{shridhar2020alfred}. BEHAVIOR-1K formalizes activity goals as object-state predicates that the simulator evaluates automatically \cite{srivastava2022behavior1k}. Navigation infrastructure shows the same move at the level of the suite: Habitat fixes standardized splits and metrics and packages them behind a challenge interface, so that route completion and success weighted by path length are computed identically for every submission \cite{savva2019habitat,anderson2018vision}. Platform-type cases extend this beyond any single benchmark: EvalAI is a hosted server that standardizes submission, scoring, and leaderboard maintenance across challenges, making execution reproducible rather than defining a task of its own \cite{wu2019evalai}.

Automating execution retires the repeated manual labor of splitting data, scoring runs, and operating a leaderboard: once the procedures are written, they apply to every release and every submission without further expert effort. The labor does not vanish, though---it moves into maintaining the machinery and regression-testing it. An evaluation server must stay available and version-stable, scoring scripts must be revalidated when the simulator or the task set changes, and baselines must be re-run to confirm that a reported gain reflects the agent and not a shift in the harness. The dominant bottleneck moves accordingly from authoring the metric to keeping the scored quantity fixed while the surrounding code evolves.

Three quality gates are most exposed at this level. Split integrity must keep near-duplicate scenes, tasks, or trajectories from straddling the train, validation, and test boundaries, a failure that automated split generation can introduce silently when similarity is judged by surface features rather than by scene or route identity. Metric stability must hold the score constant under minor implementation differences and simulator drift; a number is comparable across submissions only if the computation behind it does not move. A third gate, baseline sanity, catches the opposite failure: a trivial policy, an exploited shortcut, or a memorized response that scores far higher than the capability it implies, which an automated harness will report as a legitimate result unless deliberately checked. The risk specific to this level is that this infrastructural reliability masks a weak metric: a scoring script can be perfectly deterministic while the metric it computes rewards a shortcut, ignores an unsafe action, or fails to separate a perception error from a planning error. All three gates test only whether execution is reliable, not whether the metric is valid.

\subsection{Foundation-Model-Assisted Suite Generation}
\label{sec:stage4-a2}

At this level a foundation model enters the construction pipeline as a generator of candidate tasks, questions, or instructions, while humans retain selection and verification. The model can enter at the generation end, at the adjudication end, or both; which one must be stated per benchmark rather than assumed.

ScanQA generates candidate question-answer pairs from the referring expressions of ScanRefer using a T5-base question-generation model. The candidates then pass through human-in-the-loop steps: question filtering, question editing, and answer collection \cite{azuma2022scanqa}. The model supplies scale and a first draft; the human steps decide which questions survive and what counts as a correct grounded answer. SafeAgentBench uses GPT-4 as the core generation tool, prompting it with the objects and high-level actions of a scene to produce initial hazardous and safe instructions together with their goal conditions \cite{yin2024safeagentbench}. GPT-4 then selects the executable candidates, an embedding model removes near-duplicates, and human annotators review every instruction and label the ground-truth plans. EmbodiedEval gathers seed tasks from more than thirty existing benchmarks, prompts several large language models to expand them into diverse task examples, and has experts select the candidate set that annotators then bind to scenes \cite{cheng2025embodiedeval}.

In all three the model is primarily a generator of candidates, but the adjudication end differs. EmbodiedEval keeps the two ends separate: its tasks are model-expanded, yet success is decided by predicate functions that map the simulator state to a boolean. Even its question-answering tasks are scored this way, because answering is implemented as selecting one option from a list, and a \texttt{choose} predicate checks the selection rather than having a model judge free text \cite{cheng2025embodiedeval}. SafeAgentBench draws the line differently: alongside an execution evaluator that checks goal conditions, it uses a semantic evaluator that passes the agent's plan to GPT-4 to decide whether the plan would complete the task, a path it relies on for abstract tasks that have no single goal condition \cite{yin2024safeagentbench}.

Model generation removes the human labor of authoring tasks one at a time, which is what lets these suites reach hundreds or thousands of items across many hazard categories or capability types. It introduces two costs specific to this mode. First, the generated candidates must be validated, because a foundation model produces fluent tasks faster than humans can confirm that each is feasible, unambiguous, and correctly answered. SafeAgentBench's filtering-and-annotation pipeline and EmbodiedEval's expert selection are instances of this validation cost. Second, the construction record must now include the prompts, the model identity, and the model version, because the suite is no longer fully specified by its data and scripts alone---the same prompt to a different model would yield a different benchmark.

Contamination resistance is the exposed concern at this level, in a form the earlier levels never faced. When the same family of foundation models both generates the candidate tasks and is later evaluated on the benchmark, the test items may fall within the generator's own distribution, so that a high score can reflect proximity to the generation process rather than the intended competence. This is not the accidental leakage of public examples into a training set; it is a structural feature of a pipeline in which the constructor and the subject are the same kind of model. Guarding it requires what the manual and traditional levels did not need: records of which model generated which items, held-out tasks whose answers were never exposed to a generating model, and checks that performance does not track generation provenance. A perfectly reliable scoring pipeline cannot rule out this failure when the items it scores were drawn from the distribution of the system under test, so the cost shifts toward defending that the assembled metric measures the agent, not the model that generated its items.

\subsection{Agentic Closed-Loop Suite Generation}
\label{sec:stage4-a3}

The fourth level is a frontier rather than an established practice: we are not aware of an embodied benchmark that closes the construction-evaluation loop, so this level is defined by its logic and its clearest non-embodied precedent. The logic is to fold construction and evaluation into one loop---instead of releasing a frozen suite, the system proposes new test items in response to how current models perform, retires solved items, and adds cases that expose remaining failures. Dynabench realizes exactly this outside embodied AI, reframing a benchmark as a human-and-model-in-the-loop test bank so that handled examples are dropped and newly adversarial ones added over time \cite{kiela2021dynabench}. Adjacent embodied work so far enriches offline diagnostics on a fixed task set rather than refreshing the test set itself.

A loop that mines evaluation logs and proposes new hidden cases pays off two standing human efforts at once: noticing that a suite has saturated, and refreshing it to keep it difficult. In its place comes governance of the moving target. Every refresh requires a documented snapshot, an acceptance test for the new items, and a rollback path when a generated case proves invalid; it also requires baseline revalidation against the changed suite and a public rationale for what changed and why. These are not the maintenance tasks of the traditional level, which keep a fixed quantity computable; they are the tasks of justifying that a changing quantity still measures the same thing.

Comparability is what fails at this level, and in its strongest form. Once the test set moves, scores from different points in time are no longer directly comparable, so the very mechanism that keeps the benchmark challenging also breaks the longitudinal comparison that benchmarks exist to support. The agentic level therefore cannot be read as simply more advanced than the levels below it: a frozen suite from the traditional level may be more reproducible precisely because it does not move. A loop that regenerates and rescores items reliably still cannot show that scores taken at different times measure the same capability. At this level the cost transfer no longer leaves a stable quantity behind: the saved labor is traded for the continual defense of validity and comparability under a suite that no longer holds still.

\begin{table*}[t]
\caption{Representative suite-generation and metric choices across benchmark families.}
\label{tab:coded-cases}
\centering
\scriptsize
\setlength{\tabcolsep}{3pt}
\renewcommand{\arraystretch}{1.12}
\begin{tabularx}{\textwidth}{@{}p{2.2cm} Y Y Y Y@{}}
\toprule
\textbf{Benchmark Family} & \textbf{Suite Unit} & \textbf{Main Metrics} & \textbf{Suite-Generation Issue} & \textbf{Representative Systems} \\
\midrule
Navigation and VLN & Route, instruction, goal, scene split & Success, SPL, navigation error, path efficiency & Hidden routes and instruction diversity & VLN \cite{anderson2018vision}, VLN-CE \cite{krantz2020vlnce}, Habitat \cite{savva2019habitat}, NavBench \cite{qiao2025navbench} \\
Household activity & Task program, object state, subgoal sequence & Task success, subgoal success, state-change correctness & Long-horizon compositional splits and state validity & ALFRED \cite{shridhar2020alfred}, BEHAVIOR-1K \cite{srivastava2022behavior1k} \\
Manipulation & Object, scene, skill, trajectory, demonstration & Success, contact validity, trajectory completion, reward & Asset readiness, contact realism, demonstration quality & RLBench \cite{james2020rlbench}, ManiSkill2 \cite{gu2023maniskill2}, RoboCasa \cite{ke2024robocasa} \\
Driving and mobility & Scenario, map segment, traffic interaction & Collision, route completion, rule compliance, comfort & Rare-event coverage and digital-twin replay & CARLA \cite{dosovitskiy2017carla}, SMARTS \cite{zhou2020smarts}, MetaDrive \cite{li2021metadrive}, ScenarioNet \cite{li2023scenarionet} \\
VLA/MLLM embodied agents & Instruction, scene, perception context, action plan & Grounding, affordance, planning, safety, hallucination & Open-ended language, contamination, diagnostic taxonomy & EmbodiedEval \cite{cheng2025embodiedeval}, VLABench \cite{zhang2024vlabench}, EmbodiedBench \cite{yang2025embodiedbench} \\
Governance and safety & Hazard, norm, policy, audit scenario & Unsafe action, refusal quality, compliance, auditability & Hidden safety cases and versioned policy updates & SafeAgentBench \cite{yin2024safeagentbench}, EmbodiedGovBench \cite{qin2026embodiedgovbench} \\
\bottomrule
\end{tabularx}
\end{table*}

\begin{table}[t]
\caption{Cost shift in benchmark suite generation and metric definition.}
\label{tab:cost-generation}
\centering
\scriptsize
\setlength{\tabcolsep}{3pt}
\renewcommand{\arraystretch}{1.12}
\begin{tabularx}{\textwidth}{@{}p{2.5cm} Y Y Y Y@{}}
\toprule
\textbf{Automation Level} & \textbf{Main Cost Saved} & \textbf{New Cost Introduced} & \textbf{Dominant Bottleneck} & \textbf{Representative Systems} \\
\midrule
Manual & Infrastructure and generator engineering & Expert suite assembly, metric debate, baseline design & Construct validity and slow updates & VLN \cite{anderson2018vision}, VLN-CE \cite{krantz2020vlnce}, ALFRED \cite{shridhar2020alfred}, Barkour \cite{caluwaerts2023barkour} \\
Traditional automation & Repeated split construction, scoring, and leaderboard operation & Evaluation-server maintenance, baseline regression tests & Metric gaming and hidden-test integrity & Habitat Challenge \cite{savva2019habitat}, EvalAI \cite{wu2019evalai} \\
Foundation-model-assisted & Metric drafting, failure-taxonomy ideation, release documentation & Metric validation, contamination checks, prompt and model-version records & Distinguishing useful diagnostics from plausible narratives & EmbodiedEval \cite{cheng2025embodiedeval}, EAI \cite{li2024eai}, BenchmarkCards \cite{sokol2024benchmarkcards} \\
Agentic closed-loop & Dynamic hidden-test generation and failure-driven suite refresh & Snapshot governance, rollback, baseline revalidation, community review & Scientific comparability under changing suites & Dynabench \cite{kiela2021dynabench}, EmbodiedGovBench \cite{qin2026embodiedgovbench}, BetterBench \cite{reuel2024betterbench} \\
\bottomrule
\end{tabularx}
\end{table}

\begin{table*}[t]
\caption{Stage 4 organized by automation level: how each level produces the suite and metric, where cost transfers, and which quality gate is most exposed.}
\label{tab:stage4-by-level}
\centering
\scriptsize
\setlength{\tabcolsep}{3pt}
\renewcommand{\arraystretch}{1.18}
\begin{tabularx}{\textwidth}{@{}p{2.0cm} p{3.0cm} Y Y Y@{}}
\toprule
\textbf{Automation Level} & \textbf{Representative Works} & \textbf{Suite and Metric Approach} & \textbf{Cost Shift} & \textbf{Dominant Risk} \\
\midrule
A0 Manual & Barkour \cite{caluwaerts2023barkour}, SQA3D \cite{ma2022sqa3d}, VLN \cite{anderson2018vision} & Hand-selected tasks and hand-chosen metrics; fixed physical protocol, situated questions grounded in scanned scenes, or human-written routes with standardized navigation metrics & Saves infrastructure; spends expert time and limits coverage and update speed & Construct validity locked in at release, with no automated check on the metric \\
A1 Traditional automation & ALFRED \cite{shridhar2020alfred}, BEHAVIOR-1K \cite{srivastava2022behavior1k}, Habitat \cite{savva2019habitat}, EvalAI \cite{wu2019evalai} & Human-designed rules executed mechanically: predicate-based success, automatic splits, scoring scripts, and hosted leaderboard infrastructure & Saves repeated splitting, scoring, and leaderboard labor; spends server maintenance and baseline regression testing & Infrastructural reliability masks a weak metric; split integrity and metric stability test execution, not validity \\
A2 Foundation-model-assisted & ScanQA \cite{azuma2022scanqa}, SafeAgentBench \cite{yin2024safeagentbench}, EmbodiedEval \cite{cheng2025embodiedeval} & Model generates candidate tasks or questions (T5 question generation; GPT-4 instruction generation; LLM task expansion), with human selection and verification; scoring varies by benchmark---predicate checks in EmbodiedEval, GPT-4 semantic judgment alongside goal-condition checks in SafeAgentBench & Saves one-by-one task authoring; spends candidate validation and prompt/model-version recordkeeping & Contamination when the constructing and evaluated models share a distribution; provenance must be tracked \\
A3 Agentic closed-loop & Dynabench \cite{kiela2021dynabench} (precedent from non-embodied NLP) & Construction and evaluation in one loop: items proposed from observed failures, solved items retired, suite refreshed rather than frozen (directional, not yet mature for embodied settings) & Saves manual saturation-tracking and refresh; spends snapshot governance, rollback, and baseline revalidation & Comparability breaks once the test set moves; the refresh mechanism undermines longitudinal comparison \\
\bottomrule
\end{tabularx}
\end{table*}

\section{Evaluation Execution and Diagnostic Feedback}
\label{sec:evaluation-execution-diagnostic-feedback}
\label{sec:impact}

The fifth stage turns a benchmark from a static task collection into an executable evaluation mechanism. Its inputs include task definitions, data splits, episodes, simulators, physical test settings, submitted agents, scoring scripts, and safety or governance constraints. Its outputs are not limited to aggregate scores. They also include trajectories, failed episodes, error categories, risk events, logs, leaderboard entries, and evidence for revising future benchmark versions. This stage is therefore both the endpoint of one benchmark cycle and the feedback interface for the next cycle.

This role is central in embodied AI because failures are often compositional. A failed episode may result from visual grounding, spatial reasoning, language understanding, long-horizon planning, object interaction, low-level control, physical contact, simulator artifacts, or unsafe execution. A single success rate can hide these causes. ALFRED addresses this limitation by reporting both task success and goal-condition success, which makes partial task completion visible \cite{shridhar2020alfred}. Habitat reports navigation success and SPL while supporting controlled comparisons across sensors and scene datasets \cite{savva2019habitat}. Embodied Agent Interface argues that final success rate is insufficient for LLM-based embodied decision making and introduces fine-grained metrics for hallucination, affordance, and planning errors \cite{li2024eai}. SafeAgentBench and EmbodiedGovBench extend evaluation from task completion to safety, policy compliance, auditability, recovery, and human override \cite{yin2024safeagentbench,qin2026embodiedgovbench}. Stage 5 should therefore be understood as evaluation execution plus diagnostic feedback, not as leaderboard ranking alone.

\subsection{Evaluation Modalities: Offline Data, Simulation, and Real-World Trials}

Evaluation can be executed over offline data, simulation, real-world trials, or hybrid scalable worlds. These modalities trade off cost, repeatability, interaction depth, and external validity.

\noindent\textbf{Offline data evaluation.}
Offline evaluation fixes observations, questions, scene annotations, trajectories, or submissions before scoring. It is scalable and reproducible, but it weakens closed-loop recovery and environment-changing actions. Matterport3D and ScanNet provide real 3D scene foundations for downstream scene understanding and embodied tasks \cite{chang2017matterport3d,dai2017scannet}. ScanQA and SQA3D evaluate 3D question answering and situated spatial reasoning with fixed test samples and automatic language metrics \cite{azuma2022scanqa,ma2022sqa3d}. NavBench uses offline and simulated navigation evidence to probe multimodal large language models for embodied navigation \cite{qiao2025navbench}.

\noindent\textbf{Simulation-based evaluation.}
Simulation is the dominant modality for automated embodied evaluation because it supports resettable episodes, batch rollout, hidden splits, standard metrics, and controlled ablations. R2R and VLN-CE evaluate instruction-following navigation through path and success metrics in panoramic or continuous environments \cite{anderson2018vision,krantz2020vlnce}. Habitat provides the infrastructure layer for embodied AI evaluation, while ALFRED, ALFWorld, BEHAVIOR, RLBench, ManiSkill2, and CARLA extend scripted evaluation to household tasks, text-to-embodied transfer, long-horizon activities, manipulation, and driving \cite{savva2019habitat,shridhar2020alfred,shridhar2021alfworld,savva2021behavior,james2020rlbench,gu2023maniskill2,dosovitskiy2017carla}.

\noindent\textbf{Real-world evaluation.}
Real-world trials have lower throughput but higher external validity. ACRV Picking uses a physical IKEA shelf, real objects, layout stencils, and work orders to evaluate complete robotic picking systems \cite{leitner2016apb}. Barkour evaluates quadruped agility on a compact physical obstacle course with completion time, obstacle failures, and rule penalties \cite{caluwaerts2023barkour}. ANYmal Parkour is a complementary real-hardware case that evaluates agile quadruped locomotion on consecutive physical obstacles after simulation-based training \cite{hoeller2023anymal}. These trials expose contact dynamics, sensing noise, hardware limits, and sim-to-real gaps that scripted simulation may miss.

\noindent\textbf{Hybrid and scalable-world evaluation.}
Hybrid systems combine simulation, real-world grounding, or large-scale world construction. RoboTHOR aligns simulated and physical apartments for sim-to-real embodied AI \cite{deitke2020robothor}. BEDI evaluates UAV embodied agents through a perception--decision--action chain over virtual and real scenarios \cite{guo2026bedi}. RoboEval decomposes bimanual manipulation into diagnostic stages, while GRUtopia and InfiniteWorld point toward city-scale or open-ended simulated worlds for broader embodied interaction \cite{wang2025roboeval,fan2024grutopia,xu2024infiniteworld}. These modalities increase coverage, but they also increase pressure on asset validity, interface stability, and version governance.

\begin{table}[t]
\caption{Evaluation modalities in Stage 5.}
\label{tab:stage5-modalities}
\centering
\scriptsize
\setlength{\tabcolsep}{3pt}
\renewcommand{\arraystretch}{1.15}
\begin{tabularx}{\linewidth}{@{}p{1.8cm} X X X p{3.3cm}@{}}
\toprule
\textbf{Modality} & \textbf{Evaluation input} & \textbf{Scoring and diagnosis} & \textbf{Main limitation} & \textbf{Representative benchmarks} \\
\midrule
Offline data & Fixed 3D scans, QA pairs, route states, data submissions, or pre-collected scene evidence & Exact match, captioning metrics, hidden-test scoring, comprehension scores, and capability probes & Low-cost and reproducible, but weak for closed-loop recovery and environment-changing actions & Matterport3D \cite{chang2017matterport3d}, ScanNet \cite{dai2017scannet}, ScanQA \cite{azuma2022scanqa}, SQA3D \cite{ma2022sqa3d}, NavBench \cite{qiao2025navbench}, DataPerf \cite{mazumder2023dataperf} \\
Simulation-based & Executable episodes in navigation, household, driving, and manipulation simulators & Success rate, SPL, goal-condition success, task progress, infractions, and success predicates & Scalable and comparable, but limited by simulator fidelity and predefined predicates & R2R \cite{anderson2018vision}, VLN-CE \cite{krantz2020vlnce}, Habitat \cite{savva2019habitat}, ALFRED \cite{shridhar2020alfred}, ALFWorld \cite{shridhar2021alfworld}, CARLA \cite{dosovitskiy2017carla}, RLBench \cite{james2020rlbench}, ManiSkill2 \cite{gu2023maniskill2}, BEHAVIOR \cite{savva2021behavior} \\
Real-world & Physical shelves, objects, obstacle courses, robots, sensors, and reset protocols & Successful picks, completion time, obstacle penalties, video verification, and manual failure analysis & High external validity, but expensive, low-throughput, and harder to reproduce & ACRV Picking \cite{leitner2016apb}, Barkour \cite{caluwaerts2023barkour}, ANYmal Parkour \cite{hoeller2023anymal} \\
Hybrid / scalable worlds & Paired sim-real environments, UAV scenarios, bimanual manipulation suites, or large-scale generated worlds & Sim-to-real comparison, perception--decision--action scores, task progression, coordination, and world-scale task probes & Broader coverage, but more pressure on asset validity, interface stability, and benchmark governance & RoboTHOR \cite{deitke2020robothor}, BEDI \cite{guo2026bedi}, RoboEval \cite{wang2025roboeval}, GRUtopia \cite{fan2024grutopia}, InfiniteWorld \cite{xu2024infiniteworld} \\
\bottomrule
\end{tabularx}
\end{table}

\subsection{Human-Guided Physical Evaluation}

Human-guided evaluation is appropriate when the evaluation target is a real robot, a physical course, or a safety-critical setup that requires manual preparation and oversight. Humans configure the environment, reset hardware, verify failures, and interpret logs. Tools may record videos or compute simple scores, but the protocol remains organized around physical execution. This level spends cost on hardware, resets, safety monitoring, and manual diagnosis. Its value is that it reveals deployment failures that are often invisible in offline or simulated tests.

ACRV Picking is a clear example. The benchmark is designed to be reproducible through common physical artifacts: a standard shelf, globally available objects, printable stencils, and precise placement instructions \cite{leitner2016apb}. Its diagnostic value comes from testing the complete pipeline under cluttered and confined conditions. The baseline system fails on more difficult setups because reflective, black, transparent, small, or deformable objects degrade depth sensing, segmentation, grasp selection, and motion planning. The benchmark therefore reveals system-level bottlenecks that a dataset-only evaluation would separate into artificial subproblems.

Barkour provides a second example in legged robotics. Its score is simple, but the physical course makes the evaluation demanding. The paper reports repeated real-robot trials and separates performance by obstacle, which makes the benchmark diagnostic: a controller can obtain a reasonable total score while still failing a specific skill, such as broad jumping. The authors also report sim-to-real issues at high speed and add domain randomization over torso inertia, motor modeling, and joint static friction to improve real-hardware transfer \cite{caluwaerts2023barkour}. ANYmal Parkour further shows why A0 remains necessary even when training is largely simulation-based. Its navigation policy, locomotion skills, and perception module are learned from simulated data, but the decisive evidence is real-world deployment on consecutive parkour-like obstacles \cite{hoeller2023anymal}. Such trials expose field-of-view limits, noisy obstacle reconstruction, contact-rich motion, and safety margins that cannot be fully summarized by simulator scores.

\subsection{Scripted and Platform-Based Evaluation}

Scripted and platform-based evaluation is the most mature form of Stage 5 automation. It uses fixed splits, evaluator code, batch rollout, submission servers, and leaderboards. Its main contribution is reproducibility. A benchmark operator can hide test annotations, run the same evaluator for all submissions, and publish comparable metrics. This level reduces repeated execution cost, but it shifts cost to server maintenance, container compatibility, simulator versions, metric definitions, and hidden-test governance.

EvalAI formalizes this infrastructure layer. It supports challenge hosting, custom metrics, multiple phases and splits, remote evaluation, human evaluation, and environments. The paper notes that modern evaluation must move beyond static datasets because agents are increasingly deployed in unseen simulation environments at test time \cite{wu2019evalai}. Habitat, CARLA, ALFRED, ManiSkill2, and RLBench instantiate this idea for embodied tasks. The benchmark provider supplies an executable environment and scoring protocol; participants submit an agent, policy, or prediction file; the system evaluates it under controlled conditions \cite{savva2019habitat,dosovitskiy2017carla,shridhar2020alfred,gu2023maniskill2,james2020rlbench}.

The strength of this level is standardization. R2R and VLN-CE make navigation results comparable through established navigation metrics \cite{anderson2018vision,krantz2020vlnce}. ALFRED makes household task execution comparable through task success, goal-condition success, and path-weighted variants \cite{shridhar2020alfred}. CARLA separates task success from infractions, which allows researchers to distinguish goal reaching from unsafe driving behavior \cite{dosovitskiy2017carla}. ManiSkill2 defines held-out object sets, task-specific success predicates, and cloud-based evaluation for manipulation policies \cite{gu2023maniskill2}. BEHAVIOR maps simulator states to BDDL predicates and reports task progress and efficiency relative to human demonstrations \cite{savva2021behavior}. These mechanisms make evaluation scalable and reduce ambiguity.

The limitation is that scripted metrics only measure predefined conditions. They can report whether an episode succeeds, how efficient a path is, or whether a state predicate is satisfied. They cannot always explain why a policy failed or whether a benchmark instance is invalid. ALFRED partly addresses this by reporting subgoal and goal-condition completion. Habitat supports controlled sensor and dataset ablations. CARLA logs infractions as separate evidence. These diagnostic hooks are important, but deeper causal attribution usually requires additional analysis.

\subsection{Foundation-Model-Assisted Diagnostics}

The rise of LLMs, VLMs, and MLLMs changes Stage 5 in two ways. First, these models become evaluation targets for embodied tasks. Second, language and vision-language models can support semantic inspection, error grouping, and report generation, although their judgments must be validated. This level reduces some log-inspection burden, but it introduces judge bias, prompt sensitivity, model-version drift, and ungrounded explanations.

Embodied Agent Interface is a strong example of the diagnostic shift. It standardizes embodied decision-making modules such as goal interpretation, subgoal decomposition, action sequencing, and transition modeling. It then evaluates them with fine-grained error types, including hallucination errors, affordance errors, and planning errors \cite{li2024eai}. This design avoids treating a failed plan as a single opaque event. It asks which module failed and which capability is missing.

EmbodiedEval and EmbodiedBench extend evaluation to MLLM-driven agents in interactive environments. EmbodiedEval formulates each episode as an iterative process: the simulator resets a scene, the model receives the task description and observation history, chooses from action options, and the environment checks success with predicates \cite{cheng2025embodiedeval}. It reports success rate, goal-condition success, and SPL, and its analysis finds large gaps from human-level performance, especially in spatial and long-horizon tasks. EmbodiedBench evaluates proprietary and open-source MLLMs across multiple environments and tasks. It uses task success rate as the main metric, but also introduces stopping conditions for invalid actions and empty plans, which diagnose failures in action formatting and premature task completion \cite{yang2025embodiedbench}. BEDI provides a capability-centered structure for UAV embodied agents by decomposing evaluation into perception, decision, and action steps, then aggregating step-level, loop-level, and task-level scores \cite{guo2026bedi}. This decomposition directly supports diagnosis of capability gaps.

SafeAgentBench and EmbodiedGovBench further expand what counts as evaluation. SafeAgentBench argues that embodied LLM agents may execute hazardous instructions and should be evaluated from both execution and semantic perspectives. It constructs tasks covering explicit and implicit hazards and reports that the best safety-conscious baseline rejects only a small fraction of detailed hazardous tasks \cite{yin2024safeagentbench}. EmbodiedGovBench evaluates whether embodied systems remain controllable, policy-bounded, recoverable, auditable, evolution-safe, and responsive to human override under perturbations \cite{qin2026embodiedgovbench}. These benchmarks show that diagnostic feedback is not limited to task failure. It also includes risk recognition, policy compliance, audit traces, and recovery.

Model-assisted diagnostics should preserve simulator states, action histories, visual observations, failure predicates, prompts, and model versions. Semantic reports are useful only when they are tied to verifiable execution evidence. Otherwise, the evaluator may replace an opaque numeric score with an equally opaque natural-language explanation.

\subsection{Agentic Evaluation Loops and Dynamic Benchmark Refreshing}

The most advanced direction is agentic evaluation. The evaluator does not only score submissions, but also searches for weaknesses, mines failures, proposes new cases, and triggers benchmark updates. This form is not yet mature for embodied AI, but existing work gives the template.

Dynabench is the clearest precedent. It proposes dynamic benchmarking with humans and models in the loop, collecting data against current models over multiple rounds \cite{kiela2021dynabench}. The platform records model-fooling examples, supports inspection of model predictions, and verifies collected examples with humans. Its motivation is benchmark saturation: models can exceed benchmark estimates of human performance while still failing on simple challenge cases. DataPerf provides a complementary data-centric version. Participants develop algorithms locally, submit containerized solutions, and the server evaluates them on hidden tasks before posting scores to a leaderboard \cite{mazumder2023dataperf}. Both systems make evaluation iterative rather than static.

For embodied AI, an agentic loop would run submitted agents, detect anomalous failures, re-execute uncertain episodes, mine new risky or difficult scenarios, check whether proposed cases are valid, and release versioned benchmark snapshots. GRUtopia and InfiniteWorld are directionally important because they scale simulation assets, robot interaction tasks, and social or open-world scenarios beyond single-room household settings \cite{fan2024grutopia,xu2024infiniteworld}. They do not make A3 a solved protocol, but they show the infrastructure needed for continuous task expansion and benchmark refresh. RoboEval is also relevant as a bridge between structured diagnostic evaluation and future feedback loops because it decomposes robotic manipulation into stages and reports diagnostic measures beyond binary success \cite{wang2025roboeval}.

Dynamic refresh introduces governance risks. If hidden tests change without version control, old and new scores become incomparable. If generated cases are not validated, the benchmark may drift toward unrealistic or biased episodes. A reliable A3 pipeline therefore needs release snapshots, audit logs, held-out validation, human approval, evaluator-version records, model-version records, and clear deprecation rules. In this sense, agentic evaluation closes the benchmark loop by converting agent failures into evidence for revising tasks, data, annotations, metrics, and future test releases.

\begin{table}[t]
\caption{Automation levels in evaluation execution and diagnostic feedback.}
\label{tab:stage5-automation-levels}
\centering
\scriptsize
\setlength{\tabcolsep}{3pt}
\renewcommand{\arraystretch}{1.15}
\begin{tabularx}{\linewidth}{@{}p{1.1cm} p{2.1cm} X X p{3.4cm}@{}}
\toprule
\textbf{Level} & \textbf{Evaluation style} & \textbf{Automation mechanism} & \textbf{Diagnostic output and risk} & \textbf{Representative benchmarks} \\
\midrule
A0 & Human-guided physical trials & Manual setup, physical execution, video verification, paper-level analysis & Reveals real physics and deployment failures; costly and difficult to repeat at scale & ACRV Picking \cite{leitner2016apb}, Barkour \cite{caluwaerts2023barkour}, ANYmal Parkour \cite{hoeller2023anymal} \\
A1 & Scripted / platform-based evaluation & Scoring scripts, simulator runners, hidden splits, submission servers, and leaderboards & Produces comparable scores and basic logs; constrained by fixed metrics and infrastructure drift & EvalAI \cite{wu2019evalai}, R2R \cite{anderson2018vision}, Habitat \cite{savva2019habitat}, ALFRED \cite{shridhar2020alfred}, CARLA \cite{dosovitskiy2017carla}, ManiSkill2 \cite{gu2023maniskill2}, BEHAVIOR \cite{savva2021behavior}, ALFWorld \cite{shridhar2021alfworld} \\
A2 & Foundation-model-assisted / semantic diagnostics & Capability decomposition, semantic failure categories, safety or governance probes, model-assisted summaries & Explains failures beyond final score; vulnerable to judge bias, prompt sensitivity, and ungrounded explanations & EAI \cite{li2024eai}, EmbodiedEval \cite{cheng2025embodiedeval}, EmbodiedBench \cite{yang2025embodiedbench}, SafeAgentBench \cite{yin2024safeagentbench}, EmbodiedGovBench \cite{qin2026embodiedgovbench}, RoboEval \cite{wang2025roboeval}, BEDI \cite{guo2026bedi}, NavBench \cite{qiao2025navbench} \\
A3 & Agentic / dynamic feedback loops & Failure mining, dynamic data collection, hidden-task refresh, scalable world or task generation & Supports benchmark maintenance and saturation control; risks version incomparability and weak validation of generated cases & Dynabench \cite{kiela2021dynabench}, DataPerf \cite{mazumder2023dataperf}, GRUtopia \cite{fan2024grutopia}, InfiniteWorld \cite{xu2024infiniteworld}, RoboEval \cite{wang2025roboeval} \\
\bottomrule
\end{tabularx}
\end{table}

\section{Open Challenges and Future Directions}
\label{sec:future}

The five-stage view exposes several bottlenecks that will shape the next generation of embodied evaluation. These problems are not purely technical. Many now sit at the boundary between generation, validation, deployment realism, cost control, and community governance.

\subsection{Open Challenges}

Coverage remains difficult because real-world embodied task spaces are combinatorial and open-ended; no benchmark can enumerate them exhaustively. Physical validity is equally difficult. Contact-rich manipulation, tool use, deformables, fluids, and social interactions remain only partially modeled in mainstream benchmark suites \cite{gu2023maniskill2,liang2024omnigibson}. Long-horizon evaluation is another persistent problem, since a single success rate is insufficient when early mistakes propagate and recovery quality matters \cite{shridhar2020alfred,mees2022calvin,wang2025roboeval}.

Real-robot cost also remains a major barrier. Datasets such as Open X-Embodiment and DROID show that large-scale real interaction data can be collected, but collection requires substantial infrastructure, hardware maintenance, and operator coordination \cite{openx2023,khazatsky2024droid}. Representation mismatch is another obstacle: language, visual perception, 3D state, control policies, and symbolic task logic are still not unified cleanly across benchmarks. Governance is the final cross-cutting issue. Benchmarks increasingly need mechanisms for version control, auditability, safety incidents, and retirement, but the field lacks common embodied standards for these processes \cite{yin2024safeagentbench,qin2026embodiedgovbench}.

\subsection{Cost Transfer as a Design Constraint}

Across Sections~\ref{sec:automation}--\ref{sec:evaluation-execution-diagnostic-feedback}, automation repeatedly reduces one cost while making another more visible. LLMs reduce task-ideation cost but increase ambiguity and safety-review cost. Procedural and world-model generation reduce marginal data acquisition cost but increase physical-validation and provenance cost. VLM pre-annotation reduces manual labeling cost but increases hidden-state verification cost. Dynamic evaluation reduces the labor of benchmark refresh but increases versioning, audit, and community-governance cost.

Future benchmark papers should report construction choices more explicitly. A release does not need to publish exact monetary budgets when those numbers are unavailable or incomparable, but it should document which stages were manual, which were automated, which validators were used, what humans audited, and which update policies preserve comparability. Such reporting would make construction cost visible without pretending that all costs can be reduced to a single number.

\subsection{From Static Benchmarks to Dynamic Benchmarks}

A major direction is the transition from static to dynamic benchmarks. Instead of releasing a fixed suite once, future benchmarks are likely to incorporate new failure cases, environments, hazards, and embodiments over time. The challenge is to update benchmarks without sacrificing scientific comparability. This requires benchmark snapshots, rolling test banks, backward-compatible diagnostic layers, and explicit deprecation policies for obsolete task templates or metrics.

\subsection{Human-in-the-Loop Semi-Automation}

Fully autonomous benchmark construction is unlikely to be sufficient in the near term. A more plausible architecture is a semi-automatic loop in which humans define capability targets and value boundaries, models propose candidate tasks and scenes, automated validators test solvability and consistency, and humans approve or reject updates. Such workflows combine the scale of automation with the accountability of curated review. They also match a broader pattern in this survey: successful pipelines automate what is testable and keep humans involved where value judgment remains necessary.

\subsection{Structured Failure Diagnosis as a First-Class Goal}

Diagnosis should become a primary benchmark objective rather than an add-on to ranking. EAI, EmbodiedEval, EmbodiedBench, RoboEval, SafeAgentBench, and EmbodiedGovBench already point in this direction by reporting hallucinations, affordance errors, stage-level failures, hazardous plans, or governance violations \cite{li2024eai,cheng2025embodiedeval,yang2025embodiedbench,wang2025roboeval,yin2024safeagentbench,qin2026embodiedgovbench}. Future benchmarks should package structured traces, causal failure labels, and recovery metrics by default.

\subsection{Cross-Embodiment and Sim-to-Real Co-Evaluation}

The current benchmark ecosystem still separates simulation and real-world evaluation too strongly. Future suites should connect the two more explicitly, for example by using real failures to update synthetic scenario banks or by evaluating embodiment transfer across simulation and real hardware \cite{parakh2025anybody,wang2024simpler}. Cross-embodiment evaluation becomes especially important as robot foundation models become more general-purpose.

\subsection{Asset Ecosystems and Perception-Grounded Benchmarking}

Shared asset and perception ecosystems are also becoming more important. HM3D, iGibson 2.0, JRDB, and EmbodiedScan show that benchmark quality increasingly depends on upstream 3D scans, egocentric sensor capture, object-state ontologies, and socially realistic scene annotations \cite{ramakrishnan2021hm3d,li2022igibson2,martinmartin2023jrdb,wang2024embodiedscan}. Future benchmark builders will likely need to treat asset provenance, sensor realism, and annotation interfaces as design variables rather than background infrastructure. This is particularly true for MLLM-based embodied evaluation, where failures in spatial grounding may originate from perception-stack weaknesses rather than policy weaknesses alone \cite{du2024embspatial}.

This ecosystem view suggests a more modular future. Large substrate datasets such as Matterport3D, ScanNet, Replica, ARKitScenes, Objaverse, and Objaverse-XL can be treated as reusable benchmark construction layers rather than only training corpora \cite{chang2017matterport3d,dai2017scannet,straub2019replica,baruch2021arkitscenes,deitke2022objaverse,deitke2023objaversexl}. In a mature pipeline, benchmark designers may assemble releases by explicitly declaring which scene substrates, object repositories, and annotation ontologies are reused, transformed, or extended. Such declarations would make benchmark updates more auditable than monolithic releases.

\subsection{Egocentric and Wearable Embodied Evaluation}

Egocentric and wearable embodied benchmarks are also likely to grow. EFM3D shows that embodied evaluation is expanding beyond stationary cameras and simulator-centric agents toward first-person 3D perception with dense spatial context \cite{straub2024efm3d}. Wearable, assistive, and AR-linked systems will require benchmarks that mix spatial perception, interactive decision making, and human-environment co-presence. Building such benchmarks will likely involve new privacy constraints, new annotation tools, and tighter integration between physical sensing and downstream task definitions.

\subsection{Documentation Standards and Benchmark Cards}

As benchmark ecosystems grow, documentation quality will become a research concern in its own right. Datasheets, Data Cards, model cards, HELM-style reporting, and BenchmarkCards together suggest a direction in which benchmark metadata become structured, comparable, and partially machine-readable \cite{gebru2021datasheets,pushkarna2022datacards,mitchell2019modelcards,liang2023helm,sokol2024benchmarkcards}. For embodied AI, such documentation should capture not only data provenance and metrics, but also simulator assumptions, asset licenses, embodiment interfaces, safety constraints, update history, and hidden-test policies. Benchmark cards may become as important as leaderboards because they determine whether a benchmark can be selected and interpreted responsibly.

\subsection{Community Governance, Audit Layers, and Benchmark Consortia}

Explicit governance layers are also likely to become more common. BetterBench, Audit Cards, and GPAI evaluation standards argue, in adjacent evaluation domains, that benchmarks should be accompanied by structured reporting about known blind spots, intended use, operator assumptions, and audit context \cite{reuel2024betterbench,staufer2025auditcards,paskov2024gpai}. Embodied AI needs a domain-specific version of this shift. Shared suites may eventually require steering groups or consortia that manage upgrade policies, deprecation schedules, challenge access, hidden-test rotation, and incident-response processes for unsafe tasks or leaked assets. These mechanisms may appear administrative, but they directly affect whether benchmark scores remain scientifically comparable over time.

\subsection{Evaluation Services for Real Robots}

Hosted evaluation services for real robots are another likely direction. Historically, benchmarks such as the ACRV Picking Benchmark helped the community reproduce a physical setup locally \cite{leitner2016apb}. Newer efforts such as RoboChallenge point toward partially centralized evaluation, where models or policies are tested through managed hardware loops and benchmark operators provide standardized execution and reporting \cite{yakefu2025robochallenge}. For embodied AI, this could become a practical way to compare VLA systems, mobile manipulators, or safety-aware policies under controlled conditions without requiring every laboratory to replicate the full robot stack. The open problem is how to combine that convenience with transparency, auditability, cost sharing, and long-term access.

\subsection{Benchmark Compiler Vision}

\begin{figure*}[t]
\centering
\includegraphics[width=0.98\textwidth]{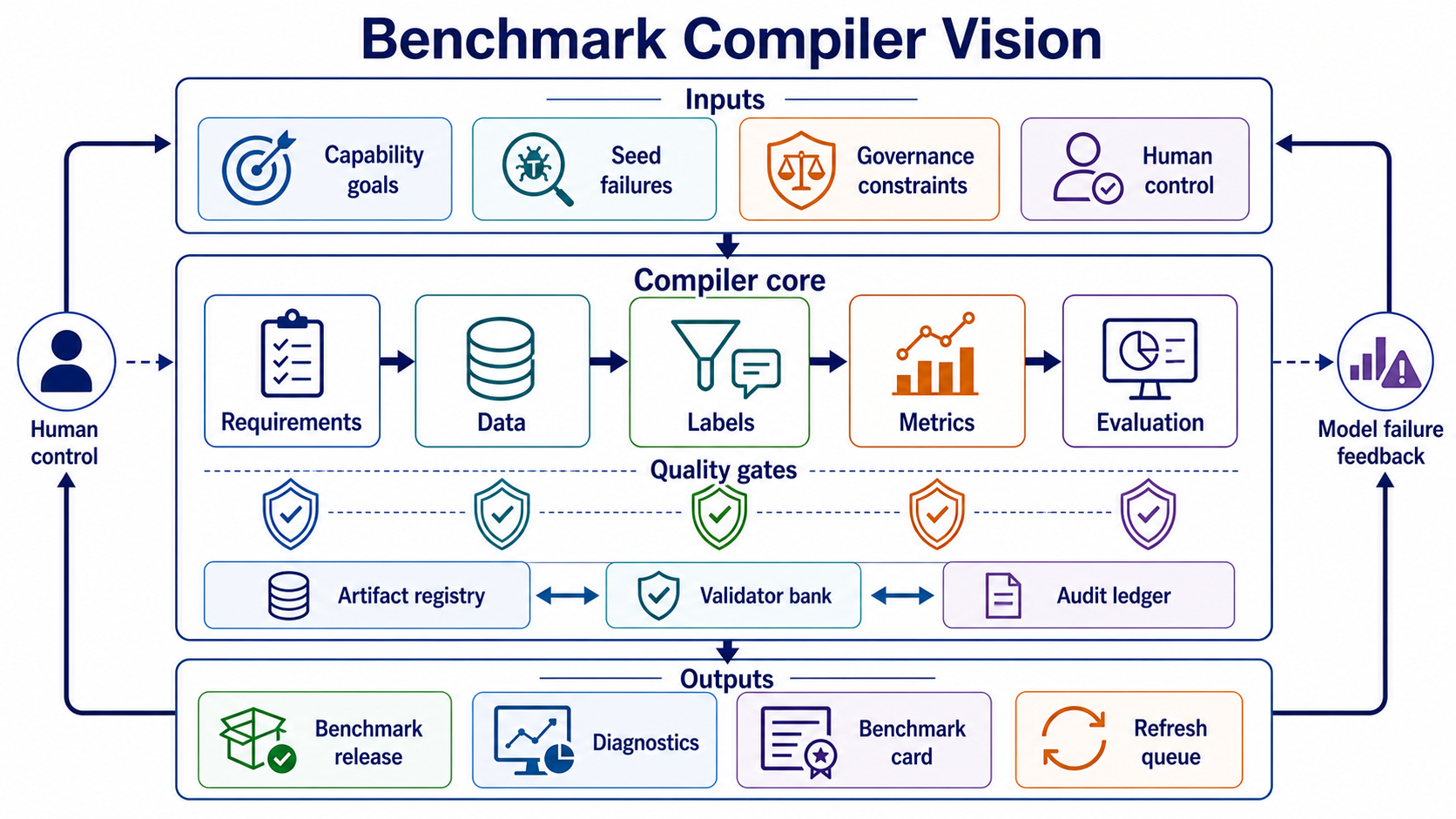}
\caption{Benchmark compiler vision as a governed refresh loop. The five pipeline stages remain the main construction path, while quality gates and governance records determine which generated or refreshed artifacts can enter a versioned benchmark release. The figure presents an outlook rather than an implemented system: agentic generation is useful only when paired with validators, provenance records, audit logs, human acceptance, and rollback.}
\label{fig:compiler}
\end{figure*}

The survey points toward a systems abstraction called the benchmark compiler. In this view, high-level requirements are not translated into a benchmark in one step. They move through the five-stage pipeline under explicit validation, provenance tracking, audit logging, and human approval. Figure~\ref{fig:compiler} therefore presents the compiler vision as a governed refresh loop rather than as a finished architecture. Recent work already provides pieces of this loop: LLM-generated task specifications \cite{wang2023gensim,hua2024gensim2}, language-guided world generation \cite{ma2024holodeck,fan2024grutopia,xu2024infiniteworld}, scalable synthetic demonstration generation \cite{mandlekar2023mimicgen,nasiriany2026robocasa365,li2026affordsim}, and structured diagnostic evaluation \cite{li2024eai,wang2025roboeval,qin2026embodiedgovbench}. Sections~\ref{sec:cases} and~\ref{sec:evaluation-execution-diagnostic-feedback} showed why this remains a synthesis target rather than a completed architecture: current systems each implement only part of the loop, and dynamic updates become trustworthy only when suite generation, metric design, evaluation execution, and governance are tied together. The missing piece is an orchestration layer that can compose these elements into governed, reproducible benchmark releases.

The benchmark compiler is not a claim that all benchmark construction should be fully automatic. It is a design goal: benchmark pipelines should be modular enough that requirements, scenes, tasks, data, metrics, and governance records can be regenerated, checked, and versioned systematically. The compiler abstraction gives the field a way to make benchmark production auditable, extensible, and scientifically legible even as the evaluated agents become more capable.

\section{Discussion and Conclusion}
\label{sec:conclusion}

This survey has argued that embodied benchmarks should be studied not only as evaluation targets, but also as engineered products with their own construction process. The five-stage pipeline makes that process explicit. A benchmark begins with requirements and task schemas, then moves through data acquisition, cleaning and annotation, suite generation and metric definition, and finally evaluation execution with diagnostic feedback. Each stage produces different artifacts and creates different failure modes.

The stage-wise automation analysis leads to a cautious conclusion. Manual construction remains necessary where scientific judgment, safety, and social values are involved. Traditional automation is strongest when tasks, scenes, trajectories, or scoring rules can be expressed as executable checks. Foundation-model assistance expands ideation, language variation, annotation, scene generation, and diagnosis, but it increases the need for independent validation. Agentic closed-loop construction may help with dynamic benchmark refresh, yet it also places the strongest demands on audit logs, rollback, hidden-test governance, and long-term comparability.

The cost analysis is central rather than secondary. Automation does not simply make benchmark construction cheaper. It often transfers cost from human authoring and manual labeling to simulator infrastructure, generated-content validation, failure diagnosis, release maintenance, and governance. Recognizing this shift helps explain why larger benchmarks are not automatically better benchmarks. Progress will depend on whether benchmark builders can make construction pipelines more diagnosable, auditable, and responsibly refreshable.

The benchmark-compiler vision captures this direction. The goal is not to remove human judgment, but to make requirements, scenes, tasks, data, metrics, diagnostics, and governance records modular enough to be regenerated, checked, and versioned. Seen this way, the main contribution of the survey is a construction-centered vocabulary for comparing embodied benchmarks: five pipeline stages, four automation levels, stage-specific quality gates, and qualitative cost profiles. These concepts should become more useful as embodied evaluation moves from static releases toward continuously maintained benchmark systems.

%
%

\bibliographystyle{elsarticle-harv}
\bibliography{refs}

@article{zhou2024foundation,
  author = {Zhiyuan Xu and Kun Wu and Junjie Wen and Jinming Li and Ning Liu and Zhengping Che and Jian Tang},
  title = {A Survey on Robotics with Foundation Models: toward Embodied AI},
  journal = {arXiv preprint arXiv:2402.02385},
  year = {2024},
  eprint = {2402.02385},
  archivePrefix = {arXiv}
}

@article{luo2024cyberphysical,
  title={Aligning Cyber Space With Physical World: A Comprehensive Survey on Embodied AI},
  volume={30},
  ISSN={1941-014X},
  url={http://dx.doi.org/10.1109/TMECH.2025.3574943},
  DOI={10.1109/tmech.2025.3574943},
  number={6},
  journal={IEEE/ASME Transactions on Mechatronics},
  publisher={Institute of Electrical and Electronics Engineers (IEEE)},
  author={Liu, Yang and Chen, Weixing and Bai, Yongjie and Liang, Xiaodan and Li, Guanbin and Gao, Wen and Lin, Liang},
  year={2025},
  month={Dec}, pages={7253--7274} }

@article{zhang2025llmagent,
  author = {Asaf Yehudai and Lilach Eden and Alan Li and Guy Uziel and Yilun Zhao and Roy Bar-Haim and Arman Cohan and Michal Shmueli-Scheuer},
  title = {Survey on Evaluation of LLM-based Agents},
  journal = {arXiv preprint arXiv:2503.16416},
  year = {2025},
  eprint = {2503.16416},
  archivePrefix = {arXiv}
}

@inproceedings{mandlekar2018roboturk,
  author = {Mandlekar, Ajay and Zhu, Yuke and Garg, Animesh and Booher, Jonathan and Spero, Max and Tung, Albert and Gao, Julian and Emmons, John and Gupta, Anchit and Orbay, Emre and Savarese, Silvio and Fei-Fei, Li},
  title = {ROBOTURK: A Crowdsourcing Platform for Robotic Skill Learning through Imitation},
  booktitle = {Proceedings of the Conference on Robot Learning},
  year = {2018},
  eprint = {1811.02790},
  archivePrefix = {arXiv}
}

@article{feng2025embodiedagi,
  author = {Yequan Wang and Aixin Sun},
  title = {Toward Embodied AGI: A Review of Embodied AI and the Road Ahead},
  journal = {arXiv preprint arXiv:2505.14235},
  year = {2025},
  eprint = {2505.14235},
  archivePrefix = {arXiv}
}

@inproceedings{chang2017matterport3d,
  title={Matterport3D: Learning from {RGB-D} Data in Indoor Environments},
  author={Chang, Angel and Dai, Angela and Funkhouser, Thomas and Halber, Maciej and Nie{\ss}ner, Matthias and Savva, Manolis and Song, Shuran and Zeng, Andy and Zhang, Yinda},
  booktitle={3DV},
  year={2017}
}

@inproceedings{dai2017scannet,
  title={ScanNet: Richly-Annotated 3D Reconstructions of Indoor Scenes},
  author={Dai, Angela and Chang, Angel X. and Savva, Manolis and Halber, Maciej and Funkhouser, Thomas and Nie{\ss}ner, Matthias},
  booktitle={CVPR},
  year={2017}
}

@inproceedings{anderson2018vision,
  title={Vision-and-Language Navigation: Interpreting Visually-Grounded Navigation Instructions in Real Environments},
  author={Anderson, Peter and Wu, Qi and Teney, Damien and Bruce, Jake and Johnson, Mark and S{\"u}nderhauf, Niko and Drummond, Tom and Gould, Stephen and van den Hengel, Anton},
  booktitle={CVPR},
  year={2018}
}

@article{kolve2017ai2thor,
  author = {Eric Kolve and Roozbeh Mottaghi and Winson Han and Eli VanderBilt and Luca Weihs and Alvaro Herrasti and Matt Deitke and Kiana Ehsani and Daniel Gordon and Yuke Zhu and Aniruddha Kembhavi and Abhinav Gupta and Ali Farhadi},
  title = {AI2-THOR: An Interactive 3D Environment for Visual AI},
  journal = {arXiv preprint arXiv:1712.05474},
  year = {2017},
  eprint = {1712.05474},
  archivePrefix = {arXiv}
}

@inproceedings{savva2019habitat,
  title={Habitat: A Platform for Embodied AI Research},
  author={Savva, Manolis and Kadian, Abhinav and Maksymets, Oleksandr and Zhao, Yili and Wijmans, Erik and Jain, Bhavana and Straub, Julian and Liu, Jia and Koltun, Vladlen and Malik, Jitendra and others},
  booktitle={ICCV},
  year={2019}
}

@inproceedings{szot2021habitat2,
  author = {Andrew Szot and Alex Clegg and Eric Undersander and Erik Wijmans and Yili Zhao and John Turner and Noah Maestre and Mustafa Mukadam and Devendra Chaplot and Oleksandr Maksymets and Aaron Gokaslan and Vladimir Vondrus and Sameer Dharur and Franziska Meier and Wojciech Galuba and Angel Chang and Zsolt Kira and Vladlen Koltun and Jitendra Malik and Manolis Savva and Dhruv Batra},
  title = {Habitat 2.0: Training Home Assistants to Rearrange their Habitat},
  booktitle = {Advances in Neural Information Processing Systems 34},
  publisher = {Curran Associates, Inc.},
  year = {2021}
}

@inproceedings{shridhar2020alfred,
  title={ALFRED: A Benchmark for Interpreting Grounded Instructions for Everyday Tasks},
  author={Shridhar, Mohit and Manuelli, Lucas and Fox, Dieter},
  booktitle={CVPR},
  year={2020}
}

@inproceedings{savva2021behavior,
  author = {Savva, Manolis and others},
  title = {BEHAVIOR: Benchmark for Everyday Household Activities in Virtual, Interactive, and Ecological Environments},
  booktitle = {Proceedings of the Conference on Robot Learning},
  publisher = {PMLR},
  year = {2021}
}

@inproceedings{liang2024omnigibson,
  author = {Chengshu Li and Ruohan Zhang and Josiah Wong and Cem Gokmen and Sanjana Srivastava and Roberto Martín-Martín and Chen Wang and Gabrael Levine and Wensi Ai and Benjamin Martinez and Hang Yin and Michael Lingelbach and Minjune Hwang and Ayano Hiranaka and Sujay Garlanka and Arman Aydin and Sharon Lee and Jiankai Sun and Mona Anvari and Manasi Sharma and Dhruva Bansal and Samuel Hunter and Kyu-Young Kim and Alan Lou and Caleb R Matthews and Ivan Villa-Renteria and Jerry Huayang Tang and Claire Tang and Fei Xia and Yunzhu Li and Silvio Savarese and Hyowon Gweon and C. Karen Liu and Jiajun Wu and Li Fei-Fei},
  title = {{BEHAVIOR}-1K: A Human-Centered, Embodied AI Benchmark with 1,000 Everyday Activities and Realistic Simulation},
  booktitle = {Proceedings of the 6th Conference on Robot Learning},
  publisher = {PMLR},
  year = {2022}
}

@inproceedings{puig2024habitat3,
  author = {Xavier Puig and Eric Undersander and Andrew Szot and Mikael Dallaire Cote and Tsung-Yen Yang and Ruslan Partsey and Ruta Desai and Alexander William Clegg and Michal Hlavac and So Yeon Min and Vladimír Vondruš and Theophile Gervet and Vincent-Pierre Berges and John M. Turner and Oleksandr Maksymets and Zsolt Kira and Mrinal Kalakrishnan and Jitendra Malik and Devendra Singh Chaplot and Unnat Jain and Dhruv Batra and Akshara Rai and Roozbeh Mottaghi},
  title = {Habitat 3.0: A Co-Habitat for Humans, Avatars and Robots},
  booktitle = {International Conference on Learning Representations},
  year = {2024}
}

@article{straub2019replica,
  author = {Julian Straub and Thomas Whelan and Lingni Ma and Yufan Chen and Erik Wijmans and Simon Green and Jakob J. Engel and Raul Mur-Artal and Carl Ren and Shobhit Verma and Anton Clarkson and Mingfei Yan and Brian Budge and Yajie Yan and Xiaqing Pan and June Yon and Yuyang Zou and Kimberly Leon and Nigel Carter and Jesus Briales and Tyler Gillingham and Elias Mueggler and Luis Pesqueira and Manolis Savva and Dhruv Batra and Hauke M. Strasdat and Renzo De Nardi and Michael Goesele and Steven Lovegrove and Richard Newcombe},
  title = {The Replica Dataset: A Digital Replica of Indoor Spaces},
  journal = {arXiv preprint arXiv:1906.05797},
  year = {2019},
  eprint = {1906.05797},
  archivePrefix = {arXiv}
}

@inproceedings{ramakrishnan2021hm3d,
  title={Habitat Matterport 3D Dataset ({HM}3D): 1000 Large-scale 3D Environments for Embodied {AI}},
  author={Ramakrishnan, Santhosh Kumar and Gokaslan, Aaron and Maksymets, Oleksandr and Clegg, Alexander and Turner, John and Undersander, Eric and Galuba, Wojciech and Westbury, Andrew and Chang, Angel and Savva, Manolis and Zhao, Yili and Batra, Dhruv},
  booktitle={Thirty-fifth Conference on Neural Information Processing Systems Datasets and Benchmarks Track},
  year={2021}
}

@inproceedings{li2022igibson2,
  title={iGibson 2.0: Object-Centric Simulation for Robot Learning of Everyday Household Tasks},
  author={Li, Chengshu and Xia, Fei and Martin-Martin, Roberto and Lingelbach, Michael and Srivastava, Sanjana and Shen, Bokui and Vainio, Kent Elliott and Gokmen, Cem and Dharan, Gokul and Jain, Tanish and Kurenkov, Andrey and Liu, Karen and Gweon, Hyowon and Wu, Jiajun and Fei-Fei, Li and Savarese, Silvio},
  booktitle={Conference on Robot Learning},
  year={2022}
}

@inproceedings{james2020rlbench,
  title={RLBench: The Robot Learning Benchmark and Learning Environment},
  author={James, Stephen and Ma, Zicong and Arrojo, Diego and Davison, Andrew},
  booktitle={IEEE Robotics and Automation Letters},
  year={2020}
}

@inproceedings{xi2020sapien,
  title={SAPIEN: A SimulAted Part-based Interactive ENvironment},
  author={Xi, N. and Batra, D. and others},
  booktitle={CVPR},
  year={2020}
}

@inproceedings{gu2023maniskill2,
  author = {Jiayuan Gu and Fanbo Xiang and Xuanlin Li and Zhan Ling and Xiqiang Liu and Tongzhou Mu and Yihe Tang and Stone Tao and Xinyue Wei and Yunchao Yao and Xiaodi Yuan and Pengwei Xie and Zhiao Huang and Rui Chen and Hao Su},
  title = {ManiSkill2: A Unified Benchmark for Generalizable Manipulation Skills},
  booktitle = {International Conference on Learning Representations},
  year = {2023}
}

@article{baruch2021arkitscenes,
  title={{ARKitScenes}: A Diverse Real-World Dataset for 3D Indoor Scene Understanding Using Mobile {RGB-D} Data},
  author={Baruch, Gilad and Chen, Zhuoyuan and Dehghan, Afshin and Dimry, Tal and Feigin, Yuri and Fu, Peter and Gebauer, Thomas and Joffe, Brandon and Kurz, Daniel and Schwartz, Arik and Shulman, Elad},
  journal={Advances in Neural Information Processing Systems},
  year={2021}
}

@inproceedings{chen2020scanrefer,
  title={ScanRefer: 3D Object Localization in {RGB-D} Scans Using Natural Language},
  author={Chen, Dave Zhenyu and Chang, Angel X. and Nie{\ss}ner, Matthias},
  booktitle={ECCV},
  year={2020}
}

@article{martinmartin2023jrdb,
  title={JRDB: A Dataset and Benchmark of Egocentric Robot Visual Perception of Humans in Built Environments},
  author={Martin-Martin, Roberto and Patel, Mihir and Rezatofighi, Hamid and Shenoi, Abhijeet and Gwak, Junyoung and Frankel, Eric and Sadeghian, Amir and Savarese, Silvio},
  journal={IEEE Transactions on Pattern Analysis and Machine Intelligence},
  volume={45},
  number={6},
  pages={6748--6765},
  year={2023}
}

@inproceedings{wang2024embodiedscan,
  title={EmbodiedScan: A Holistic Multi-Modal 3D Perception Suite Towards Embodied {AI}},
  author={Wang, Tai and Mao, Xiaohan and Zhu, Chenming and Xu, Runsen and Lyu, Ruiyuan and Li, Peisen and Chen, Xiao and Zhang, Wenwei and Chen, Kai and Xue, Tianfan and Liu, Xihui and Lu, Cewu and Lin, Dahua and Pang, Jiangmiao},
  booktitle={CVPR},
  year={2024}
}

@inproceedings{du2024embspatial,
  title={EmbSpatial-Bench: Benchmarking Spatial Understanding for Embodied Tasks with Large Vision-Language Models},
  url={http://dx.doi.org/10.18653/v1/2024.acl-short.33},
  DOI={10.18653/v1/2024.acl-short.33},
  booktitle={Proceedings of the 62nd Annual Meeting of the Association for Computational Linguistics (Volume 2: Short Papers)},
  publisher={Association for Computational Linguistics},
  author={Du, Mengfei and Wu, Binhao and Li, Zejun and Huang, Xuanjing and Wei, Zhongyu},
  year={2024},
  pages={346--355} }

@article{mees2022calvin,
  title={CALVIN: A Benchmark for Language-Conditioned Policy Learning for Long-Horizon Robot Manipulation Tasks},
  author={Mees, Oier and Hermann, Lukas and Rosete-Beas, Erick and Burgard, Wolfram},
  journal={IEEE Robotics and Automation Letters},
  year={2022}
}

@inproceedings{liu2023libero, series={NeurIPS 2023},
  title={LIBERO: Benchmarking Knowledge Transfer for Lifelong Robot Learning},
  url={http://dx.doi.org/10.52202/075280-1939},
  DOI={10.52202/075280-1939},
  booktitle={Advances in Neural Information Processing Systems 36},
  publisher={Neural Information Processing Systems Foundation, Inc. (NeurIPS)},
  author={Feng, Yihao and Gao, Chongkai and Liu, Bo and Liu, Qiang and Stone, Peter and Zhu, Yifeng and Zhu, Yuke},
  year={2023},
  pages={44776--44791},
  collection={NeurIPS 2023} }

@inproceedings{yenamandra2023homerobot,
  author = {Sriram Yenamandra and Arun Ramachandran and Karmesh Yadav and Austin Wang and Mukul Khanna and Theophile Gervet and Tsung-Yen Yang and Vidhi Jain and Alexander William Clegg and John Turner and Zsolt Kira and Manolis Savva and Angel Chang and Devendra Singh Chaplot and Dhruv Batra and Roozbeh Mottaghi and Yonatan Bisk and Chris Paxton},
  title = {HomeRobot: Open-Vocabulary Mobile Manipulation},
  booktitle = {Proceedings of the Conference on Robot Learning},
  publisher = {PMLR},
  year = {2023}
}

@inproceedings{openx2023,
  title={Open X-Embodiment: Robotic Learning Datasets and RT-X Models : Open X-Embodiment Collaboration},
  url={http://dx.doi.org/10.1109/ICRA57147.2024.10611477},
  DOI={10.1109/icra57147.2024.10611477},
  booktitle={2024 IEEE International Conference on Robotics and Automation (ICRA)},
  publisher={IEEE},
  author={O'Neill, Abby and Rehman, Abdul and Maddukuri, Abhiram and Gupta, Abhishek and Padalkar, Abhishek and Lee, Abraham and Pooley, Acorn and Gupta, Agrim and Mandlekar, Ajay and Jain, Ajinkya and Tung, Albert and Bewley, Alex and Herzog, Alex and Irpan, Alex and Khazatsky, Alexander and Rai, Anant and Gupta, Anchit and Wang, Andrew and Singh, Anikait and Garg, Animesh and Kembhavi, Aniruddha and Xie, Annie and Brohan, Anthony and Raffin, Antonin and Sharma, Archit and Yavary, Arefeh and Jain, Arhan and Balakrishna, Ashwin and Wahid, Ayzaan and Burgess-Limerick, Ben and Kim, Beomjoon and Schölkopf, Bernhard and Wulfe, Blake and Ichter, Brian and Lu, Cewu and Xu, Charles and Le, Charlotte and Finn, Chelsea and Wang, Chen and Xu, Chenfeng and Chi, Cheng and Huang, Chenguang and Chan, Christine and Agia, Christopher and Pan, Chuer and Fu, Chuyuan and Devin, Coline and Xu, Danfei and Morton, Daniel and Driess, Danny and Chen, Daphne and Pathak, Deepak and Shah, Dhruv and Büchler, Dieter and Jayaraman, Dinesh and Kalashnikov, Dmitry and Sadigh, Dorsa and Johns, Edward and Foster, Ethan and Liu, Fangchen and Ceola, Federico and Xia, Fei and Zhao, Feiyu and Stulp, Freek and Zhou, Gaoyue and Sukhatme, Gaurav S. and Salhotra, Gautam and Yan, Ge and Feng, Gilbert and Schiavi, Giulio and Berseth, Glen and Kahn, Gregory and Wang, Guanzhi and Su, Hao and Fang, Hao-Shu and Shi, Haochen and Bao, Henghui and Ben Amor, Heni and Christensen, Henrik I and Furuta, Hiroki and Walke, Homer and Fang, Hongjie and Ha, Huy and Mordatch, Igor and Radosavovic, Ilija and Leal, Isabel and Liang, Jacky and Abou-Chakra, Jad and Kim, Jaehyung and Drake, Jaimyn and Peters, Jan and Schneider, Jan and Hsu, Jasmine and Bohg, Jeannette and Bingham, Jeffrey and Wu, Jeffrey and Gao, Jensen and Hu, Jiaheng and Wu, Jiajun and Wu, Jialin and Sun, Jiankai and Luo, Jianlan and Gu, Jiayuan and Tan, Jie and Oh, Jihoon and Wu, Jimmy and Lu, Jingpei and Yang, Jingyun and Malik, Jitendra and Silvério, João and Hejna, Joey and Booher, Jonathan and Tompson, Jonathan and Yang, Jonathan and Salvador, Jordi and Lim, Joseph J. and Han, Junhyek and Wang, Kaiyuan and Rao, Kanishka and Pertsch, Karl and Hausman, Karol and Go, Keegan and Gopalakrishnan, Keerthana and Goldberg, Ken and Byrne, Kendra and Oslund, Kenneth and Kawaharazuka, Kento and Black, Kevin and Lin, Kevin and Zhang, Kevin and Ehsani, Kiana and Lekkala, Kiran and Ellis, Kirsty and Rana, Krishan and Srinivasan, Krishnan and Fang, Kuan and Singh, Kunal Pratap and Zeng, Kuo-Hao and Hatch, Kyle and Hsu, Kyle and Itti, Laurent and Chen, Lawrence Yunliang and Pinto, Lerrel and Fei-Fei, Li and Tan, Liam and Fan, Linxi Jim and Ott, Lionel and Lee, Lisa and Weihs, Luca and Chen, Magnum and Lepert, Marion and Memmel, Marius and Tomizuka, Masayoshi and Itkina, Masha and Castro, Mateo Guaman and Spero, Max and Du, Maximilian and Ahn, Michael and Yip, Michael C. and Zhang, Mingtong and Ding, Mingyu and Heo, Minho and Srirama, Mohan Kumar and Sharma, Mohit and Kim, Moo Jin and Kanazawa, Naoaki and Hansen, Nicklas and Heess, Nicolas and Joshi, Nikhil J and Suenderhauf, Niko and Liu, Ning and Di Palo, Norman and Shafiullah, Nur Muhammad Mahi and Mees, Oier and Kroemer, Oliver and Bastani, Osbert and Sanketi, Pannag R and Miller, Patrick Tree and Yin, Patrick and Wohlhart, Paul and Xu, Peng and Fagan, Peter David and Mitrano, Peter and Sermanet, Pierre and Abbeel, Pieter and Sundaresan, Priya and Chen, Qiuyu and Vuong, Quan and Rafailov, Rafael and Tian, Ran and Doshi, Ria and Martín-Martín, Roberto and Baijal, Rohan and Scalise, Rosario and Hendrix, Rose and Lin, Roy and Qian, Runjia and Zhang, Ruohan and Mendonca, Russell and Shah, Rutav and Hoque, Ryan and Julian, Ryan and Bustamante, Samuel and Kirmani, Sean and Levine, Sergey and Lin, Shan and Moore, Sherry and Bahl, Shikhar and Dass, Shivin and Sonawani, Shubham and Song, Shuran and Xu, Sichun and Haldar, Siddhant and Karamcheti, Siddharth and Adebola, Simeon and Guist, Simon and Nasiriany, Soroush and Schaal, Stefan and Welker, Stefan and Tian, Stephen and Ramamoorthy, Subramanian and Dasari, Sudeep and Belkhale, Suneel and Park, Sungjae and Nair, Suraj and Mirchandani, Suvir and Osa, Takayuki and Gupta, Tanmay and Harada, Tatsuya and Matsushima, Tatsuya and Xiao, Ted and Kollar, Thomas and Yu, Tianhe and Ding, Tianli and Davchev, Todor and Zhao, Tony Z. and Armstrong, Travis and Darrell, Trevor and Chung, Trinity and Jain, Vidhi and Vanhoucke, Vincent and Zhan, Wei and Zhou, Wenxuan and Burgard, Wolfram and Chen, Xi and Wang, Xiaolong and Zhu, Xinghao and Geng, Xinyang and Liu, Xiyuan and Liangwei, Xu and Li, Xuanlin and Lu, Yao and Ma, Yecheng Jason and Kim, Yejin and Chebotar, Yevgen and Zhou, Yifan and Zhu, Yifeng and Wu, Yilin and Xu, Ying and Wang, Yixuan and Bisk, Yonatan and Cho, Yoonyoung and Lee, Youngwoon and Cui, Yuchen and Cao, Yue and Wu, Yueh-Hua and Tang, Yujin and Zhu, Yuke and Zhang, Yunchu and Jiang, Yunfan and Li, Yunshuang and Li, Yunzhu and Iwasawa, Yusuke and Matsuo, Yutaka and Ma, Zehan and Xu, Zhuo and Cui, Zichen Jeff and Zhang, Zichen and Lin, Zipeng},
  year={2024},
  month={May}, pages={6892--6903} }

@inproceedings{khazatsky2024droid, series={RSS2024},
  title={DROID: A Large-Scale In-The-Wild Robot Manipulation Dataset},
  url={http://dx.doi.org/10.15607/RSS.2024.XX.120},
  DOI={10.15607/rss.2024.xx.120},
  booktitle={Robotics: Science and Systems XX},
  publisher={Robotics: Science and Systems Foundation},
  author={Khazatsky, Alexander and Pertsch, Karl and Nair, Suraj and Balakrishna, Ashwin and Dasari, Sudeep and Karamcheti, Siddharth and Nasiriany, Soroush and Srirama, Mohan and Chen, Lawrence and Ellis, Kirsty and Fagan, Peter and Hejna, Joey and Itkina, Masha and Lepert, Marion and Ma, Yecheng and Miller, Patrick and Wu, Jimmy and Belkhale, Suneel and Dass, Shivin and Ha, Huy and Jain, Arhan and Lee, Abraham and Lee, Youngwoon and Memmel, Marius and Park, Sungjae and Radosavovic, Ilija and Wang, Kaiyuan and Zhan, Albert and Black, Kevin and Chi, Cheng and Hatch, Kyle and Lin, Shan and Lu, Jingpei and Mercat, Jean and Rehman, Abdul and Sanketi, Pannag and Sharma, Archit and Simpson, Cody and Vuong, Quan and Walke, Homer and Wulfe, Blake and Xiao, Ted and Yang, Jonathan and Yavary, Arefeh and Zhao, Tony and Agia, Christopher and Baijal, Rohan and Castro, Mateo and Chen, Daphne and Chen, Qiuyu and Chung, Trinity and Drake, Jaimyn and Foster, Ethan and Gao, Jensen and Herrera, David and Heo, Minho and Hsu, Kyle and Hu, Jiaheng and Jackson, Donovon and Le, Charlotte and Li, Yunshuang and Lin, Roy and Ma, Zehan and Maddukuri, Abhiram and Mirchandani, Suvir and Morton, Daniel and Nguyen, Tony and O'Neill, Abigail and Scalise, Rosario and Seale, Derick and Son, Victor and Tian, Stephen and Tran, Emi and Wang, Andrew and Wu, Yilin and Xie, Annie and Yang, Jingyun and Yin, Patrick and Zhang, Yunchu and Bastani, Osbert and Berseth, Glen and Bohg, Jeannette and Goldberg, Ken and Gupta, Abhinav and Gupta, Abhishek and Jayaraman, Dinesh and Lim, Joseph and Malik, Jitendra and Martín-Martín, Roberto and Ramamoorthy, Subramanian and Sadigh, Dorsa and Song, Shuran and Wu, Jiajun and Yip, Michael and Zhu, Yuke and Kollar, Thomas and Levine, Sergey and Finn, Chelsea},
  year={2024},
  month={July}, collection={RSS2024} }

@article{parakh2025anybody,
  author = {Meenal Parakh and Alexandre Kirchmeyer and Beining Han and Jia Deng},
  title = {AnyBody: A Benchmark Suite for Cross-Embodiment Manipulation},
  journal = {arXiv preprint arXiv:2505.14986},
  year = {2025},
  eprint = {2505.14986},
  archivePrefix = {arXiv}
}

@inproceedings{wang2024simpler,
  author = {Xuanlin Li and Kyle Hsu and Jiayuan Gu and Karl Pertsch and Oier Mees and Homer Rich Walke and Chuyuan Fu and Ishikaa Lunawat and Isabel Sieh and Sean Kirmani and Sergey Levine and Jiajun Wu and Chelsea Finn and Hao Su and Quan Vuong and Ted Xiao},
  title = {Evaluating Real-World Robot Manipulation Policies in Simulation},
  booktitle = {Proceedings of the 8th Conference on Robot Learning},
  publisher = {PMLR},
  year = {2024}
}

@article{straub2024efm3d,
  author = {Julian Straub and Daniel DeTone and Tianwei Shen and Nan Yang and Chris Sweeney and Richard Newcombe},
  title = {EFM3D: A Benchmark for Measuring Progress Towards 3D Egocentric Foundation Models},
  journal = {arXiv preprint arXiv:2406.10224},
  year = {2024},
  eprint = {2406.10224},
  archivePrefix = {arXiv}
}

@inproceedings{deitke2022objaverse,
  title={Objaverse: A Universe of Annotated 3D Objects},
  url={http://dx.doi.org/10.1109/CVPR52729.2023.01263},
  DOI={10.1109/cvpr52729.2023.01263},
  booktitle={2023 IEEE/CVF Conference on Computer Vision and Pattern Recognition (CVPR)},
  publisher={IEEE},
  author={Deitke, Matt and Schwenk, Dustin and Salvador, Jordi and Weihs, Luca and Michel, Oscar and VanderBilt, Eli and Schmidt, Ludwig and Ehsanit, Kiana and Kembhavi, Aniruddha and Farhadi, Ali},
  year={2023},
  month={June}, pages={13142--13153} }

@inproceedings{deitke2023objaversexl, series={NeurIPS 2023},
  title={Objaverse-XL: A Universe of 10M+ 3D Objects},
  url={http://dx.doi.org/10.52202/075280-1554},
  DOI={10.52202/075280-1554},
  booktitle={Advances in Neural Information Processing Systems 36},
  publisher={Neural Information Processing Systems Foundation, Inc. (NeurIPS)},
  author={Deitke, Matt and Ehsani, Kiana and Fan, Alan and Farhadi, Ali and Gadre, Samir Yitzhak and Gkioxari, Georgia and Kembhavi, Aniruddha and Kusupati, Aditya and Laforte, Christian and Liu, Ruoshi and Michel, Oscar and Ngo, Huong and Schmidt, Ludwig and Vanderbilt, Eli and Voleti, Vikram and Vondrick, Carl and Wallingford, Matthew},
  year={2023},
  pages={35799--35813},
  collection={NeurIPS 2023} }

@inproceedings{ke2024robocasa, series={RSS2024},
  title={RoboCasa: Large-Scale Simulation of Household Tasks for Generalist Robots},
  url={http://dx.doi.org/10.15607/RSS.2024.XX.050},
  DOI={10.15607/rss.2024.xx.050},
  booktitle={Robotics: Science and Systems XX},
  publisher={Robotics: Science and Systems Foundation},
  author={Nasiriany, Soroush and Maddukuri, Abhiram and Zhang, Lance and Parikh, Adeet and Lo, Aaron and Joshi, Abhishek and Mandlekar, Ajay and Zhu, Yuke},
  year={2024},
  month={July}, collection={RSS2024} }

@article{nasiriany2026robocasa365,
  author = {Soroush Nasiriany and Sepehr Nasiriany and Abhiram Maddukuri and Yuke Zhu},
  title = {RoboCasa365: A Large-Scale Simulation Framework for Training and Benchmarking Generalist Robots},
  journal = {arXiv preprint arXiv:2603.04356},
  year = {2026},
  eprint = {2603.04356},
  archivePrefix = {arXiv}
}

@inproceedings{zhang2024vlabench,
  title={VLABench: A Large-Scale Benchmark for Language-Conditioned Robotics Manipulation with Long-Horizon Reasoning Tasks},
  url={http://dx.doi.org/10.1109/ICCV51701.2025.01037},
  DOI={10.1109/iccv51701.2025.01037},
  booktitle={2025 IEEE/CVF International Conference on Computer Vision (ICCV)},
  publisher={IEEE},
  author={Zhang, Shiduo and Xu, Zhe and Liu, Peiju and Yu, Xiaopeng and Li, Yuan and Gao, Qinghui and Fei, Zhaoye and Yin, Zhangyue and Wu, Zuxuan and Jiang, Yu-Gang and Qiu, Xipeng},
  year={2025},
  month={Oct}, pages={11142--11152} }

@article{li2024eai,
  title={Embodied Agent Interface: Benchmarking LLMs for Embodied Decision Making},
  author={Li, Manling and Zhao, Shiyu and Wang, Qineng and Wang, Kangrui and Zhou, Yu and others},
  journal={Advances in Neural Information Processing Systems},
  year={2024}
}

@article{cheng2025embodiedeval,
  author = {Zhili Cheng and Yuge Tu and Ran Li and Shiqi Dai and Jinyi Hu and Shengding Hu and Jiahao Li and Yang Shi and Tianyu Yu and Weize Chen and Lei Shi and Maosong Sun},
  title = {EmbodiedEval: Evaluate Multimodal LLMs as Embodied Agents},
  journal = {arXiv preprint arXiv:2501.11858},
  year = {2025},
  eprint = {2501.11858},
  archivePrefix = {arXiv}
}

@article{yang2025embodiedbench,
  author = {Rui Yang and Hanyang Chen and Junyu Zhang and Mark Zhao and Cheng Qian and Kangrui Wang and Qineng Wang and Teja Venkat Koripella and Marziyeh Movahedi and Manling Li and Heng Ji and Huan Zhang and Tong Zhang},
  title = {EmbodiedBench: Comprehensive Benchmarking Multi-modal Large Language Models for Vision-Driven Embodied Agents},
  journal = {arXiv preprint arXiv:2502.09560},
  year = {2025},
  eprint = {2502.09560},
  archivePrefix = {arXiv}
}

@article{yin2024safeagentbench,
  author = {Sheng Yin and Xianghe Pang and Yuanzhuo Ding and Menglan Chen and Yutong Bi and Yichen Xiong and Wenhao Huang and Zhen Xiang and Jing Shao and Siheng Chen},
  title = {SafeAgentBench: A Benchmark for Safe Task Planning of Embodied LLM Agents},
  journal = {arXiv preprint arXiv:2412.13178},
  year = {2024},
  eprint = {2412.13178},
  archivePrefix = {arXiv}
}

@article{qin2026embodiedgovbench,
  author = {Xue Qin and Simin Luan and John See and Cong Yang and Zhijun Li},
  title = {EmbodiedGovBench: A Benchmark for Governance, Recovery, and Upgrade Safety in Embodied Agent Systems},
  journal = {arXiv preprint arXiv:2604.11174},
  year = {2026},
  eprint = {2604.11174},
  archivePrefix = {arXiv}
}

@inproceedings{deitke2022procthor,
  author = {Matt Deitke and Eli VanderBilt and Alvaro Herrasti and Luca Weihs and Jordi Salvador and Kiana Ehsani and Winson Han and Eric Kolve and Ali Farhadi and Aniruddha Kembhavi and Roozbeh Mottaghi},
  title = {ProcTHOR: Large-Scale Embodied AI Using Procedural Generation},
  booktitle = {Advances in Neural Information Processing Systems 35},
  publisher = {Curran Associates, Inc.},
  year = {2022}
}

@inproceedings{mandlekar2023mimicgen,
  title={MimicGen: A Data Generation System for Scalable Robot Learning using Human Demonstrations},
  author={Mandlekar, Ajay and Nasiriany, Soroush and Wen, Bowen and Akinola, Iretiayo and Narang, Yashraj and Fan, Linxi and Zhu, Yuke and Fox, Dieter},
  booktitle={Conference on Robot Learning},
  year={2023}
}

@inproceedings{wang2023gensim,
  author = {Wang, Lirui and Ling, Yiyang and Yuan, Zhecheng and Shridhar, Mohit and Bao, Chen and Qin, Yuzhe and Wang, Bailin and Xu, Huazhe and Wang, Xiaolong},
  title = {GenSim: Generating Robotic Simulation Tasks via Large Language Models},
  booktitle = {International Conference on Learning Representations},
  year = {2024}
}

@article{hua2024gensim2,
  author = {Pu Hua and Minghuan Liu and Annabella Macaluso and Yunfeng Lin and Weinan Zhang and Huazhe Xu and Lirui Wang},
  title = {GenSim2: Scaling Robot Data Generation with Multi-modal and Reasoning LLMs},
  journal = {arXiv preprint arXiv:2410.03645},
  year = {2024},
  eprint = {2410.03645},
  archivePrefix = {arXiv}
}

@inproceedings{yang2023robogen,
  author = {Yufei Wang and Zhou Xian and Feng Chen and Tsun-Hsuan Wang and Yian Wang and Katerina Fragkiadaki and Zackory Erickson and David Held and Chuang Gan},
  title = {RoboGen: Towards Unleashing Infinite Data for Automated Robot Learning via Generative Simulation},
  booktitle = {Proceedings of the 41st International Conference on Machine Learning},
  publisher = {PMLR},
  year = {2024}
}

@inproceedings{ma2024holodeck,
  title={Holodeck: Language Guided Generation of 3D Embodied AI Environments},
  url={http://dx.doi.org/10.1109/CVPR52733.2024.01536},
  DOI={10.1109/cvpr52733.2024.01536},
  booktitle={2024 IEEE/CVF Conference on Computer Vision and Pattern Recognition (CVPR)},
  publisher={IEEE},
  author={Yang, Yue and Sun, Fan-Yun and Weihs, Luca and Vanderbilt, Eli and Herrasti, Alvaro and Han, Winson and Wu, Jiajun and Haber, Nick and Krishna, Ranjay and Liu, Lingjie and Callison-Burch, Chris and Yatskar, Mark and Kembhavi, Aniruddha and Clark, Christopher},
  year={2024},
  month={June}, pages={16277--16287} }

@inproceedings{ma2024eureka,
  author = {Yecheng Jason Ma and William Liang and Guanzhi Wang and De-An Huang and Osbert Bastani and Dinesh Jayaraman and Yuke Zhu and Linxi Fan and Anima Anandkumar},
  title = {Eureka: Human-Level Reward Design via Coding Large Language Models},
  booktitle = {International Conference on Learning Representations},
  year = {2024}
}

@inproceedings{vuong2024dreureka, series={RSS2024},
  title={DrEureka: Language Model Guided Sim-To-Real Transfer},
  url={http://dx.doi.org/10.15607/RSS.2024.XX.094},
  DOI={10.15607/rss.2024.xx.094},
  booktitle={Robotics: Science and Systems XX},
  publisher={Robotics: Science and Systems Foundation},
  author={Ma, Yecheng and Liang, William and Wang, Hung-Ju and Zhu, Yuke and Fan, Linxi and Bastani, Osbert and Jayaraman, Dinesh},
  year={2024},
  month={July}, collection={RSS2024} }

@article{fan2024grutopia,
  author = {Hanqing Wang and Jiahe Chen and Wensi Huang and Qingwei Ben and Tai Wang and Boyu Mi and Tao Huang and Siheng Zhao and Yilun Chen and Sizhe Yang and Peizhou Cao and Wenye Yu and Zichao Ye and Jialun Li and Junfeng Long and Zirui Wang and Huiling Wang and Ying Zhao and Zhongying Tu and Yu Qiao and Dahua Lin and Jiangmiao Pang},
  title = {GRUtopia: Dream General Robots in a City at Scale},
  journal = {arXiv preprint arXiv:2407.10943},
  year = {2024},
  eprint = {2407.10943},
  archivePrefix = {arXiv}
}

@article{xu2024infiniteworld,
  author = {Pengzhen Ren and Min Li and Zhen Luo and Xinshuai Song and Ziwei Chen and Weijia Liufu and Yixuan Yang and Hao Zheng and Rongtao Xu and Zitong Huang and Tongsheng Ding and Luyang Xie and Kaidong Zhang and Changfei Fu and Yang Liu and Liang Lin and Feng Zheng and Xiaodan Liang},
  title = {InfiniteWorld: A Unified Scalable Simulation Framework for General Visual-Language Robot Interaction},
  journal = {arXiv preprint arXiv:2412.05789},
  year = {2024},
  eprint = {2412.05789},
  archivePrefix = {arXiv}
}

@article{wang2025roboeval,
  author = {Yi Ru Wang and Carter Ung and Christopher Tan and Grant Tannert and Jiafei Duan and Josephine Li and Anh Le and Rishabh Oswal and Markus Grotz and Wilbert Pumacay and Yuquan Deng and Ranjay Krishna and Dieter Fox and Siddhartha Srinivasa},
  title = {RoboEval: Where Robotic Manipulation Meets Structured and Scalable Evaluation},
  journal = {arXiv preprint arXiv:2507.00435},
  year = {2025},
  eprint = {2507.00435},
  archivePrefix = {arXiv}
}

@article{li2026affordsim,
  author = {Mingyang Li and Haofan Xu and Haowen Sun and Xinzhe Chen and Sihua Ren and Liqi Huang and Xinyang Sui and Chenyang Miao and Jiawei Ye and Qiongjie Cui and Zeyang Liu and Xingyu Chen and Xuguang Lan},
  title = {AffordSim: A Scalable Data Generator and Benchmark for Affordance-Aware Robotic Manipulation},
  journal = {arXiv preprint arXiv:2604.11674},
  year = {2026},
  eprint = {2604.11674},
  archivePrefix = {arXiv}
}

@inproceedings{kiela2021dynabench,
  title={Dynabench: Rethinking Benchmarking in NLP},
  author={Kiela, Douwe and Bartolo, Max and Nie, Yixin and Kaushik, Divyansh and Geiger, Atticus and Wu, Zhengxuan and Vidgen, Bertie and Prasad, Grusha and Singh, Amanpreet and Ringshia, Pratik and others},
  booktitle={NAACL},
  year={2021}
}

@inproceedings{mazumder2023dataperf,
  title={DataPerf: Benchmarks for Data-Centric {AI} Development},
  author={Mazumder, Mark and Banbury, Colby and Yao, Xiaozhe and Karlas, Bojan and Gaviria Rojas, William and Diamos, Sudnya and Diamos, Greg and He, Lynn and Aroyo, Lora and Acun, Bilge and others},
  booktitle={Advances in Neural Information Processing Systems},
  year={2023}
}

@article{liang2023helm,
  title={Holistic Evaluation of Language Models},
  author={Liang, Percy and Bommasani, Rishi and Lee, Tony and Tsipras, Dimitris and Soylu, Dilara and Yasunaga, Michihiro and Zhang, Yian and Narayanan, Deepak and others},
  journal={Transactions on Machine Learning Research},
  year={2023}
}

@inproceedings{wu2019evalai,
  title={EvalAI: Towards Better Evaluation Systems for AI Agents},
  author={Wu, Y. and others},
  booktitle={CVPR Workshops},
  year={2019}
}

@article{gebru2021datasheets,
  title={Datasheets for Datasets},
  author={Gebru, Timnit and Morgenstern, Jamie and Vecchione, Briana and Vaughan, Jennifer and Wallach, Hanna and Daum{\'e}, Hal and Crawford, Kate},
  journal={Communications of the ACM},
  volume={64},
  number={12},
  pages={86--92},
  year={2021}
}

@inproceedings{pushkarna2022datacards, series={FAccT '22},
  title={Data Cards: Purposeful and Transparent Dataset Documentation for Responsible AI},
  url={http://dx.doi.org/10.1145/3531146.3533231},
  DOI={10.1145/3531146.3533231},
  booktitle={2022 ACM Conference on Fairness Accountability and Transparency},
  publisher={ACM},
  author={Pushkarna, Mahima and Zaldivar, Andrew and Kjartansson, Oddur},
  year={2022},
  month={June}, pages={1776--1826},
  collection={FAccT '22} }

@inproceedings{mitchell2019modelcards,
  title={Model Cards for Model Reporting},
  author={Mitchell, Margaret and Wu, Simone and Zaldivar, Andrew and Barnes, Parker and Vasserman, Lucy and Hutchinson, Ben and Spitzer, Elena and Raji, Inioluwa and Gebru, Timnit},
  booktitle={FAT*},
  year={2019}
}

@article{sokol2024benchmarkcards,
  author = {Anna Sokol and Elizabeth Daly and Michael Hind and David Piorkowski and Xiangliang Zhang and Nuno Moniz and Nitesh Chawla},
  title = {BenchmarkCards: Standardized Documentation for Large Language Model Benchmarks},
  journal = {arXiv preprint arXiv:2410.12974},
  year = {2024},
  eprint = {2410.12974},
  archivePrefix = {arXiv}
}

@inproceedings{leitner2016apb,
  title={The ACRV picking benchmark: A robotic shelf picking benchmark to foster reproducible research},
  url={http://dx.doi.org/10.1109/ICRA.2017.7989545},
  DOI={10.1109/icra.2017.7989545},
  booktitle={2017 IEEE International Conference on Robotics and Automation (ICRA)},
  publisher={IEEE},
  author={Leitner, Jurgen and Tow, Adam W. and Sunderhauf, Niko and Dean, Jake E. and Durham, Joseph W. and Cooper, Matthew and Eich, Markus and Lehnert, Christopher and Mangels, Ruben and McCool, Christopher and Kujala, Peter T. and Nicholson, Lachlan and Pham, Trung and Sergeant, James and Wu, Liao and Zhang, Fangyi and Upcroft, Ben and Corke, Peter},
  year={2017},
  month={May}, pages={4705--4712} }

@article{paskov2024gpai,
  author = {Patricia Paskov and Lukas Berglund and Everett Smith and Lisa Soder},
  title = {GPAI Evaluations Standards Taskforce: Towards Effective AI Governance},
  journal = {arXiv preprint arXiv:2411.13808},
  year = {2024},
  eprint = {2411.13808},
  archivePrefix = {arXiv}
}

@inproceedings{reuel2024betterbench, series={NeurIPS 2024},
  title={BetterBench: Assessing AI Benchmarks, Uncovering Issues, and Establishing Best Practices},
  url={http://dx.doi.org/10.52202/079017-0685},
  DOI={10.52202/079017-0685},
  booktitle={Advances in Neural Information Processing Systems 37},
  publisher={Neural Information Processing Systems Foundation, Inc. (NeurIPS)},
  author={Hardy, Amelia and Hardy, Malcolm and Kochenderfer, Mykel and Lamparth, Max and Reuel, Anka and Smith, Chandler},
  year={2024},
  pages={21763--21813},
  collection={NeurIPS 2024} }

@article{staufer2025auditcards,
  author = {Leon Staufer and Mick Yang and Anka Reuel and Stephen Casper},
  title = {Audit Cards: Contextualizing AI Evaluations},
  journal = {arXiv preprint arXiv:2504.13839},
  year = {2025},
  eprint = {2504.13839},
  archivePrefix = {arXiv}
}

@article{yakefu2025robochallenge,
  author = {Adina Yakefu and Bin Xie and Chongyang Xu and Enwen Zhang and Erjin Zhou and Fan Jia and Haitao Yang and Haoqiang Fan and Haowei Zhang and Hongyang Peng and Jing Tan and Junwen Huang and Kai Liu and Kaixin Liu and Kefan Gu and Qinglun Zhang and Ruitao Zhang and Saike Huang and Shen Cheng and Shuaicheng Liu and Tiancai Wang and Tiezhen Wang and Wei Sun and Wenbin Tang and Yajun Wei and Yang Chen and Youqiang Gui and Yucheng Zhao and Yunchao Ma and Yunfei Wei and Yunhuan Yang and Yutong Guo and Ze Chen and Zhengyuan Du and Ziheng Zhang and Ziming Liu and Ziwei Yan},
  title = {RoboChallenge: Large-scale Real-robot Evaluation of Embodied Policies},
  journal = {arXiv preprint arXiv:2510.17950},
  year = {2025},
  eprint = {2510.17950},
  archivePrefix = {arXiv}
}

@inproceedings{shah2017airsim,
  title={AirSim: High-Fidelity Visual and Physical Simulation for Autonomous Vehicles},
  author={Shah, Shital and Dey, Debadeepta and Lovett, Chris and Kapoor, Ashish},
  booktitle={Field and Service Robotics},
  year={2017}
}

@inproceedings{dosovitskiy2017carla,
  title={{CARLA}: An Open Urban Driving Simulator},
  author={Dosovitskiy, Alexey and Ros, German and Codevilla, Felipe and Lopez, Antonio and Koltun, Vladlen},
  booktitle={Conference on Robot Learning},
  year={2017}
}

@inproceedings{zhou2020smarts,
  author = {Ming Zhou and Jun Luo and Julian Villella and Yaodong Yang and David Rusu and Jiayu Miao and Weinan Zhang and Montgomery Alban and Iman Fadakar and Zheng Chen and Aurora Chongxi Huang and Ying Wen and Kimia Hassanzadeh and Daniel Graves and Dong Chen and Zhengbang Zhu and Nhat Nguyen and Mohamed Elsayed and Kun Shao and Sanjeevan Ahilan and Baokuan Zhang and Jiannan Wu and Zhengang Fu and Kasra Rezaee and Peyman Yadmellat and Mohsen Rohani and Nicolas Perez Nieves and Yihan Ni and Seyedershad Banijamali and Alexander Cowen Rivers and Zheng Tian and Daniel Palenicek and Haitham bou Ammar and Hongbo Zhang and Wulong Liu and Jianye Hao and Jun Wang},
  title = {{SMARTS}: Scalable Multi-Agent Reinforcement Learning Training School for Autonomous Driving},
  booktitle = {Proceedings of the 4th Conference on Robot Learning},
  publisher = {PMLR},
  year = {2020}
}

@article{li2021metadrive,
  title={MetaDrive: Composing Diverse Driving Scenarios for Generalizable Reinforcement Learning},
  author={Li, Quanyi and Peng, Zhenghao and Feng, Lan and others},
  journal={IEEE Transactions on Pattern Analysis and Machine Intelligence},
  year={2021}
}

@inproceedings{song2021flightmare,
  title={Flightmare: A Flexible Quadrotor Simulator},
  author={Song, Yunlong and Naji, Selim and Kaufmann, Elia and Loquercio, Antonio and Scaramuzza, Davide},
  booktitle={Conference on Robot Learning},
  year={2021}
}

@article{liu2023aerialvln,
  title={AerialVLN: Vision-and-Language Navigation for UAVs},
  author={Liu, Shubo and Zhang, Hongsheng and Qi, Yuankai and Wang, Peng and Zhang, Yanning and Wu, Qi},
  journal={ICCV},
  year={2023}
}

@inproceedings{li2023scenarionet, series={NeurIPS 2023},
  title={ScenarioNet: Open-Source Platform for Large-Scale Traffic Scenario Simulation and Modeling},
  url={http://dx.doi.org/10.52202/075280-0172},
  DOI={10.52202/075280-0172},
  booktitle={Advances in Neural Information Processing Systems 36},
  publisher={Neural Information Processing Systems Foundation, Inc. (NeurIPS)},
  author={Duan, Chenda and Feng, Lan and Li, Quanyi and Liu, Zhizheng and Mo, Wenjie and Peng, Zhenghao (Mark) and Zhou, Bolei},
  year={2023},
  pages={3894--3920},
  collection={NeurIPS 2023} }

@article{guo2026bedi,
  title={{BEDI}: A Comprehensive Benchmark for Evaluating Embodied Agents on UAVs},
  author={Guo, Mingning and Wu, Mengwei and He, Jiarun and Li, Shaoxian and Li, Haifeng and Tao, Chao},
  journal={ISPRS Journal of Photogrammetry and Remote Sensing},
  year={2026}
}

@article{yao2024aeroverse,
  author = {Fanglong Yao and Yuanchang Yue and Youzhi Liu and Xian Sun and Kun Fu},
  title = {AeroVerse: UAV-Agent Benchmark Suite for Simulating, Pre-training, Finetuning, and Evaluating Aerospace Embodied World Models},
  journal = {arXiv preprint arXiv:2408.15511},
  year = {2024},
  eprint = {2408.15511},
  archivePrefix = {arXiv}
}

@article{hu2023gaia1,
  author = {Anthony Hu and Lloyd Russell and Hudson Yeo and Zak Murez and George Fedoseev and Alex Kendall and Jamie Shotton and Gianluca Corrado},
  title = {GAIA-1: A Generative World Model for Autonomous Driving},
  journal = {arXiv preprint arXiv:2309.17080},
  year = {2023},
  eprint = {2309.17080},
  archivePrefix = {arXiv}
}

@inbook{wang2023drivedreamer,
  title={DriveDreamer: Towards Real-World-Drive World Models for Autonomous Driving},
  ISBN={9783031731952},
  ISSN={1611-3349},
  url={http://dx.doi.org/10.1007/978-3-031-73195-2_4},
  DOI={10.1007/978-3-031-73195-2_4},
  booktitle={Computer Vision -- ECCV 2024},
  publisher={Springer Nature Switzerland},
  author={Wang, Xiaofeng and Zhu, Zheng and Huang, Guan and Chen, Xinze and Zhu, Jiagang and Lu, Jiwen},
  year={2024},
  month={Nov}, pages={55--72} }

@inproceedings{bruce2024genie,
  author = {Bruce, Jake and Dennis, Michael and Edwards, Ashley and Parker-Holder, Jack and Shi, Yuge and Hughes, Edward and Lai, Matthew and Mavalankar, Aditi and Steigerwald, Richie and Apps, Chris and others},
  title = {Genie: Generative Interactive Environments},
  booktitle = {Proceedings of the 41st International Conference on Machine Learning},
  publisher = {PMLR},
  year = {2024}
}

@inproceedings{das2018embodiedqa,
  author = {Das, Abhishek and Datta, Samyak and Gkioxari, Georgia and Lee, Stefan and Parikh, Devi and Batra, Dhruv},
  title = {Embodied Question Answering},
  booktitle = {Proceedings of the IEEE Conference on Computer Vision and Pattern Recognition},
  year = {2018}
}

@inproceedings{thomason2020cvdn,
  author = {Thomason, Jesse and Murray, Michael and Cakmak, Maya and Zettlemoyer, Luke},
  title = {Vision-and-Dialog Navigation},
  booktitle = {Proceedings of the Conference on Robot Learning},
  year = {2020}
}

@inproceedings{qi2020reverie,
  author = {Qi, Yuankai and Wu, Qi and Shen, Chunhua and Zhang, Hanwang and Wang, Jianfeng and Yuille, Alan and Torr, Philip},
  title = {REVERIE: Remote Embodied Visual Referring Expression in Real Indoor Environments},
  booktitle = {Proceedings of the IEEE/CVF Conference on Computer Vision and Pattern Recognition},
  year = {2020}
}

@inproceedings{ku2020rxr,
  author = {Ku, Alexander and Anderson, Peter and Patel, Roma and Ie, Eugene and Baldridge, Jason},
  title = {Room-Across-Room: Multilingual Vision-and-Language Navigation with Dense Spatiotemporal Grounding},
  booktitle = {Proceedings of the 2020 Conference on Empirical Methods in Natural Language Processing},
  year = {2020}
}

@inproceedings{deitke2020robothor,
  author = {Deitke, Matt and Han, Winson and Kolve, Eric and Mottaghi, Roozbeh and Salvador, Jordi and Schwenk, Dustin and VanderBilt, Eli and Wallingford, Matthew and Weihs, Luca and Herrasti, Alvaro and Gordon, Daniel and Ehsani, Kiana and Farhadi, Ali},
  title = {RoboTHOR: An Open Simulation-to-Real Embodied AI Platform},
  booktitle = {Proceedings of the IEEE/CVF Conference on Computer Vision and Pattern Recognition},
  year = {2020}
}

@inproceedings{krantz2020vlnce,
  author = {Krantz, Jacob and Wijmans, Erik and Majumdar, Arjun and Batra, Dhruv and Lee, Stefan},
  title = {Beyond the Nav-Graph: Vision-and-Language Navigation in Continuous Environments},
  booktitle = {Proceedings of the European Conference on Computer Vision},
  year = {2020}
}

@inproceedings{padmakumar2022teach,
  author = {Padmakumar, Vishakh and Thomason, Jesse and Shrivastava, Ayush and Lange, Patrick and Narayan-Chen, Anjali and Gella, Spandana and Piramuthu, Robinson and Tur, Gokhan and Hakkani-Tur, Dilek},
  title = {TEACh: Task-driven Embodied Agents that Chat},
  booktitle = {Proceedings of the AAAI Conference on Artificial Intelligence},
  year = {2022}
}

@inproceedings{shridhar2021alfworld,
  author = {Shridhar, Mohit and Yuan, Xingdi and C{\^o}t{\'e}, Marc-Alexandre and Bisk, Yonatan and Trischler, Adam and Hausknecht, Matthew},
  title = {ALFWorld: Aligning Text and Embodied Environments for Interactive Learning},
  booktitle = {Proceedings of the International Conference on Learning Representations},
  year = {2021}
}

@inproceedings{srivastava2022behavior1k,
  author = {Srivastava, Sanjana and Li, Chengshu and Lingelbach, Michael and Mart{\'i}n-Mart{\'i}n, Roberto and Xia, Fei and Vainio, Kent Elliott and Lian, Zheng and Gokmen, Cem and Buch, Shyamal and Liu, Karen and Savarese, Silvio and Gweon, Hyowon and Wu, Jiajun and Fei-Fei, Li},
  title = {BEHAVIOR-1K: A Benchmark for Embodied AI with 1,000 Everyday Activities and Realistic Simulation},
  booktitle = {Proceedings of the Conference on Robot Learning},
  year = {2022}
}

@inproceedings{yu2020metaworld,
  author = {Yu, Tianhe and Quillen, Deirdre and He, Zhanpeng and Julian, Ryan and Hausman, Karol and Finn, Chelsea and Levine, Sergey},
  title = {Meta-World: A Benchmark and Evaluation for Multi-Task and Meta Reinforcement Learning},
  booktitle = {Proceedings of the Conference on Robot Learning},
  year = {2020}
}

@article{zhu2020robosuite,
  author = {Zhu, Yuke and Wong, Josiah and Mandlekar, Ajay and Mart{\'i}n-Mart{\'i}n, Roberto and Joshi, Abhishek and Nasiriany, Soroush and Zhu, Yifeng},
  title = {robosuite: A Modular Simulation Framework and Benchmark for Robot Learning},
  journal = {arXiv preprint arXiv:2009.12293},
  year = {2020}
}

@inproceedings{mandlekar2021robomimic,
  author = {Mandlekar, Ajay and Xu, Danfei and Wong, Josiah and Nasiriany, Soroush and Wang, Chen and Kulkarni, Rohun and Fei-Fei, Li and Savarese, Silvio and Zhu, Yuke and Mart{\'i}n-Mart{\'i}n, Roberto},
  title = {What Matters in Learning from Offline Human Demonstrations for Robot Manipulation},
  booktitle = {Proceedings of the Conference on Robot Learning},
  year = {2021}
}

@inproceedings{makoviychuk2021isaacgym,
  author = {Makoviychuk, Viktor and Wawrzyniak, Lukasz and Guo, Yunrong and Lu, Michelle and Storey, Kier and Macklin, Miles and Hoeller, David and Rudin, Nikita and Allshire, Arthur and Handa, Ankur and State, Gavriel},
  title = {Isaac Gym: High Performance GPU-Based Physics Simulation for Robot Learning},
  booktitle = {Advances in Neural Information Processing Systems},
  year = {2021}
}

@inproceedings{padalkar2023bridgedata,
  author = {Padalkar, Abhinav and Pooley, Archie and Jain, Ajay and Bewley, Alex and Herzog, Alexander and Irpan, Alex and Khazatsky, Alexander and Rai, Akshara and Singh, Anikait and Brohan, Anthony and Raffin, Antonin and others},
  title = {BridgeData V2: A Dataset for Robot Learning at Scale},
  booktitle = {Proceedings of the Conference on Robot Learning},
  year = {2023}
}

@inproceedings{fang2023rh20t,
  author = {Fang, Hao-Shu and Fang, Hongjie and Tang, Zhenyu and Liu, Jirong and Wang, Chenxi and Wang, Junbo and Zhu, Haoyi and Lu, Cewu},
  title = {RH20T: A Comprehensive Robotic Dataset for Learning Diverse Skills in One-Shot},
  booktitle = {Proceedings of the IEEE International Conference on Robotics and Automation},
  year = {2024}
}

@inproceedings{fu2024mobilealoha,
  author = {Fu, Zipeng and Zhao, Tony Z. and Finn, Chelsea},
  title = {Mobile ALOHA: Learning Bimanual Mobile Manipulation with Low-Cost Whole-Body Teleoperation},
  booktitle = {Proceedings of the Conference on Robot Learning},
  year = {2024}
}

@inproceedings{brohan2022rt1,
  author = {Brohan, Anthony and Brown, Noah and Carbajal, Justice and Chebotar, Yevgen and Driess, Danny and Finn, Chelsea and Florence, Pete and Gopalakrishnan, Keerthana and Hausman, Karol and Herzog, Alexander and Hsu, Jasmine and Ichter, Brian and Irpan, Alex and Joshi, Nikhil and Julian, Ryan and Kalashnikov, Dmitry and Kuang, Yuheng and Lee, Kanishka and Levine, Sergey and Lu, Yao and Michalewski, Henryk and Mordatch, Igor and Pertsch, Karl and Rao, Kanishka and Reymann, Karol and Ryoo, Michael and Salazar, Grecia and Sanketi, Pannag and Sermanet, Pierre and Singh, Jaspiar and Singh, Anikait and Soricut, Radu and Tran, Kevin and Vanhoucke, Vincent and Vuong, Quan and Xia, Fei and Xiao, Ted and Xu, Peng and Xu, Sichun and Yu, Tianhe and Zitkovich, Brianna},
  title = {RT-1: Robotics Transformer for Real-World Control at Scale},
  booktitle = {Proceedings of Robotics: Science and Systems},
  year = {2023}
}

@inproceedings{brohan2023rt2,
  author = {Brohan, Anthony and Brown, Noah and Carbajal, Justice and Chebotar, Yevgen and Driess, Danny and Finn, Chelsea and Florence, Pete and Gopalakrishnan, Keerthana and Hausman, Karol and Herzog, Alexander and Hsu, Jasmine and Ichter, Brian and Irpan, Alex and Jackson, Tobin and Jesmonth, Shaunak and Joshi, Nikhil and Julian, Ryan and Kalashnikov, Dmitry and Kuang, Yuheng and Lee, Kanishka and Levine, Sergey and Lu, Yao and Michalewski, Henryk and Mordatch, Igor and Pertsch, Karl and Rao, Kanishka and Reymann, Karol and Ryoo, Michael and Salazar, Grecia and Sanketi, Pannag and Sermanet, Pierre and Singh, Jaspiar and Singh, Anikait and Soricut, Radu and Tran, Kevin and Vanhoucke, Vincent and Vuong, Quan and Xia, Fei and Xiao, Ted and Xu, Peng and Xu, Sichun and Yu, Tianhe and Zitkovich, Brianna},
  title = {RT-2: Vision-Language-Action Models Transfer Web Knowledge to Robotic Control},
  booktitle = {Proceedings of the Conference on Robot Learning},
  year = {2023}
}

@inproceedings{ahn2022saycan,
  author = {Ahn, Michael and Brohan, Anthony and Brown, Noah and Chebotar, Yevgen and Cortes, Omar and David, Byron and Finn, Chelsea and Fu, Chuyuan and Gopalakrishnan, Keerthana and Hausman, Karol and Herzog, Alexander and Ho, Daniel and Hsu, Jasmine and Ichter, Brian and Irpan, Alex and Jang, Eric and Ruano, Rosario Jauregui and Jeffrey, Kyle and Jesmonth, Shaunak and Joshi, Nikhil and Julian, Ryan and Kalashnikov, Dmitry and Kuang, Yuheng and Lee, Kanishka and Levine, Sergey and Lu, Yao and Luu, Linda and Parada, Carolina and Pastor, Peter and Quiambao, Jeannette and Rao, Kanishka and Rettinghouse, Jonathon and Reyes, Diego and Sermanet, Pierre and Sievers, Nicolas and Tan, Clayton and Toshev, Alexander and Vanhoucke, Vincent and Xia, Fei and Xiao, Ted and Xu, Peng and Xu, Sichun and Yan, Mengyuan and Yu, Andy},
  title = {Do As I Can, Not As I Say: Grounding Language in Robotic Affordances},
  booktitle = {Proceedings of the Conference on Robot Learning},
  year = {2022}
}

@inproceedings{driess2023palme,
  author = {Driess, Danny and Xia, Fei and Sajjadi, Mehdi S. M. and Lynch, Corey and Chowdhery, Aakanksha and Ichter, Brian and Wahid, Ayzaan and Tompson, Jonathan and Vuong, Quan and Yu, Tianhe and Huang, Wenlong and Chebotar, Yevgen and Sermanet, Pierre and Duckworth, Daniel and Levine, Sergey and Vanhoucke, Vincent and Hausman, Karol and Toussaint, Marc and Greff, Klaus and Zeng, Andy and Mordatch, Igor and Florence, Pete},
  title = {PaLM-E: An Embodied Multimodal Language Model},
  booktitle = {Proceedings of the 40th International Conference on Machine Learning},
  publisher = {PMLR},
  year = {2023}
}

@inproceedings{jiang2022vima,
  author = {Jiang, Yunfan and Gupta, Agrim and Zhang, Zichen and Wang, Guanzhi and Dou, Yongqiang and Chen, Yanjun and Fei-Fei, Li and Anandkumar, Anima and Zhu, Yuke and Fan, Linxi},
  title = {VIMA: General Robot Manipulation with Multimodal Prompts},
  booktitle = {Proceedings of the 40th International Conference on Machine Learning},
  publisher = {PMLR},
  year = {2023}
}

@inproceedings{liang2023codeaspolicies,
  author = {Liang, Jacky and Huang, Wenlong and Xia, Fei and Xu, Peng and Hausman, Karol and Ichter, Brian and Florence, Pete and Zeng, Andy},
  title = {Code as Policies: Language Model Programs for Embodied Control},
  booktitle = {Proceedings of the IEEE International Conference on Robotics and Automation},
  year = {2023}
}

@inproceedings{huang2023voxposer,
  author = {Huang, Wenlong and Wang, Chen and Zhang, Ruohan and Li, Yunzhu and Wu, Jiajun and Fei-Fei, Li},
  title = {VoxPoser: Composable 3D Value Maps for Robotic Manipulation with Language Models},
  booktitle = {Proceedings of the Conference on Robot Learning},
  year = {2023}
}

@inproceedings{grauman2021ego4d,
  author = {Grauman, Kristen and Westbury, Andrew and Byrne, Eugene and Chavis, Zachary and Furnari, Antonino and Girdhar, Rohit and Hamburger, Jackson and Jiang, Hao and Liu, Miao and Liu, Xingyu and Martin, Miguel and Nagarajan, Tushar and Radosavovic, Ilija and Ramakrishnan, Santhosh K. and Ryan, Fiona and Sharma, Jayant and Wray, Michael and Xu, Mengmeng and Xu, Eric Zhongcong and Zhao, Chen and Bansal, Siddhant and Batra, Dhruv and Cartillier, Vincent and Crane, Sean and Do, Tien and Doulaty, Marzieh and Erapalli, Apoorva and Feichtenhofer, Christoph and Fragomeni, Andrea and Fu, Qichen and Gebreselasie, Abrham and Gonz{\'a}lez, Cristina and Hillis, James and Huang, Xuhua and Huang, Yifei and Jia, Wenqi and Khoo, Wei and Kol{\'a}r, Jachym and Kottur, Satwik and Kumar, Anurag and Landini, Federico and Li, Chao and Li, Yanghao and Mangalam, Karttikeya and Modhugu, Rahul and Munro, Jonathan and Murrell, Tullie and Nishiyasu, Takuma and Price, Will and Puentes, Paola Ruiz and Ramazanova, Merey and Sari, Leda and Somasundaram, Kiran and Southerland, Audrey and Sugano, Yusuke and Tao, Ruijie and Vo, Minh and Wang, Yuchen and Wu, Xindi and Yagi, Takuma and Zhao, Yunyi and Zhu, Xuehan and Arbelaez, Pablo and Crandall, David and Damen, Dima and Farinella, Giovanni Maria and Fuegen, Christian and Ghanem, Bernard and Ithapu, Vamsi Krishna and Javidi, Tara and Joo, Hanbyul and Kitani, Kris and Li, Haizhou and Newcombe, Richard and Oliva, Aude and Park, Hyun Soo and Rehg, James M. and Sato, Yoichi and Shi, Jianbo and Shou, Mike Zheng and Torralba, Antonio and Torresani, Lorenzo and Yan, Mingfei and Malik, Jitendra},
  title = {Ego4D: Around the World in 3,000 Hours of Egocentric Video},
  booktitle = {Proceedings of the IEEE/CVF Conference on Computer Vision and Pattern Recognition},
  year = {2022}
}

@inproceedings{azuma2022scanqa,
  title={Scanqa: 3d question answering for spatial scene understanding},
  author={Azuma, Daichi and Miyanishi, Taiki and Kurita, Shuhei and Kawanabe, Motoaki},
  booktitle={proceedings of the IEEE/CVF conference on computer vision and pattern recognition},
  pages={19129--19139},
  year={2022}
}

@article{ma2022sqa3d,
  title={Sqa3d: Situated question answering in 3d scenes},
  author={Ma, Xiaojian and Yong, Silong and Zheng, Zilong and Li, Qing and Liang, Yitao and Zhu, Song-Chun and Huang, Siyuan},
  journal={arXiv preprint arXiv:2210.07474},
  year={2022}
}

@inproceedings{grauman2024ego,
  title={Ego-exo4d: Understanding skilled human activity from first-and third-person perspectives},
  author={Grauman, Kristen and Westbury, Andrew and Torresani, Lorenzo and Kitani, Kris and Malik, Jitendra and Afouras, Triantafyllos and Ashutosh, Kumar and Baiyya, Vijay and Bansal, Siddhant and Boote, Bikram and others},
  booktitle={Proceedings of the IEEE/CVF Conference on Computer Vision and Pattern Recognition},
  pages={19383--19400},
  year={2024}
}

@misc{qiao2025navbench,
  title = {{NavBench}: Probing Multimodal Large Language Models for Embodied Navigation},
  author = {Qiao, Yanyuan and Hong, Haodong and Lyu, Wenqi and An, Dong and Zhang, Siqi and Xie, Yutong and Wang, Xinyu and Wu, Qi},
  year = {2025}
}

@misc{caluwaerts2023barkour,
  title = {{Barkour}: Benchmarking Animal-Level Agility with Quadruped Robots},
  author = {Caluwaerts, Ken and Iscen, Atil and Kew, J. Chase and Yu, Wenhao and Zhang, Tingnan and Freeman, Daniel and Lee, Kuang-Huei and Lee, Lisa and Saliceti, Stefano and Zhuang, Vincent and others},
  year = {2023},
  eprint = {2305.14654},
  archivePrefix = {arXiv},
  primaryClass = {cs.RO}
}

@article{hoeller2023anymal,
  title = {{ANYmal} Parkour: Learning Agile Navigation for Quadrupedal Robots},
  author = {Hoeller, David and Rudin, Nikita and Sako, Dhionis and Hutter, Marco},
  journal = {Science Robotics},
  year = {2024},
  note = {arXiv:2306.14874}
}

\end{document}